\documentclass[11pt]{imsart}
\usepackage{geometry}
\newgeometry{vmargin={20mm}, hmargin={38mm,38mm}}   % set the margins

\RequirePackage{amsthm,amsmath,amsfonts} %,amssymb
\RequirePackage[authoryear]{natbib}%% uncomment this for author-year citations
\RequirePackage[colorlinks,citecolor=blue,urlcolor=blue]{hyperref}
\RequirePackage{graphicx}

\startlocaldefs
\numberwithin{equation}{section}
\theoremstyle{plain}

\newtheorem{theorem}{Theorem}[section]
\newtheorem{lemma}[theorem]{Lemma}
\newtheorem{proposition}{Proposition}[section]

\theoremstyle{definition}

\newtheorem{example}{Example}[section]

\newtheorem{remark}{Remark}[section]
\newtheorem{assumption}{Assumption}

\theoremstyle{remark}

\usepackage{enumerate}

\usepackage{url} % not crucial - just used below for the URL 

\usepackage{tabularx,booktabs}
\usepackage{stackengine}
\usepackage[normalem]{ulem}
\usepackage{cleveref}
\usepackage{multirow}
\usepackage{algorithmic}
\usepackage{algorithm}
\usepackage{dsfont}
\usepackage{empheq}
\usepackage{bm}
\usepackage{bbm}
\usepackage{hhline}
\usepackage{mathtools}
\usepackage{stmaryrd}
\usepackage{algorithm}
\usepackage{subcaption}
\usepackage{cleveref}

\newcommand{\mb}{\mathbb}
\newcommand{\E}{\mathbb{E}}
\newcommand{\s}{\geq}
\newcommand{\m}{\leq}

\newcommand{\1}{\mathds{1}}
\newcommand{\tti}{\underset{t \rightarrow +\infty}{\longrightarrow}}

\newcommand{\ie}{i.e. }
\makeatletter
\let\oldabs\abs
\def\abs{\@ifstar{\oldabs}{\oldabs*}}
\let\oldnorm\norm
\def\norm{\@ifstar{\oldnorm}{\oldnorm*}}
\makeatother

\newcommand{\PP}[1][]{\ifthenelse{\equal{#1}{}}{\ensuremath{\mathbb{P}}}{\ensuremath{\mathbb{P}\left\{ #1 \right\} }}}
\newcommand{\EE}[1][]{\ifthenelse{\equal{#1}{}}{\ensuremath{\mathbb E}}{\ensuremath{{\mathbb E}\left[ #1 \right]}}}
\def\E{\EE}

\def\Var{\mathbb{V}\mathrm{ar}}

\newcommand{\new}[1]{{#1}}

\newcommand{\sphere}{\mathbb{S}}

\newcommand{\rset}{\mathbb{R}}

\newcommand{\ind}{\mathds{1}}
\newcommand{\un}{\ind}
\newcommand{\ud}{\,\mathrm{d}}
\newcommand{\comment}[1]{\ignorespaces}

\newcommand{\riskinf}{R_\infty}
\newcommand{\riskext}{R_{P_\infty}}
\newcommand{\bayesext}{f_{P_\infty}^*}
\newcommand{\riskempnk}{\hat{R}_{k}}
\newcommand{\risknktilde}{\Tilde{R}_{t_{n,k}}}
\newcommand{\hatf}{\hat{f}_{k}}
\newcommand{\hath}{\hat{h}_{k}}
\newcommand{\risknk}{R_{t_{n,k}}}
\newcommand{\B}{\mathbb{B}}
\newcommand{\hH}{h_{\mathcal{H}}}
\newcommand{\Mo}{\mathbb{M}_{\mathbb{O}}}

\endlocaldefs

\begin{document}
\begin{frontmatter}
  \title{On Regression in Extreme Regions} %\title{A sample article title}
\runtitle{On Regression in Extreme Regions}

%\thankstext{T1}{A sample additional note to the title.}

\begin{aug}
%%%%%%%%%%%%%%%%%%%%%%%%%%%%%%%%%%%%%%%%%%%%%%%
%% Only one address is permitted per author. %%
%% Only division, organization and e-mail is %%
%% included in the address.                  %%
%% Additional information can be included in %%
%% the Acknowledgments section if necessary. %%
%% ORCID can be inserted by command:         %%
%% \orcid{0000-0000-0000-0000}               %%
%%%%%%%%%%%%%%%%%%%%%%%%%%%%%%%%%%%%%%%%%%%%%%%
  \author[A]{%\inits{S.C.}
    \fnms{Stephan}~\snm{Cl\'emen\c{c}on}\ead[label=e1]{stephan.clemencon@telecom-paris.fr}},
  \author[A]{%\inits{N.H.}
    \fnms{Nathan}~\snm{Huet}\ead[label=e2]{nathan.huet@telecom-paris.fr}}
\and
\author[B]{%\inits{A.S.}
  \fnms{Anne}~\snm{Sabourin}\ead[label=e3]{anne.sabourin@u-paris.fr}}
%%%%%%%%%%%%%%%%%%%%%%%%%%%%%%%%%%%%%%%%%%%%%%
%% Addresses                                %%
%%%%%%%%%%%%%%%%%%%%%%%%%%%%%%%%%%%%%%%%%%%%%%
\address[A]{LTCI, T\'el\'ecom Paris, Institut Polytechnique de Paris, France\printead[presep={,\ }]{e1}\printead[presep={,\ }]{e2}}

%\address[A]{LTCI, T\'el\'ecom Paris, Institut Polytechnique de Paris\printead[presep={,\ }]{e2}}

\address[B]{MAP5 - CNRS UMR 8145, Universit\'e Paris Cit\'e, France \printead[presep={,\ }]{e3}}

% \author[A]{\fnms{First}~\snm{Author}\ead[label=e1]{first@somewhere.com}},
% \author[B]{\fnms{Second}~\snm{Author}\ead[label=e2]{second@somewhere.com}\orcid{0000-0000-0000-0000}}
% \and
% \author[B]{\fnms{Third}~\snm{Author}\ead[label=e3]{third@somewhere.com}}
% %%%%%%%%%%%%%%%%%%%%%%%%%%%%%%%%%%%%%%%%%%%%%%
% %% Addresses                                %%
% %%%%%%%%%%%%%%%%%%%%%%%%%%%%%%%%%%%%%%%%%%%%%%
% \address[A]{Department,
% University or Company Name\printead[presep={,\ }]{e1}}

% \address[B]{Department,
% University or Company Name\printead[presep={,\ }]{e2,e3}}
% \runauthor{F. Author et al.}
\end{aug}

\begin{abstract}

We establish a statistical learning theoretical framework aimed at extrapolation, or out-of-domain generalization, on the unobserved tails of covariates in continuous regression problems. Our strategy involves performing statistical regression on a subsample of observations with continuous labels that are the furthest away from the origin, focusing specifically on their angular components. The underlying assumptions of our approach are grounded in the theory of multivariate regular variation, a cornerstone of extreme value theory.
We address the stylized problem of nonparametric least squares regression with predictors chosen from a Vapnik-Chervonenkis class.

This work contributes to a broader initiative to develop statistical learning theoretical foundations for supervised learning strategies that enhance performance on the supposedly heavy tails of covariates. Previous efforts in this area have focused exclusively on binary classification on extreme covariates.
Although the continuous target setting necessitates different techniques and regularity assumptions, our main results echo findings from earlier studies. We quantify the predictive performance on tail regions in terms of excess risk, presenting it as a finite sample risk bound with a clear bias-variance decomposition. Numerical experiments with simulated and real data illustrate our theoretical findings.

\end{abstract}



\end{frontmatter}
%%%%%%%%%%%%%%%%%%%%%%%%%%%%%%%%%%%%%%%%%%%%%%
%% Please use \tableofcontents for articles %%
%% with 50 pages and more                   %%
%%%%%%%%%%%%%%%%%%%%%%%%%%%%%%%%%%%%%%%%%%%%%%
%\tableofcontents

\section{Introduction}

In the standard supervised learning setup, $(X,Y)$ is a pair of random
variables with distribution $P$, where the target
$Y \in\mathcal{Y} \subset \rset$ is a real-valued random variable (the
output) and the predictor (or covariable) $X \in\mathcal{X}$ models
some input information hopefully useful to predict $Y$.  Given a cost
function $c(y, \hat y)\ge 0$, the classical problem is to build, from
a training dataset
$\mathcal{D}_n=\{(X_1,Y_1),\; \ldots,\; (X_n,Y_n) \}$ composed of
$n\geq 1$ independent copies of $(X,Y)$, a mapping
$f:\rset^d\rightarrow \mathbb{R}$ in order to compute a `good'
prediction $f(X)$ for $Y$, with risk
$ R_P(f)=\mathbb{E}\left[c(Y, f(X)) \right] $ as small as possible. A
natural choice for the cost function with a continuous target is the
squared error loss $c(y,\hat y)=(y-\hat y)^2$, while binary
classification problems are typically formalized with the Hamming loss
$c(y,\hat y) = \un\{y\neq \hat y\}$, although convex surrogate losses
are then typically preferred in practice.  A natural strategy consists
in solving the Empirical Risk Minimization (ERM) problem
$\min_{f\in \mathcal{F}}R_{\hat{P}_n}(f)$, where $\mathcal{F}$ is a
class of functions sufficiently rich to include an approximate
minimizer of $R_P$ and $\hat{P}_n$ is an empirical version of $P$
based on $\mathcal{D}_n$.  The performance of predictive functions
$\hat{f}$ obtained this way has been extensively investigated in the
statistical learning literature as reviewed in \emph{e.g.}
\cite{devroye2013probabilistic,GKKW02,lugosi2002pattern,Massart2007}.
Confidence upper bounds for the excess of quadratic risk
$R_P(\hat{f})-R_P^*=\mathbb{E}[(Y-\hat{f}(X))^2\mid
\mathcal{D}_n]-R_P^*$ have been established in \cite{LM16} by means of
concentration inequalities for empirical processes \cite{BLM2013}.

Here we consider the different problem of building prediction
functions which would be reliable in a `crisis scenario', or under
`covariate shift', where the covariates vector takes unusually large
values and thus belongs to regions where few or even no such large
examples have been observed in the past. The goal pursued here is not
to predict extreme realizations of the target, but rather to learn a
regression function which would behave well with test data with a
covariate vector belonging to an unseen domain, namely with an
unusually large norm. An example of application would be to predict
the direction of the wind (a bounded variable between $0$ and $2\pi$,
say), given unusually large values of (potentially unbounded)
explanatory variables such as temperature, pressure, wind speed at
several locations. Another application would be predicting the
proportion of patients admitted to a specific hospital department,
given a large number of patients across all departments and other
external explanatory variables that may take extreme values (such as
temperature, air quality, \dots).  Addressing this generic task from
the perspective of multivariate extreme value theory is a natural
strategy that has gained increasing interest in recent years, with a
variety of viewpoints further detailed in the paragraph `Related
works' below in this introduction. The closest existing works focus on
classification on the tails of the covariates, as discussed next.

\paragraph{Supervised learning on covariate  tails, Regression vs. Classification.}

The present work is part of a broader effort to establish the
theoretical foundations for learning algorithms dedicated to covariate
tail extrapolation, with finite sample statistical guarantees
regarding the excess risk of the algorithm. We operate within a
model-agnostic, nonparametric framework grounded in multivariate
Extreme Value Analysis (EVA).  While previous works
\citep{aghbalou2022cross,clemencon+j+s+s:2021,jalalzai2018binary} have
focused exclusively on binary classification problems where
\(\mathcal{Y} = \{\pm 1\}\), we address here the related yet distinct
problem of continuous regression, which may be considered as a `second
chapter' in learning theory, following binary classification.

It has long been recognized in the statistical learning and
mathematical statistics literature that binary classification and
continuous regression, although similar in spirit, necessitate
different analyses and yield distinct statistical results. The latter
is generally regarded as more challenging than the former, as
discussed in Chapters 6 and 7 of \cite{devroye2013probabilistic} and
Chapter 1, Section 1.4 of \cite{GKKW02}. It is thus not immediately
evident that probabilistic and statistical results obtained in the
simpler context of classification should readily extend to the more
complex continuous setting considered here.  In particular, the main
tail regularity assumptions made in \cite{jalalzai2018binary} do not
easily translate to the continuous target setting, as discussed in the
following dedicated paragraph.

For mathematical convenience and in line with the envisioned
applications, we assume that the target \( Y \) is bounded, where
\(\mathcal{Y} = [-M, M]\) for some \( M > 0 \). In statistical
modeling, the assumption of a bounded target can be contentious,
particularly when dealing with unbounded covariates and multivariate
extremes.  Also considering extreme covariates, rather than targets,
is somewhat unconventional in the field of EVA, where extremality
typically concerns the target, not the covariates, \emph{e.g.} in the
problem of extreme quantile regression. This is further discussed in
Remark~\ref{rem:HToutput-input}.

With this in mind, we propose a rescaling mechanism applicable to
unbounded targets in Example~\ref{ex:predictionRVvect} and
Proposition~\ref{prop:exampleMissingComponent}, which is designed to
enforce the bounded target assumption effectively. This approach is
supported by our numerical experiments, including those conducted on a
real dataset.  This connection with relatively standard multivariate
EVA setups constitutes a major improvement upon the work of
\cite{jalalzai2018binary}.

The goal of supervised learning in the covariates' tail as formalized
in \cite{jalalzai2018binary} is to achieve good prediction performance
on the event that $\|X\|$ is unusually large, namely when their norm
$\lVert X\rVert$ exceeds some (asymptotically) large threshold $t>0$.
The choice of the norm is unimportant in theory, and is typically
determined by the application context.  The threshold $t$ depends on
the observations, since `large' should be naturally understood as
large with respect to the vast majority of data observed. Hence,
extreme observations are rare by nature and severely underrepresented
in the training dataset.  Consequently, the impact of prediction
errors in extreme regions of the input space on the global regression
error of $\hat{f}$ is generally negligible.  Indeed, the law of total
probability yields
   \begin{equation}\label{eq:decomp1}
     \begin{aligned}
       R_P(f)~=~&\mathbb{P}\{ \lVert X \rVert \s t \}
                  \mathbb{E}\left[c\left(Y-f(X)\right)\mid \lVert X \rVert \s t \right]\\
                ~+~     &\mathbb{P}\{ \lVert X \rVert < t \}
                  \mathbb{E}\left[c\left(Y-f(X)\right)\mid \lVert X \rVert < t \right].
       \end{aligned}
\end{equation}
The above decomposition involves a conditional error term relative to
excesses of $\lVert X\rVert$ above $t$, the \emph{conditional
  risk}, 
\begin{equation*}
 R_{t}(f):=\mathbb{E}\left[c\left(Y- f(X)\right)\mid \lVert X \rVert \s t \right]. 
 \end{equation*}

 The informal purpose of our analysis, as in
 \cite{jalalzai2018binary}, is to construct a predictive function
 $\hat f$ that (approximately) minimizes $R_t(f)$ for all $t > t_0$,
 with $t_0$ being a large threshold. Since an approximate minimizer of
 $R_t$ might not be suitable for minimizing $R_{t'}$ when $t' > t$, to
 ensure robust extrapolation performance for our learned function, our
 formal focus is on minimizing the \emph{asymptotic conditional risk}
 defined as
 \begin{equation}\label{eq:asym_cond_risk}
\riskinf(f) := \limsup_{t\to +\infty} R_t(f) = \limsup_{t\to +\infty} \EE [c(Y - f(X) )\ | \ \lVert X\rVert \s t ].
\end{equation}
With the quadratic cost for regression, any function that coincides
with the regression function $f^*(x)=\E [Y \mid X=x]$ on the region
$ \{ x \in \mathcal{X}, \lVert x\rVert \s t \}$ for some $t>0$,
minimizes the risk functional $R_t$, and thus also $\riskinf$.  In
other words $R_{\infty}^*:= \inf_{f}R_{\infty}(f) = R_{\infty}(f^*)$.
However, the straightforward theoretical solution $f^*$ is of course
unknown.  In view of Equation~\eqref{eq:decomp1} it is evident that an
estimate $\hat f$ of $f^*$ produced by an ERM strategy with good
overall empirical performance, may not necessarily enjoy good
performance when restricted to extreme regions.  Put another way,
there is no guarantee that the conditional risk $R_t(\hat f)$ (or
$\riskinf(\hat f)$) would be small.  To summarize, the
\emph{Supervised learning problem on extremes} refers here to the task
of constructing a prediction function $\hat f$ based on
$\mathcal{D}_n$ which approximately minimizes $\riskinf$.  For
simplicity, we consider only the quadratic cost function
$c(y,\hat y)=(y-\hat y)^2$ throughout, although our results may be
straightforwardly extended to other natural losses, such as the
pinball loss for quantile regression.

 \paragraph{Tail regularity assumptions.}
 In order to develop a specific ERM framework relative to $\riskinf$
 with provable guarantees, regularity assumptions are required
 regarding the tail behavior of the pair $(X,Y)$, with respect to the
 first component.  Heuristically, the aim of these assumptions (in
 \cite{jalalzai2018binary} and in the present work) is to ensure that,
 for a fixed \( x \neq 0 \), as \( t \to +\infty \), the regression
 function \( f^*(tx) \) converges to a limit. By construction, this
 limit depends solely on the direction \( x/\|x\| \), and subsequent
 arguments aim to provide guarantees for an extrapolation strategy
 based on learning a prediction function that takes as input only the
 angular component of the covariates with the largest norm.
 \emph{Multivariate regular variation} hypotheses are very flexible in
 the sense that they correspond to a large nonparametric class of
 heavy-tailed distributions. These assumptions, or slightly weaker
 ones such as \emph{Maximum Domain of Attraction} conditions are at
 the heart of EVA (e.g., the monographs
 \cite{dHF06,resnick2013extreme}). They are frequently used in
 applications where the impact of extreme observations should be
 enhanced, or not neglected at the minimum.  For classification on
 extreme covariates, \cite{jalalzai2018binary} assume that the class
 of conditional distributions $\mathcal{L}(X~|~Y = \pm 1)$ are
 multivariate regularly varying with identical tail index, without
 obvious possible extension to continuous targets. Indeed one natural
 approach would be to assume regular variation of the conditional
 distributions $\mathcal{L}(X|Y=y)$, almost everywhere. However this
 would lead to measure theoretic complications, and it would be
 difficult to verify in practice and on theoretical examples.  We
 propose to bypass this issue \emph{via} \emph{one-component} regular
 variation assumptions stated and discussed in
 Section~\ref{subsec:framework} concerning the joint behavior of the
 pair $(t^{-1}X,Y)$, conditioned on $\|X\|>t$, for large $t$, see our
 Assumption~\ref{hyp:limit}.  Once again, our focus is not on extremes
 of the target, but on those of the covariates, thus the rescaling
 operation (multiplication by $t^{-1}$) affects only the
 covariate. This asymmetric treatment of different components of the
 pair $(X,Y)$ and the concept of one-component regular variation has
 become relatively standard in the EVA literature \citep[see
 \emph{e.g.}][]{engelke2020graphical,segers2020one}, or Section 3.2 in
 \cite{kulik2020heavy} and bibliographic notes of Section 3 in the
 latter reference, although leveraging it for nonparametric regression
 is, to our best knowledge, new.

\paragraph{Related works.}

Considering prediction from extreme values of the covariates, although
far less documented than prediction of an extreme target, is not
entirely unexplored from a methodological and applied perspective.
Parametric modeling approaches with applications have been considered
in \cite{cooley2012approximating,de2022regression}. These works assume
that the tail model (as \(t \to +\infty\)) for the pair \((X, Y)\) is
attained at the observed covariate \(x\), focusing on explicit
expressions for conditional distributions within this limiting
framework.  Recently, modeling strategies for `cascading extremes'
with neural networks have been proposed in \cite{de2025kolmogorov}.
Our approach stands in complete contrast to these works. Specifically,
our goal is to propose an analysis that accounts for the
sub-asymptotic nature of the observations within an ERM framework that
is agnostic to possible parametric structures of the generative
process.  Regarding alternatives to the least squares error,
\cite{buritica2024progression} addresses the problem of quantile
regression, which involves the pinball loss, extending the present
framework but focusing on one-dimensional covariates with no obvious
extensions to multivariate settings.  In higher dimensional settings,
prediction guarantees with a risk involving an additional LASSO-type
term are considered in the final section of the overview paper
\cite{clemenccon2025weak}, building upon an earlier, publicly
available version of the present work, and assuming additional
structural linearity conditions.  From an applied perspective, the
rescaling mechanism accommodating unbounded targets has been
implemented in \cite{huet2025multi}, based on the same earlier version
of this work, and it has been compared with parametric modeling
frameworks in the Multivariate Generalized Pareto setup
\citep{kiriliouk2019peaks,rootzen2018multivariate,rootzen2006multivariate}
for the purpose of reconstructing missing values in sea level and skew
surge multivariate time series.  An application of the classification
setting to sentiment analysis and label preserving data augmentation
with large language models has also been worked out in
\cite{jalalzai2020heavy}.

As mentioned above, learning theory on extreme covariates has been
explored in several earlier works focused on classification. In
\cite{clemencon+j+s+s:2021}, the guarantees of
\cite{jalalzai2018binary} are extended to scenarios involving
preliminary rank-based transformations of the input $X$. Their
argument relies on controlling the deviations of the angular measure
over a class of sets, with no obvious generalization to regression
problems. The question of marginal standardization remains open in a
regression context. In another direction, \cite{aghbalou2022cross}
establish guarantees for cross-validation strategies in
classification, aiming to evaluate the generalization risk of ERM
algorithms, using the classification problem formalized in
\cite{jalalzai2018binary} as a leading example.

The idea of rescaling an unbounded target for prediction in
multivariate vectors, that was already present in the previously
mentioned earlier version of this work, has been since adapted to the
classification case (Example 2.1 in
\cite{aghbalou2022cross}). Conversely, the latter reference involves
general (real-valued) loss functions for classification, with proof
techniques similar to the ones that we employ at intermediate steps of
our analysis (Proposition~\ref{prop:concentr1}).  However, the
analysis in \cite{aghbalou2022cross} is purely statistical, leveraging
only the low probability of the event \(\{\|X\| > t\}\). Their focus
is on the error at finite levels $t$, not on the structure of the
solutions as \(t \to +\infty\).  Differently, our main result,
Theorem~\ref{cor:concentr2}, concerns the excess of $R_\infty$ risk,
at infinite levels, and incorporates additional bias terms, with
discussions of sufficient conditions for these bias terms to vanish as
the training threshold goes to infinity.

Broadening the perspective to encompass the machine learning
literature, the problem of regression in extreme regions can be
likened to a specific transfer learning or out-of-domain
generalization problem, see \emph{e.g.}
\cite{panSurveyTransferLearning2010,zhou2022domain}. Indeed, the
objective is to learn a regression function that is nearly optimal in
the target (limit) extremal domain, based on source training data in a
pre-asymptotic regime. Unlike pre-existing transfer learning and
domain adaptation approaches, the methodology we develop does not rely
on inverse probability weighting \citep{pmlr-v63-clemencon64},
estimating or learning propensity score functions
\citep{pmlr-v139-bertail21a}, or the use of Markov kernels
\citep{pfister2024extrapolationaware}. Instead, it exploits a
multivariate regular variation assumption to estimate the target loss
with guarantees. The problem at hand could also be viewed as a
specific, yet unaddressed, few-shot learning problem \citep{Wang}.

\paragraph{Contributions and paper organization.}
The goal of this paper is to complete the framework initiated in
\cite{jalalzai2018binary} and to establish a theoretical foundation
for Regression on Extremes. We consider a generic algorithmic approach
that naturally extends the method proposed for binary classification
in \cite{jalalzai2018binary}. This approach involves making
predictions based on the \emph{direction} (or \emph{angle}) of the
largest observations.

Our main contributions are twofold. First, we introduce a new set of
assumptions under which the primary structural results for
classification, specifically the particular form of optimal predictors
of angular nature, continue to hold in the case of a continuous
target. We carefully discuss these assumptions by proposing
sufficient, and arguably more interpretable, conditions, and we
provide examples directly related with classical regular variation
assumptions of densities. We also explore how and when to normalize an
unbounded target to satisfy our assumption of a bounded target.
Second, from a statistical learning perspective, our main result
establishes a bias-variance decomposition of the limit conditional
risk \( R_\infty \), where the bias term arises from the combination
of a model bias and an observation bias due to the nonasymptotic
nature of the observed data.

The paper is organized as follows. The algorithmic approach we
consider for Regression on Extremes is detailed in
Section~\ref{sec:background}. The probability framework we employ for
regression in extreme regions is described extensively therein. In
Section~\ref{sec:main}, we present our working assumptions and
demonstrate that a predictive rule using only the angular information,
i.e., of the form \( f(X) = h(X/\lVert X \rVert) \), where \( h \) is
a real-valued function defined on the hypersphere
\( \sphere = \{x \in \rset^d: \lVert x \rVert = 1\} \), achieves the
best possible performance with respect to the asymptotic risk.
Subsequently, we study the performance of a predictive rule learned by
minimizing an empirical version of~\eqref{eq:asym_cond_risk} based on
a fraction \( k/n \) of the training dataset, corresponding to the
largest \( \lVert X_i \rVert \)'s. Nonasymptotic bounds for the excess
of asymptotic risk of such an empirical (pre-asymptotic) risk
minimizer are established, demonstrating its near optimality.  Beyond
these theoretical guarantees, the performance of empirical risk
minimization on extreme covariates is supported by various numerical
experiments presented in Section~\ref{sec:exp}. Concluding remarks are
collected in Section~\ref{sec:conclusion}. To enhance readability,
certain technical details are deferred to the Appendix.

\section{A Regular Variation Framework for Regression}\label{sec:background} %Background and Preliminaries}

In this section, we propose a probabilistic framework in which
regression on extremes may be addressed, together with a dedicated
algorithmic approach, the latter being analyzed next in the subsequent
sections.  Here and throughout, $(X,Y)$ is a pair of random variables
defined on a probability space $(\Omega,\; \mathcal{A},\; \mathbb{P})$
with distribution $P$, where $Y$ is real-valued with marginal
distribution $G$ and $X=(X^{(1)},\; \ldots,\; X^{(d)})$ takes its
values in $\rset^d$, $d\geq 1$. We sometimes denote by
$\mathcal{L}(Z)$ the distribution of a random variable $Z$.  Recall
from the Introduction section that $\lVert \cdot\rVert$ is any norm on
$\mathbb{R}^d$. We denote by $\sphere$ the unit sphere for this norm
and by $\B := \{x \in \rset^d, \lVert x \rVert \m 1 \}$ the unit
ball. Let $E=\rset^d\setminus \{0_{\rset^d}\}$ be the punctured
Euclidean space.  For any measurable subset $A$ of $\rset^d$ we denote
by $\mathcal{B}(A)$ the Borel $\sigma$-algebra on $A$.  The boundary
and the closure of $A$ are respectively denoted by $\partial A$ and
$\bar{A}$, and we set $tA=\{tx:\; x\in A\}$ for all $t\in \mathbb{R}$.
By $\1\{\mathcal{E}\}$ is meant the indicator function of any event
$\mathcal{E}$ and the integer part of any $u\in \mathbb{R}$ is denoted
by $\lfloor u \rfloor$. For any $x \in E$, we denote by
$\theta(x) = \lVert x \rVert^{-1} x$ the angular component of $x$ for
conciseness.
\subsection{Least Squares Minimization on Extremes - The  {\sc ROXANE} Algorithm}

To help the reader follow the overall  workflow of the paper, we begin immediately by introducing the algorithm {\sc ROXANE} ({\sc R}egression \textsc{O}n e\textsc{X}treme \textsc{AN}gl\textsc{E}s) that we promote to solve the regression problem on extremes stated in the introduction, formulated as the minimization of the risk functional  $\riskinf$ defined in~\eqref{eq:asym_cond_risk}, \new{thus generalizing the binary classification framework introduced in \cite{jalalzai2018binary}}. The remainder of this work aims at developing a framework that fully justifies Algorithm~\ref{algo} below. 
\begin{algorithm}[ht]
   \caption{{\sc R}egression \textsc{O}n e\textsc{X}treme \textsc{AN}gl\textsc{E}s ({\sc ROXANE})}
   \label{algo}
\begin{algorithmic}
  \STATE {\bfseries INPUT:} Training dataset $\mathcal{D}_n=\{(X_1,Y_1),\; \ldots,\;(X_n,Y_n)\}$ with $(X_i,Y_i) \in \rset^d \times \rset$ ; %,  $i=1,\; \ldots,\; n$;
  class $\mathcal{H}$ of predictive functions $h:\mathbb{S}\to \mathbb{R}$; number $k\leq n$ of `extreme' observations among training data.
   \STATE {\bfseries Truncation:}  Sort the training data by decreasing order of magnitude of their norm, so that the sorted sample $\{ X_{(i)}, i\le n\}$ satisfies % input information
$
\lVert X_{(1)}\rVert  \geq \ldots \geq \lVert X_{(n)}\rVert
$. 
Form a training set made of $k$ \textit{extreme training observations}
\begin{equation*}
\left\{ \left(X_{(1)}, Y_{(1)}\right),\; \ldots,\;\left(X_{(k)}, Y_{(k)}\right)  \right\}.
\end{equation*}

\STATE {\bfseries Empirical quadratic risk minimization:} based on the extreme training dataset, solve the optimization problem
\begin{equation}\label{eq:emp_min_extr}
\min_{h\in \mathcal{H}}\frac{1}{k}\sum_{i=1}^k\left( Y_{(i)}-h\left(\theta\left(X_{(i)}\right)\right) \right)^2,
\end{equation}
where  $\theta(x) = x/\|x\|$ for any $x\in\rset^d\setminus\{0\}$.  %producing the

\STATE {\bfseries OUTPUT:} Solution $\hat{h}$ to problem \eqref{eq:emp_min_extr} and predictive function
$ \widehat f(x) = (\hat{h}\circ \theta)(x)$ to be used for predictions of $Y$ based on new examples $X$ such that $\lVert X \rVert\ge \lVert X_{(k)}\rVert$.
\end{algorithmic}
\end{algorithm}

%~\\

The {\sc ROXANE} algorithm can be implemented with any optimization
heuristic solving the quadratic risk minimization problem
\eqref{eq:emp_min_extr}, refer to e.g., \cite{GKKW02}. The study of
dedicated numerical techniques is beyond the scope of the present
paper.

A key feature of the {\sc ROXANE} Algorithm is that its training step
involves the \emph{angular} component of extremes solely. It returns a
prediction function $\widehat f$ which only depends on the angular
component $\theta(X)$ of a new input $X$. This apparently arbitrary
choice turns out to be fully justified under regular variation
assumptions, which are introduced and discussed in the following
subsections. To wit, the main theoretical advantage of considering
angular prediction function is to ensure the convergence of the
conditional risk $R_t$, as $t\to+\infty$.  In practice, rescaling all
extremes (in the training set and in new examples) onto a bounded set
allows a drastic increase in the density of available training
examples and a clear extrapolation method beyond the envelope of
observed examples.

After recalling some minimal background about multivariate regular
variation (Section~\ref{subsec:ht}), we introduce in
Section~\ref{subsec:framework} a modified version of the standard
regularly varying framework (\emph{regular variation with respect to
  the first component}) which is suitable for the regression problem
considered here, in the sense that the {\sc ROXANE} Algorithm turns
out to enjoy probabilistic and statistical guarantees in this context.
We thoroughly discuss the relevance of our assumptions by working out
several sufficient conditions and examples.  We state our main
probabilistic results in Section~\ref{subsec:approach}, establishing
connections between different risks and their corresponding
minimizers, thus bringing a first (probabilistic) justification
regarding the angular nature of the prediction function in
Algorithm~\ref{algo}. Statistical guarantees are deferred to
Section~\ref{sec:stats}.

\subsection{Background on Multivariate Regular Variation}\label{subsec:ht}

The goal of heavy-tail analysis is to study phenomena that are not
ruled by averaging effects but determined by extreme values. To
investigate the behavior of a random vector $X$ valued in $E$, far
from its center of mass, a classic assumption is that $X$'s
distribution is \textit{multivariate regularly varying} with tail
index $\alpha>0$, \ie there exist a nonzero Borel measure $\nu$ on
$E$, finite on all Borel measurable subsets of $E$ bounded away from
zero and a \textit{regularly varying} function $b(t)$ with index
$\alpha$, \ie $b(tx)/b(t) \rightarrow x^\alpha$ as
$t \rightarrow +\infty$, such that
\begin{equation}\label{eq:multivariate}
b(t)\mathbb{P}\left\{ X\in  t A \right\}\rightarrow \nu(A) \text{ as } t\rightarrow +\infty,
\end{equation}
for any Borel measurable set $A\subset E$ bounded away from zero
($0\notin \partial A$) and such that $\nu(\partial A)=0$.  The latter
convergence is referred to as vague convergence in
$[-\infty,+\infty]^d\setminus{\{0_{\rset^d}\}}$ (see
\cite{resnick2013extreme}, Section~3.4), or equivalently as
$\mathbb{M}_0$-convergence in $E$ (see
\cite{hult2006regular,lindskog2014regularly}). The \emph{limit
  measure} $\nu$ is provably homogeneous of degree $-\alpha$:
$\nu(tA)=t^{-\alpha} \nu(A)$ for all $t>0$ and Borel set $A\subset E$
bounded away from the origin. One may refer to
\cite{resnick2013extreme} for alternative
formulations/characterizations of the regular variation property and
its application to MEVT. It follows from the homogeneity property that
the pushforward measure of $\nu$ by the polar coordinates
transformation $x\in E\mapsto (\lVert x\rVert, \theta(x))$ is the
tensor product given by
\begin{equation*}
\nu \left\{ x\in E:\; \lVert x\rVert \geq r,\;   \theta(x)\in B  \right\} =r^{-\alpha}\Phi(A),
\end{equation*}
for all $B\in \mathcal{B}(\mathbb{S})$ and $r \s 1$, where $\Phi$ is a
finite positive measure on $\mathbb{S}$, referred to as the
\textit{angular measure} of the limit measure $\nu$.  The regular
variation assumption~\eqref{eq:multivariate} implies that the
conditional distribution of $(\lVert X\lVert/t,\theta(X))$ given
$\lVert X\rVert\s t$ converges as $t \to + \infty$: for all $r \s 1$
and $B \in \mathcal{B}(\mb{S})$ with $\Phi(\partial B) = 0$, we have
\begin{equation*}\label{eq:multivariate_bis}
\mb{P} \Big\{ t^{-1}\lVert X\rVert \ge r, \theta(X) \in B \mid \lVert X \rVert \ge t\Big\} \tti cr^{-\alpha} \Phi(B),
\end{equation*}
where $c= \Phi(\mb{S})^{-1} = \nu(E\setminus\B)^{-1}$ Hence, the
radial and angular components of the random variable $X$ are
asymptotically independent with standard Pareto distribution of
parameter $\alpha$ and normalized angular measure $c\Phi$ as
respective asymptotic marginal distributions. The angular measure
$\Phi$ describes exhaustively the dependence structure of the
components $X^{(j)}$'s given that $\lVert X \rVert$ is large, \ie the
directions $\theta(X)$ in which extremes occur with largest
probability.

Heavy-tailed models have been the subject of much attention in the
statistical machine-learning literature. Among many other works, the
regular variation assumption is used in \cite{ohannessian2012rare} for
rare event probability estimation, in \cite{Achab_2017} or
\cite{carpentier2014extreme} in the context of stochastic bandit
problems, in \cite{goix2015learning} for the statistical recovery of
the dependence structure in the extremes, in \cite{goix2017sparse} for
dimensionality reduction in extreme regions and in
\cite{brownlees2015empirical} for predictive problems with
heavy-tailed losses.

\subsection{Regular Variation with respect to the First Component}\label{subsec:framework}
We now describe rigorously the framework we consider for regression in
extreme regions, which may be seen as a natural, `one-component'
extension of standard multivariate regular variation assumptions
recalled in Section~\ref{subsec:ht}.

For simplicity, we suppose that $Y$ is bounded through this
paper. This assumption can be naturally relaxed at the price of
additional technicalities (\ie tail decay hypotheses). A more detailed
discussion of the relevance of our hypotheses to the literature on
statistical learning and EVA is given in
Remark~\ref{rem:HToutput-input}.
\begin{assumption}\label{hyp:bound}
  The random variable $Y$ is bounded: there exists $M \in(0,+\infty)$
  such that with probability one, $Y\in I =
  [-M,M]$. 
\end{assumption}

The following hypothesis concerns the asymptotics, as
$t\rightarrow +\infty$, of the conditional distribution of the pair
$(X,Y)$ given that $\lVert X \rVert >t$. It may be viewed as
one-component extension of the classic regular variation assumption
~\eqref{eq:multivariate}.
\begin{assumption}\label{hyp:limit}
There exists a non null Borel measure $\mu$ on $\mathbb{O} = E\times I$, which is finite on sets bounded away from $\mathbb{C} = \{0\}\times I$,  and a regularly varying function $b(t)$ with index $\alpha>0$ such that
\begin{equation}\label{eq:cond_reg_var}
  \lim_{t\rightarrow +\infty}
  b(t) \mathbb{P}\left\{ t^{-1}X\in A, Y \in C \right\}
  =\mu( A \times C),
\end{equation}
for all $A \in \mathcal{B}(E)$ bounded away from zero and $C\in \mathcal{B}(I)$  such that $\mu(\partial (A \times C))=0$.
\end{assumption}

Assumption~\ref{hyp:limit} could be understood as a multivariate
extension of the \emph{One-Component Regular Variation} framework
developed in~\cite{hitz2016one} \new{or of the "partial regular
  variation" (with scaling function $c(t) \equiv 1$) described in
  Chapter~3.2 of \cite{kulik2020heavy}}. It fits into the framework of
Regular Variation in $\mathbb{M}_{\mathbb{O}}$ developed
in~\cite{lindskog2014regularly} as an extension
of~\cite{hult2006regular}, where
$\mathbb{O} = E\times I = (\rset^d\times I )\setminus ( \{0\}\times
I)$ and where the scalar multiplication is defined as
$\lambda(x,y) = (\lambda x,y)$. More details regarding the connections
between Assumption~\ref{hyp:limit} and~\cite{lindskog2014regularly}
are provided in
Appendix~\ref{sec:cond_mult_reg_var}. 

\begin{remark}[On Assumptions~\ref{hyp:bound} and~\ref{hyp:limit}:
  Heavy-tailed input or output?] \label{rem:HToutput-input} A classic
  way of relaxing Assumption~\ref{hyp:bound} is to assume that the
  cost function (or $Y$ itself in the case of least squares
  regression) has subgaussian tails, as developed \emph{e.g.}
  in~\cite{LM16}. It is even possible to consider heavy-tailed losses
  (or noises) at the price of additional `small ball' conditions on
  the class of predictors \citep{Mendelson2018}, and substantially
  more technicality. We do not pursue this idea further in this work,
  because our primary focus is on extreme covariates, not on extreme
  targets, which is \emph{not} contradictory to
  Assumption~\ref{hyp:bound}. Our goal is to address a problem which
  can be viewed as one of 'out-of-domain generalization', rather than
  a regression problem involving an unbounded or heavy-tailed noise
  (or loss).  Indeed, regarding Assumption~\ref{hyp:limit}, attention
  should be paid to the fact that the heavy-tails (\emph{i.e.} regular
  variation) assumption is here on the distribution of the input
  random variable $X$, in contrast to other works devoted to
  regression such as \cite{brownlees2015empirical} or
  \cite{lugosi2016risk} where it is the loss/response that is
  supposedly heavy-tailed.

  In the supervised EVA literature, similarly, the vast majority of
  existing works in a regression context are concerned with extreme
  values of the target, typically in extreme quantiles
  regression~\citep{chavez2014extreme,daouia2023optimal,el2012estimation}.
  Recent examples considering high dimensional settings and supervised
  dimension reduction include
  \cite{aghbalou2021tail,bousebata2023extreme,gardes2018tail,girard2024functional},
  LASSO-type high-dimensional regression \citep{de2022extreme}, and
  Machine Learning methodology such as gradient boosting
  \citep{velthoen2023gradient} or random forests
  \citep{gnecco2022extremal}.

  We show in Example~\ref{ex:predictionRVvect} that the present
  framework and in particular the boundedness assumption is indeed
  relevant in some classic multivariate EVA problems, where the goal
  would be to predict the \emph{relative} contribution of a given
  component of a heavy-tailed random vector.

\end{remark}

\begin{remark}[Pre-Processing]\label{rk:emp_std}
  Because the goal of this paper is to explain main ideas to tackle
  the problem of regression on extremes, the input are assumed to be
  regularly varying with same marginal index while in practice, this
  condition may be satisfied only after some marginal standardization.
  This is a recurrent theme in multivariate extreme value theory.  For
  binary-valued $Y$, in the classification setting,
  \cite{clemencon+j+s+s:2021} consider a marginal standardization
  based on ranks, following~\cite{Einmahl2001,Einmahl2009}. They prove
  an upper bound on the statistical error term induced by this
  transformation which is of the same order of magnitude as the error
  when marginal distributions are known, a simplified case considered
  in~\cite{jalalzai2018binary}.  In our experiments with real data,
  this pre-processing step is not necessary.  We leave this technical
  and potentially difficult question outside the scope of this paper.
\end{remark}

In the sequel we refer to the limit measure $\mu$ as the \textit{joint
  limit measure} of $(X,Y)$.  Under Assumption \ref{hyp:limit}, $X$'s
marginal distribution is regularly varying with \emph{marginal limit
  measure}
\begin{equation*}
  \mu_X( A ) = \lim_{t \rightarrow +\infty} b(t)\PP[X\in t A]
  = \lim_{t \rightarrow +\infty} b(t)\PP[ X\in tA, Y\in I]= \mu(A \times I),
\end{equation*}
with $A\in \mathcal{B}(E)$ bounded away from zero and such that
$\mu(\partial (A \times I))=0$.  We also naturally introduce the 
\textit{joint angular measure} of $(X,Y)$ 
denoted by $\Phi$, which is a finite measure on $\sphere\times I$ given by
\begin{equation}
  \label{eq:definePhiJoint}
  \Phi(B\times C) = \mu \{(x,y)\in E\times I: \|x\|\ge 1, \theta(x)\in B, y\in C\}. 
\end{equation}
With this notation, under Assumption~\ref{hyp:limit} it holds that
\begin{equation}\label{eq:cond_reg_var3}
  \frac{\PP\{\theta(X) \in B, \,Y \in C,\, \lVert X\rVert  \s t r \}}{
    \mb{P}\{\lVert X\rVert \s t \}} \tti \, c\, r^{-\alpha} \Phi(B \times C),
\end{equation}
where
$c = \Phi(\mb{S}\times I)^{-1}=\mu((E\setminus\B) \times I)^{-1}$, for
all $ C \in \mathcal{B}(I)$, $B \in \mathcal{B}(\mb{S})$, such that
$\Phi(\partial (B\times A))=0$ and $r \s 1$. The latter statement is
proved in Appendix~\ref{sec:cond_mult_reg_var},
Theorem~\ref{thm:limit}.  To lighten the notation, we assume without
loss of generality that $b$ is chosen so that
$\mu((E\setminus\B) \times I) = 1$ and thus $c=1$ and $\Phi$ is a
probability measure on $\sphere \times I$. In particular, the joint
limit measure $\mu$ and the joint angular measure $\Phi$ are linked
through the relation
\begin{equation*}\label{eq:rel_reg_var}
\mu(\{ x \in E: \;  \lVert x \rVert \s r , \theta(x) \in B \}\times C) = r^{-\alpha} \Phi(B \times C)
\end{equation*}
for all $C \in \mathcal{B}(I), B \in \mathcal{B}(\mb{S})$ and $r>0$.
Observe that
\begin{equation*}
  \lim_{t\rightarrow +\infty}\frac{\mathbb{P}
    \left\{ \theta(X)\in B,Y \in C, \lVert X \rVert \geq t \right\} }
  {\PP\left\{ \lVert X \rVert \s t \right\}}=\Phi(B \times C),  
\end{equation*}
for all $B \in \mathcal{B}(\mb{S}), C \in \mathcal{B}(I), $ such that
$\Phi( \partial (B \times C) )=0$. In words, $\Phi$ is the asymptotic
joint probability distribution of $(\theta(X),Y)$ given that
$\lVert X \rVert \s t$ as $t \to + \infty$.

Let $P_{\infty}$ denote the limit conditional distribution on
$E\setminus\B\times I$ of the pair $(X/t,Y)$ given that $\|X\|\s t$,
\ie
\begin{equation}\label{eq:Pinf}
  P_\infty (A\times C)
  = \lim_{t\to+\infty}\PP[ X/t \in A\, , Y \in C\, \mid \lVert X \rVert \s t]  
\end{equation}
for all $A \in \mathcal{B}(E\setminus\B)$ and $C\in \mathcal{B}(I)$
such that $\mu(\partial (A \times C))=0$, and let
$(X_\infty,Y_\infty)$ denote a random pair with distribution
$P_\infty$.  It follows immediately from~\eqref{eq:cond_reg_var3} and
from our choice $c=1$, that $P_\infty$ indeed exists and is determined
by $(\Phi,\alpha)$, namely
\begin{equation*}%\label{eq:Pinf}
  \begin{aligned}
&  P_\infty\{(x,y): \lVert x \rVert >r, \theta(x)\in B, y\in C \}  \\
&  = \lim_{t\to+\infty} \PP[\lVert X \rVert / t \s r, \theta(X) \in B, Y\in C \mid \lVert X \rVert \s t] \\
&  = r^{-\alpha}\Phi(B\times C),       
  \end{aligned}
\end{equation*}
where $B,C,r$ are as in Equation~\eqref{eq:cond_reg_var3}. In other
words, if $T$ denotes the pseudo-polar transformation with respect to
the first component $T(x,y) = ( \lVert x \rVert, \theta (x), y)$ on
$E\setminus\B\times I$, and if $\nu_\alpha$ is the Pareto measure
$\nu_\alpha([x,\infty)) = x^{-\alpha} $, then the following tensor
product decomposition holds true in polar coordinates, \ie
$P_\infty\circ T^{-1} = \nu_\alpha \otimes \Phi$.

Observe that, under Assumptions~\ref{hyp:bound} and~\ref{hyp:limit},
the random variable $Y_{\infty}$ is almost-surely bounded in amplitude
by $M<+\infty$.

Equipped with these notations, it is natural to consider the squared
error loss of a prediction function $f$, under the distribution
$P_\infty$. We call this key quantity the \emph{extreme quadratic
  risk}, denoted by $\riskext$, defined as 
\begin{equation*}\label{eq:extreme_risk}
    \riskext(f):=\mathbb{E}\left[ \left( Y_{\infty}-f(X_{\infty}) \right)^2 \right],
\end{equation*}
for $f \in \mathcal{F}$ a class of real-valued bounded
Borel-measurable functions defined on $E\setminus\B$.  As will become
clear in the subsequent analysis, although our objective $\riskinf$
and the extreme risk $\riskext$ are two different functionals, they
turn out to be connected through their minimizers under an additional
technical assumption stated below.  In the sequel we let $\bayesext$
denote the minimizer pf $\riskext$ among all measurable
functions. Standard arguments from statistical learning theory show
immediately that $\bayesext$ is defined (up to a negligible set) by a
conditional expectation,
$\bayesext(X_{\infty})=\mathbb{E}[Y_{\infty} \mid X_{\infty}]$.

An additional technical regularity assumption is necessary to obtain the main results of this section, stated next and discussed below. 
\begin{assumption}\label{hyp:cont_reg_func}
  
  The extreme regression function $\bayesext$ is continuous on
  $\rset^d\setminus\{0_{\mb{R}^d}\}$ and as $t$ tends to infinity,
  \begin{equation*}
    % \label{eq:L1convRegfun}
    \EE[ |f^*(X) - \bayesext(X)| \;\big|\, \lVert X\rVert\s t] \to 0.
  \end{equation*}
\end{assumption}

Although Assumption~\ref{hyp:cont_reg_func} may seem difficult to
verify in practice, the next proposition supports its
soundness. Indeed we show that it is automatically satisfied as soon
as Assumptions~\ref{hyp:bound} and \ref{hyp:limit} hold true, under
mild additional regularity conditions regarding the uniform
convergence of regular varying densities towards limit
densities. These additional regularity are standard in the EVA
literature. More precisely, Condition (iii) in
Proposition~\ref{prop:sufficientConditionsForAssum3} below is a
`one-component variant' of standard assumptions regarding regular
variations of densities (\cite{cai2011estimation,de1987regular}),
further discussed in Example~\ref{ex:predictionRVvect} below.

\begin{proposition}[Sufficient conditions for Assumption~\ref{hyp:cont_reg_func}]\label{prop:sufficientConditionsForAssum3}
  Let $(X,Y)$ satisfy Assumptions~\ref{hyp:bound}
  and~\ref{hyp:limit}. Then Assumption~\ref{hyp:cont_reg_func} also
  holds if one of the three conditions (i), (ii), (iii) below holds
  \begin{enumerate}
  \item[(i)] \label{suff1} The regression function $f^*$ is continuous  on $\{x\in \rset^d: \lVert x \rVert \ge 1\}$ and  as $t \rightarrow +\infty$, 
\begin{equation}\label{eq:uniform_conv}
  \sup_{\lVert x \rVert\s t} | f^*(x) - \bayesext(x) | \to 0;
\end{equation}

\item[(ii)] The conditional distributions of $Y$ given $X=x$
  (\emph{resp.} $Y_\infty$ given $X_\infty = x$) admit densities
  $p_{Y|x}(y)$ (\emph{resp.} $p_{Y|x}^{\infty}(y)$) w.r.t. the
  Lebesgue measure on $I$, for all $x\neq 0$. In addition for all
  $y \in I$, the mapping $x \mapsto p_{Y|x}(y)$ (\emph{resp.}
  $x\mapsto p_{Y|x}(y)$) is continuous, and
  $\sup_{\lVert x \rVert \s 1, y \in I} p_{Y|x}(y)<+\infty$.  Finally
  the following uniform convergence holds true,
  \begin{equation}
    \label{eq:cvCondDensity}
    \sup_{\lVert x\rVert\ge t, y\in I} |p_{Y|x}(y) - p_{Y|x}^\infty(y) | \tti 0;
  \end{equation}

\item[(iii)] The random pair $(X,Y)$ (resp. $(X_\infty,Y_\infty)$) has
  a continuous density $p$ (resp. $q$) w.r.t. the Lebesgue measure,
  and the densities converge uniformly, in the sense that
\begin{equation} \label{eq:densities}
\sup_{(\omega,y) \in \mb{S}\times I} \big|\,  b(t) t^d p(t\omega,y) - q(\omega,y) \, \big| \tti 0,
\end{equation}
where $b(t) = \PP[\lVert X\rVert\s t]^{-1}$.  In addition, $q$ is
uniformly lower bounded on the unit sphere by a positive constant,
  \begin{equation}\label{eq:diff_zero}
  \inf_{\omega\in\sphere, y\in I} q(\omega, y)>0.
  \end{equation}
  \end{enumerate}
\end{proposition}

\begin{proof} 
  We show that if Assumptions \ref{hyp:bound} and~\ref{hyp:limit} both
  hold true, then each condition (i), (ii), or (iii) of the statement
  imply Assumption~\ref{hyp:cont_reg_func}. In fact we show that
  (iii)$\Rightarrow$ (ii) $\Rightarrow$ (i) $\Rightarrow$
  Assumption~\ref{hyp:cont_reg_func}.

  \noindent {\bfseries Condition (i) $\Rightarrow$
    Assumption~\ref{hyp:cont_reg_func}.}  The continuity of
  $\bayesext$ follows from the continuity of $f^*$ and the uniform
  convergence~\eqref{eq:uniform_conv}. Also, the convergence in
  Assumption~\ref{hyp:cont_reg_func} is a direct consequence of
  convergence~\eqref{eq:uniform_conv}.

\noindent {\bfseries Condition (ii) $\Rightarrow$ Condition (i).} 
For $x \in \mb{R}^d$ such that $\lVert x\rVert \s t \ge 1$, we have
\begin{equation*}
  \begin{aligned}
|f^*(x) - \bayesext(x)|  ~=~ &\Big|\int_{y \in I} y p_{Y|x}(y)dy - \int_{y \in I} y p^\infty_{Y|x}(y)dy \Big| \\
  &  \m ~ M^2 \sup_{\lVert x\rVert  \s t,y \in I}|p_{Y|x}(y) - p^\infty_{Y|x}(y)|.
    \end{aligned}
\end{equation*}
Thus, uniform convergence in~\eqref{eq:uniform_conv} follows
from~\eqref{eq:cvCondDensity}. The continuity of $f^*$ is ensured by
an application of the dominated convergence theorem to the parametric
integral $f^*(x)= \int_{I} y p_{Y|x}(y) \ud y$, using the fact that
for all $y \in I$, $x \mapsto p_{Y|x}(y)$ is continuous and that
$\sup_{\lVert x\rVert \s 1, y \in I}p_{Y|x}(y)<+\infty$.

\noindent {\bfseries Condition (iii) $\Rightarrow$ Condition (ii).} 
We first show that uniform convergence~\eqref{eq:cvCondDensity} holds
true. The density $q$ of $\mu$ is necessarily homogeneous in its first
component, $q(tx, y) = t^{-\alpha-d} q(x,y)$ for $x\neq 0$. This
follows from the homogeneity of $\mu$ and a change of variable in the
first component when integrating over a region $tA\times B$ where
$A\subset \rset^d\setminus\{0\} $ and $B\subset I$. Thus for
$x \in \mb{R}^d$ with $\lVert x\rVert \ge 1$ and $y \in I$, we have
\begin{equation*}
  p_{Y|x}(y) = \frac{p(x,y)}{p_X(x)} \quad \mbox{and} \quad p^\infty_{Y|x}(y)
  =\frac{q(x,y)}{q_X(x)} = \frac{q(x/\lVert x\rVert ,y)}{q_X(x/\lVert x\rVert )}, 
\end{equation*}
where we denote by $p_X$ (resp. $q_X$) the marginal density of $X$
(resp. $X_\infty$) given by $p_X(x) = \int_I p(x,y) dy$ (resp.
$q_X(x) = \int_I q(x,y) dy$).  Then, for
$x \in \rset^d\setminus\{0\}, y\in I$, introducing the function
$h(t) = t^d b(t)$, the left-hand side in
Equation~\eqref{eq:cvCondDensity} writes as
\begin{multline}
  \Big|\frac{p(x,y)}{p_X(x)}-\frac{q(x/\lVert x\rVert ,y)}{q_X(x/\lVert x\rVert )}\Big| =  \bigg|\frac{h(\lVert x\rVert ) p(x,y)}{h(\lVert x\rVert ) p_X(x)} - \frac{q(x/\lVert x\rVert ,y)}{q_X(x/\lVert x\rVert )} \bigg|\\ 
  \m \underbrace{h(\lVert x\rVert ) p(x,y) \bigg|\frac{1}{h(\lVert x\rVert ) p_X(x)} - \frac{1}{q_X(x/\lVert x\rVert )} \bigg|}_{A(x,y)}  \\\quad  +~\underbrace{\frac{ \Big|h(\lVert x\rVert ) p(x,y)- q(x/\lVert x\rVert ,y) \Big|}{q_X(x/\lVert x\rVert )} }_{B(x,y)}. \label{eq:decompos-conditional}
\end{multline}

Regarding the numerator of the term $B(x,y)$ above,  for $\lVert x\rVert  \ge  t$, 
\begin{align*}
|h(\lVert x\rVert ) p(x,y)- q(x/\lVert x\rVert ,y)| &= |h(t(\lVert x\rVert /t)) p(t(\lVert x\rVert /t)(x/\lVert x\rVert ),y)- q(x/\lVert x\rVert ,y)| \\
& \m \sup_{ s \s t,(\omega,y) \in \mb{S}\times I} | h(s) p(s\omega,y) - q(\omega,y) | \rightarrow 0,
\end{align*}
as $t$ tends to infinity,  by uniform convergence~\eqref{eq:densities}.

This, together  with the lower bound~\eqref{eq:diff_zero} on $q$,  implies that as $t\to+\infty$, 
\begin{equation*}
  \sup_{\lVert x\rVert >t, y\in I} B(x,y) \to 0. 
\end{equation*}

Turning to the term $A(x,y)$ in~\eqref{eq:decompos-conditional},  we have 
\begin{equation*}
A(x,y) =  h(\lVert x\rVert ) p(x,y) \bigg|\frac{h(\lVert x\rVert )p_X(x)-q_X(x/\lVert x\rVert )}{h(\lVert x\rVert )p_X(x)q_X(x/\lVert x\rVert )}\bigg|.
\end{equation*}
Also, for $\lVert x\rVert >t$, 
\begin{align}
|h(\lVert x\rVert )p_X(x) - q_X(x/\lVert x\rVert )| &= \Big| \int_I ( h(\lVert x\rVert )p(x,y) - q(x/\lVert x\rVert ,y) ) dy \Big| \nonumber \\
&\le 2M \sup_{s \s t, (\omega,y) \in \mb{S}\times I} | h(s) p(s\omega,y) - q(\omega,y) | := U(t) \label{eq:unifcvmarginal1}, 
\end{align}
where the upper bound $U(t)$ vanishes as $t\to+\infty$ because of~\eqref{eq:densities}. 
Now, for $\lVert x\rVert >t$ and $y\in I$, 
\begin{equation*}
  A(x,y) \le  \frac{ \sup_{\lVert x\rVert \ge t, y\in I}h(\lVert x\rVert ) p(x,y) }{
    \inf_{\lVert x\rVert >t}h(\lVert x\rVert )p_X(x)  \inf_{\omega \in \sphere }q_X(\omega)} U(t).
\end{equation*}
Regarding the numerator of the above display, recall that the density
function $q$ is continuous on the compact set $\sphere$, whence it is
upper bounded. Because of uniform convergence~\eqref{eq:densities}, it
is also true that %for $t$ large enough,
$\sup_{\lVert x\rVert \ge t, y\in I}h(\lVert x\rVert ) p(x,y)$ is
upper bounded by a finite constant for $t$ large enough.  In addition,
our lower bound assumption~\eqref{eq:diff_zero} on $q$ together with
uniform convergence~\eqref{eq:unifcvmarginal1} show that the
denominator is ultimately (as $t\to+\infty$) lower bounded by a
positive constant. Summarizing, we have shown that
$\sup_{\lVert x\rVert >t, y\in \sphere} A(x,y) \to 0$ as $t\to\infty$,
finishing the proof of~\eqref{eq:cvCondDensity}.

It remains to prove that for all $y \in I$, the function
$x \mapsto p(x,y)/p_X(x)$ is continuous and that $p(x,y)/p_X(x)$ is
uniformly bounded.  For all $y \in I$, the continuity of
$x \mapsto p(x,y)/p_X(x)$ follows from the continuity of $p$. Also,
for $x \in \rset^d$ and $y\in I$, we have
\begin{equation*}
  \frac{p(x,y)}{p_X(x)} = \frac{h(\lVert x\rVert )p(x,y)}{h(\lVert x\rVert )p_X(x)}.
\end{equation*}
The numerator uniformly converges to $q$, which is uniformly
bounded. The denominator uniformly converges to $q_X$, which is
uniformly lower bounded by Equation~\eqref{eq:diff_zero}. Then
$\sup_{\lVert x\rVert \s 1, y \in I}(p(x,y)/p_X(x))$ is finite, which
concludes the proof. 
\end{proof}

We now work out several examples of regression settings in which our
Assumptions~\ref{hyp:bound}, \ref{hyp:limit}
and~\ref{hyp:cont_reg_func} are satisfied.

\begin{example}[Noise model with heavy-tailed random design]\label{ex:noise}
  Suppose that $X$ is a regularly varying random vector in $\rset^d$,
  independent from a real-valued random variable $\varepsilon$
  modeling some noise and consider a target
\begin{equation*} \label{eq:example}
Y=g(X,\varepsilon),
\end{equation*}
where $g: \mb{R}^d \times \mb{R} \rightarrow \mb{R}$ is a bounded,
continuous mapping. Assume also that there exists a function
$g_\theta:\mathbb{S}\times \mb{R}\rightarrow \mathbb{R}$ such that,
for all $z \in \mb{R}$
\begin{equation}\label{eq:condition_example}
\sup_{\lVert x\rVert \s t }| g(x,z)- g_\theta(x/\lVert x\rVert,z)|\rightarrow 0,
\end{equation}
as $t\rightarrow +\infty$.
Then, the  random pair $(X,Y)$ fulfills Assumptions~\ref{hyp:bound}, ~\ref{hyp:limit} and \ref{hyp:cont_reg_func}.  
\end{example}
The proof of the claim made in Example~\ref{ex:noise} is deferred to
Appendix~\ref{sec:proofs_section_proba}, Section~\ref{sec:examples}.
Concrete examples arise within the broader context of this generic
example, such as the additive noise model
$Y = \Tilde{g}(X) + \varepsilon$ and the multiplicative noise model
$Y = \varepsilon \Tilde{g}(X)$. In both cases,
Condition~\eqref{eq:condition_example} holds true whenever $\Tilde g$
satisfies the similar condition
  \begin{equation*}
    \sup_{\lVert x\rVert \s t}| \Tilde{g}(x)- \Tilde{g}_\theta(\theta(x))|\rightarrow 0, 
  \end{equation*}
  for some angular function $\Tilde{g}_\theta$,   with minor additional regularity assumptions.   We work out the details of these two sub-examples in  Section~\ref{sec:examples}, Propositions~\ref{ex:add_noise} and~\ref{ex:multiplicativeNoise}, from Appendix~\ref{sec:proofs_section_proba}. 

  The next example establishes a strong connection between the
  considered regression setting and typical concrete situations
  considered in Extreme Value Analysis where the goal is to predict
  the occurrence and/or the intensity of unusually large
  events. 
  
  \begin{example}[Predicting  a missing component in a regularly varying vector]\label{ex:predictionRVvect}
    
    In this example we show that our assumptions are met when
    considering a random vector $\tilde X$ with a regularly varying
    \emph{density}, where the target $Y$ is one missing component from
    the vector, or more precisely a normalized version of that missing
    component. The normalization allows to satisfy our boundedness
    constraint Assumption~\ref{hyp:bound}. We believe this example
    could be particularly useful in applications, for imputation of
    missing data with heavy tails. It should be noted that such a
    problem is the main motivation
    behind~\cite{cooley2012approximating}, whose aim is to estimate
    the full conditional distribution of the missing component given
    the observed ones. \new{The present example was initially
      developed in an earlier arXiv version of this work and has since
      been adapted in Example 2.1 of \cite{aghbalou2022cross} to a
      simpler situation where the goal is to predict an exceedance
      over a high threshold by the missing component, given that the
      other components are unusually high.}

    Let $\tilde X\in\rset^{d+1}$ have continuous density $p$
    and % let b regularly varying with tail index $\alpha$. Let
    $b(t)=1/\PP[\lVert \tilde X\rVert\s t]$, where
    $\lVert \,\cdot\,\rVert$ is the $L^p$ norm on $\rset^{d+1}$ for
    some $p\in[1,+\infty)$ .  Assume that $b$ is regularly varying
    with index $\alpha$ for some $\alpha>0$, and that there exists a
    positive function $q$ on $\rset^{d+1}$ such that for all
    $\tilde x\neq 0_{\rset^{d+1}}$,
  \begin{equation}
    \label{eq:pointwiseCVDensity}
   t^{d+1}b(t)   p(t\tilde x)  - q(\tilde x) \tti 0. 
  \end{equation}
  Assume in addition that the convergence is uniform on the sphere,
  % form of regular variation of the density $p$ of $X$,
  \begin{equation}
    \label{eq:cvDensity}
    \sup_{\omega \in\sphere_{d+1}} %\frac{
      | t^{d+1}b(t) p(t\omega)   - q(\omega) | %}{q(x)}
      \tti 0, 
    \end{equation} 
    where $\sphere_{d+1}$ denotes the unit sphere of $\rset^{d+1}$.
    This assumption is used in
    \cite{cai2011estimation,de1987regular}. It is shown in these
    references that~\eqref{eq:pointwiseCVDensity}
    and~(\ref{eq:cvDensity}) imply that $\tilde X$ is regularly
    varying with index $\alpha$.  More precisely with
    $\mu(A) = \int_A q(\tilde x)\ud \tilde x$ for any measurable set
    $A\subset E$, we have
    $b(t) \PP[\tilde X/t \in \,\cdot\, ] \to \mu(\,\cdot\,)$ in the
    sense of vague convergence. Necessarily $q$ is homogeneous of
    order $-\alpha
    -d-1$.  % necessarily regularly varying with tail index $-\alpha-d-1$.
    Also the continuity of $p$ implies that of $q$. Assume finally
    that $\min_{\omega\in\sphere_{d+1}} q(\omega)>0$.  Another useful
    feature of this setting is that, if \eqref{eq:pointwiseCVDensity}
    and (\ref{eq:cvDensity}) hold, then also
  \begin{equation}
    \label{eq:cvDensitybis}
    \sup_{\lVert \tilde x\rVert \ge 1} %\frac{
      | p(t \tilde x) t^{d+1}b(t)  - q(\tilde x) | %}{q(x)}
      \tti 0. 
    \end{equation}

    Let $ X = (\tilde X_1,\ldots, \tilde X_d)$ and
    $Y =\tilde X_{d+1}/\lVert \tilde X\rVert$. The norm
    $\lVert x\rVert$ also denotes the $L^p$ norm in $\rset^d$ when it
    is clear from the context that $x\in\rset^d$.  It is important to
    observe that predicting $Y$ allows to predict $\tilde X_{d+1}$, as
\begin{equation*}
  Y =\frac{\tilde X_{d+1}}{\lVert\tilde X\rVert_p} \quad \Longleftrightarrow \quad 
  \tilde X_{d+1} = \frac{Y\| X\|_p}{(1-|Y|^p)^{1/p}}.
\end{equation*}
In our experiments with real data we consider the present prediction
example on a financial dataset. Importantly,
Proposition~\ref{prop:exampleMissingComponent} below shows that the
transformed pair $(X,Y)$ obtained by the transformations described
above satisfies our required assumptions, and also gives an explicit
expression for the limit pair $(X_\infty,Y_\infty)$ in this setting.
 \end{example}

\begin{proposition}\label{prop:exampleMissingComponent}
  Let $\tilde X\in\rset^{d+1}$ be a regularly varying random vector as
  in Example~\ref{ex:predictionRVvect}, namely, assume that $\tilde X$
  has regularly varying density $p$ satisfying~\eqref{eq:cvDensitybis}
  where $b(t)=\PP[\lVert \tilde X\rVert\ge t]$ and $q$ is uniformly
  lower bounded on the unit sphere,
  $\inf_{\omega\in\sphere_{d+1}} q(\omega)>0$. Let
  $ X = (\tilde X_1,\ldots, \tilde X_d)$ and
  $Y =\tilde X_{d+1}/\lVert \tilde X\rVert$. Then the following
  assertions hold true.
 \begin{itemize}
 \item[(i)] The pair $(X,Y)$ satisfies Assumptions~\ref{hyp:bound},~\ref{hyp:limit} and~\ref{hyp:cont_reg_func}; 
 \item[(ii)] The limit pair $(X_\infty,Y_\infty)$ for $(X,Y)$ defined
   in~\eqref{eq:Pinf} has distribution
\begin{equation*}
  \mathcal{L}\Big(\big( \tilde X_{\infty,1:d},  \frac{\tilde X_{\infty, d+1}}{ \lVert\tilde X_{\infty} \rVert} \big) \,\big|\, \lVert\tilde X_{\infty, 1:d}\rVert\ge 1 \Big), 
\end{equation*}
where $\tilde X_{\infty,1:d}$ denotes the $d$-dimensional vector
$(\tilde X_{\infty,1}, \ldots, \tilde X_{\infty,d})$.
\end{itemize} 
\end{proposition}
\begin{proof} 
Let $\tilde E= \rset^{d+1}\setminus\{0_{\rset^{d+1}}\}$, $E= \rset^d\setminus\{0_{\rset^d}\}$, 
 and for simplicity let us denote  by $\B_d$ both the $d$-dimensional unit ball  and its image by the canonical embedding $\rset^d\to \rset^{d+1}$, \ie
$\B_d = \{\tilde x\in\rset^{d+1}:  \lVert (\tilde x_1,\ldots, \tilde x_d)\rVert  \le 1, \tilde x_{d+1}\in\rset \}$.  
For $\tilde x\in\rset^{d+1}$ we denote by $x$ the first $d$ coordinates of $\tilde x$, $x = (\tilde x_1,\ldots, \tilde x_d)$.   Denote by $\varphi$ the continuous mapping sending $\tilde X$ to $(X,Y)$, \ie
   \begin{align*}
     \varphi: \quad  E\times \rset
     & \to E \times (-1,1)  \\
     \tilde x=(x,z)
     & \mapsto  (x, y) = (x, z/\lVert (x,z )\rVert ). 
   \end{align*}
   Equipped with these notations, we may proceed with the proof.\\~
   
\noindent
{\bfseries (a)} Assumption~\ref{hyp:bound}  is trivially satisfied because $|Y|\le 1$.\\~

\noindent {\bfseries (b)} We now show that Assumption~\ref{hyp:limit}
holds with limit pair $(X_\infty, Y_\infty)$ as in the second part of
the statement.  With the notations introduced above, the pair defined
in the statement may be written as
$( X_\infty, Y_\infty) = \varphi(\tilde X_\infty)$, where
$\tilde X_\infty$ is well defined by regular variation of the full
vector $\tilde X$.  We need to show that for any bounded, continuous
function $g$,
\begin{equation*}
  \EE[g( X/t, Y) \,| \,\lVert \tilde X\rVert  \s t ] \to \EE[ g\circ \varphi(\tilde X_\infty)\,|\, \lVert  \tilde X_{\infty,1:d}\rVert \ge 1 ] \text{ as } t\to\infty. 
\end{equation*}
However $( X/t, Y) = \varphi(\tilde X/t) $ and
$\lVert X\rVert \s t \Rightarrow \lVert \tilde X\rVert \s t$. Thus
\begin{align*}
  &  \EE[g( X/t, Y) \, | \, \lVert X\rVert  \ge t] \\
  & =  \frac{ \EE[ g \circ \varphi(\tilde X/t)\un\{\lVert X/t \rVert  \ge 1 \} \un\{\lVert  \tilde X/t \rVert  \ge 1 \} ]}{
    \PP[\lVert  \tilde X/t \rVert  \ge 1]} \frac{\PP[\lVert  \tilde X/t \rVert  \ge 1]}{\PP[\lVert  X/t \rVert  \ge 1]} \\
  &=  \EE[g\circ \varphi( \tilde X/t)\un\{\lVert  X/t \rVert  \ge 1 \}  \, | \, \lVert  \tilde X\rVert  \ge t]
    \frac{\PP[\lVert  \tilde X/t \rVert  \ge 1]}{\PP[\lVert   X/t \rVert  \ge 1]} \\
  & \to \EE[g\circ \varphi( \tilde X_\infty)\un\{\lVert  \tilde X_{\infty,1:d} \rVert  \ge 1 \}]
    \frac{1}{\PP[ \lVert \tilde X_{\infty, 1:d}\rVert \ge 1]}, 
\end{align*}
where the convergence of the first term in the latter expression is
obtained by approaching the (discontinuous) function
$\un\{\lVert z\rVert \ge 1\}$ by continuous ones and using the fact
that the boundary of $\B_d$ in $\rset^{d+1}$ is not a cone, whence it
cannot carry any positive $\mu$-mass (a standard feature of radially
homogeneous measures).\\~

\noindent {\bfseries (c) } We now prove that
Assumption~\ref{hyp:cont_reg_func} holds true by proving the stronger
condition~(\ref{eq:uniform_conv}) which rephrases in our setting as
\begin{equation}\label{eq:unifCvregextr_proof}
  \sup_{\lVert x\rVert =1} |f^*(t x) - \bayesext(t x) | \tti 0. 
\end{equation}
Indeed if~(\ref{eq:unifCvregextr_proof}) holds, then
$ \sup_{s \s t}\sup_{\lVert x\rVert =1} |f^*(t x) - \bayesext(t x) |
\tti 0$, so that
\begin{align*}
  \sup_{\lVert x\rVert  \s t} |f^*(x) - \bayesext(x) | & = \sup_{\lVert x\rVert  \s 1} |f^*(t x) - \bayesext(t x) |\\
  &= \sup_{s \s t} \sup_{\lVert  x\rVert  = 1} |f^*(s x) - \bayesext(s x) |\\
  &\tti 0.
\end{align*}
% \item $(ii)$

For $x \in \rset^d$ such that $\lVert x\rVert \ge 1$, $f^*(x)$ and
$\bayesext(x)$ may be written in terms of integrals
  \begin{align*}
    f^*(x) &= \int_{z\in\rset}
             \frac{z}{\lVert (x,z)\rVert }
             \frac{p(x,z)}{p(x)} \ud z, 
  \end{align*}
  where for simplicity we denote by $p(x)$ the marginal density of the
  first $d$ components of $\tilde X$ at $x$, and also by $p(x, z)$,
  the joint density at $\tilde x= (x,z)$.

  In the present setting, $\bayesext$ is defined as
  $ \bayesext(X_\infty) = \EE[Y_\infty\,\mid\, X_\infty] $. Introduce
  a random vector $\tilde Z = (\tilde Z_1,\ldots,\tilde Z_{d+1})$
  distributed as
  $\mathcal{L}(\tilde X_\infty \,|\, \lVert \tilde X_{\infty,
    1:d}\rVert \ge 1)$. Then $\tilde Z$ has density $C q(x, z)$ on
  $\B_d^c\times \rset$, and marginal density for its first $d$
  components, $C q(x) := \int_\rset Cq(x,z) \ud z$.  With these
  notations we have
  $(X_\infty, Y_\infty) \overset{d}{=} (\tilde Z_{1:d}, \tilde
  Z_{d+1}/\lVert \tilde Z_{1:d}\rVert ) $, whence
  $\bayesext(\tilde Z_{1:d}) = \EE[\tilde Z_{d+1}/\lVert\tilde Z\rVert
  \, |\, \tilde Z_{1:d}]$ almost surely.  We obtain, for
  $\lVert x\rVert \ge 1$,
\begin{equation*}
  \bayesext(x)
  = \int_\rset \frac{z}{\lVert(x, z)\rVert} \frac{ C q(x,z)}{C q(x)} \ud z
  = \int_\rset \frac{z}{\lVert( x, z)\rVert} \frac{  q(x,z)}{ q(x)} \ud z. 
\end{equation*}
Combining the  latter two displays we obtain 
\begin{equation}
  \label{eq:diffRegFun}
|f^* (x) - \bayesext(x) | \le
\int_{z\in\rset}  \left|\frac{p(x,z)}{p(x)} -
  \frac{q(x,z)}{q(x)}  \right| \ud z.   
\end{equation}

Introduce as in Lemma~\ref{lem:Unifcvmarginal} the function
$h(t)= t^{d+1}/\PP(\|\tilde X\|\s t)$.  For $\lVert x\rVert = 1$, by a
change of variable $r= z/t$ in~(\ref{eq:diffRegFun}), we obtain
 \begin{align*}
   |f^* (t x) - \bayesext(t x) |
   & \le
     \int_{r\in\rset}  \left|\frac{p(tx,tr )}{p(t x)} -
     \frac{q( t x,t r )}{q(t x)}  \right| t \ud r   \\
  & = \int_{r\in\rset} \left|\frac{ h(t) p(t  x,tr )}{ t^{-1}
     h(t)p(t  x)} -
     \frac{q(  x,  r )}{q(  x)}  \right|  \ud r,    
\end{align*}
since  by  homogeneity of $q$, it holds that $q(t x ,t r)= t^{-d-1-\alpha} q( x, r)$ while $q(t  x) = t^{-d-\alpha} q( x)$.
Thus
 \begin{align}
   \sup_{\lVert x \rVert = 1}   |f^* (t x) - \bayesext(t x) |
   & \le 
\int_{r\in\rset} \underbrace{ \sup_{\lVert  x\rVert =1} \left|\frac{ h(t) p(t x,tr )}{ t^{-1}
     h(t)p(t x)} -
     \frac{q(x,  r )}{q( x)}  \right|  }_{J(t,r)}\ud r.\label{eq:boundDifft}
 \end{align}
 We have the following controls over the quantities in the latter integrand:

1.   $q(x)$ is lower bounded by a positive constant (Lemma~\ref{lem:boundsMarginalQ})
 
2. $ \sup_{\lVert x\rVert =1} |h(t)t^{-1} p(t x) - q( x)|\tti 0$ (Lemma~\ref{lem:Unifcvmarginal}),
 
3.  For all fixed $r$,    because of (\ref{eq:cvDensitybis}), and since $\lVert ( x, r)\rVert  \ge \lVert x\rVert $,   
\begin{equation*}
  \sup_{\lVert x\rVert =1} |h(t) p(t x, tr  ) - q(x,r )|
  \le \sup_{\lVert  \tilde u\rVert \ge 1} |h(t) p(t \tilde u) - q(\tilde u )|
  \tti 0. 
\end{equation*}
Thus, combining 1., 2. and 3. above, for fixed $r$, the integrand
$J(t,r)$ in~(\ref{eq:boundDifft}) converges to $0$ as $t\to
+\infty$. In order to apply the dominated convergence theorem, we
verify that $J(t,r)$ is upper bounded by an integrable function of
$r$. The argument is somewhat similar to the one in the proof of
Lemma~\ref{lem:Unifcvmarginal}.  We decompose the integrand as
\begin{align*}
  J(t,r)
  & \le
    \underbrace{ \sup_{\lVert x\rVert  = 1} \frac{h(t)}{ h(t\lVert (x,r)\rVert )}}_{A(t,r)}
    \underbrace{\sup_{\lVert x\rVert  = 1}
    \frac{ h(t\lVert (x,r)\rVert ) p\Big( t\lVert (x, r)\rVert  \theta(x,r) \Big)
    }{t^{-1}h(t)p(t x)}}_{B(t,r)}  \\
  &+  \underbrace{ \sup_{\lVert x\rVert  = 1} \frac{q(x,  r )}{q(  x)}}_{C(t,r)} ~
    = ~A(t,r) B(t,r) + C(t,r). 
\end{align*}
   
From the proof of Lemma~\ref{lem:Unifcvmarginal} (see
Equation~(\ref{eq:boundA_Karamata})) we have that for $t\ge t_0$ large
enough, and for all $r\in\rset$,
\begin{equation*}
  \label{eq:boundA_Karamata_main}
  A(t,r) \le 2 \lVert (x,r)\rVert ^{-d-\alpha/2-1} \le 2 (1+r^p)^{\frac{-d-\alpha/2-1}{p}},   
\end{equation*}
an integrable function of $r$.

The numerator and the denominator in the definition of $B(t,r)$
converge as $t\to +\infty$, uniformly over $\lVert x\rVert \ge 1$ and
$r\in\rset$, respectively to $q(x,r)$ and $q(x)$. The latter quantity
is lower bounded (Lemma~\ref{lem:boundsMarginalQ}) and $q( x,r)$ is
uniformly bounded for $\lVert x\rVert = 1$ (by homogeneity). Thus, for
some constant $C>0$, for all $t\ge t_1$ with some large enough
$t_1\ge t_0$, we have
\begin{equation*}
  B(t,r)\le C.
\end{equation*}

By homogeneity of $q$ and Lemma~\ref{lem:boundsMarginalQ} again, we
have
\begin{align*}
  C(t,r) &\le \sup_{\lVert  x\rVert  = 1}\lVert  (x,r)\rVert ^{-\alpha-d-1} \frac{\max_{\omega \in \sphere_{d+1}} q(\omega)}{c} \\
         & = (1 + r^p)^{\frac{-\alpha-d-1}{p}} \frac{\max_{\omega\in \sphere_{d+1}} q(\omega)}{c},
\end{align*}
which is an integrable function of $r$.

Combining the bounds regarding $A(t,r), B(t,r), C(t,r)$, we have shown
that $A(t,r) B(t,r) + C(t,r)$ is upper bounded by an integrable
function of $r$. The proof of the condition~\eqref{eq:uniform_conv} is
complete.  It remains to show that $\bayesext$ is continuous on
$\lVert x\rVert \s 1$. Recall that for
$x\in \rset^d \setminus \{0_{\rset^d}\}$,
\begin{equation*}
\bayesext(x) = \frac{1}{q(x)}\int_\rset \frac{z}{\lVert (x,z)\rVert }q(x,z)dz.
\end{equation*}
The continuity of $p$ implies that of $q$ by
Equation~\eqref{eq:cvDensity}. By homogeneity of $q$, we have
\begin{align*}
  \frac{z}{\lVert (x,z)\rVert }q(x,z) \m q(x,z) & =   \lVert (x,z)\rVert ^{-d-\alpha-1}q\big(\theta(x,z)\big) \\
  &\m (1+z^p)^{\frac{-d-\alpha-1}{p}}\max_{\omega \in \sphere_{d+1}}q(\omega).
\end{align*}
Since $z \mapsto (1+z^p)^{\frac{-d-\alpha-1}{p}}$ is integrable over
$\rset$, the dominated convergence theorem for continuity applies
twice and entails that the functions
$x \mapsto \int_\rset (z/\lVert (x,z)\rVert )q(x,z)dz$ and
$x \mapsto 1/q(x)$ are continuous and then $\bayesext$ is
continuous. 
\end{proof}
 
As shown in upcoming sections,
Assumptions~\ref{hyp:bound},~\ref{hyp:limit}
and~\ref{hyp:cont_reg_func} provide sufficient regularity and
stability conditions allowing to justify the \emph{angular} ERM
approach taken in Algorithm~\ref{algo}.

\section{Regression on Extremes - Main Results}\label{sec:main}

The analysis carried out in this section aims to provide a solid theoretical foundation for the ROXANE algorithm introduced in Section~\ref{sec:background} and establish its generalization properties w.r.t. the limit distribution $P_{\infty}$. Several steps are required in this purpose. Subsection~\ref{subsec:approach} deals with the performance criteria related to the conditional distributions and the limit distribution, and their minimizers as well. It shows that a solution of the regression problem in the limit regime can be asymptotically recovered by solving the regression problem in a preasymptotic regime over a class of angular functions. Subsection~\ref{sec:stats} then studies the statistical counterparts of these problems and their solutions.

\subsection{Structural Analysis of Minimizers: Conditional, Asymptotic and Extreme Risks}
\label{subsec:approach}

The aim of this subsection is double: $(i)$ to show that under the
assumptions previously listed, the extreme quadratic risk $\riskext$
is minimized by angular prediction functions, that is functions
depending on the input through the angle only; $(ii)$ Although
$\riskinf$ and $\riskext$ are different risk functionals, they are
connected through % in the sense that
their respective minimizers and minimum values.

The first objective $(i)$ above is easily tackled.  Indeed, the
discussion below Equation~\eqref{eq:Pinf} shows that, under
Assumption~\ref{hyp:limit}, letting $\Theta_\infty = \theta(X_\infty)$
denote the angular component of $X_\infty$, the random pair
$(\Theta_\infty,Y_\infty)$ is independent from the norm
$\lVert X_\infty\rVert$, and in particular $Y_\infty$ and
$\lVert X_\infty\rVert$ are independent. Hence, the only useful piece
of information carried by $X_\infty$ to predict $Y_\infty$ is its
angular component $\Theta_\infty$. As a consequence the Bayes
regression function satisfies
$\bayesext(X_\infty) = \E[Y_\infty \mid X_\infty ] = \E[Y_\infty \mid
\Theta_\infty ]$ almost-surely. As a consequence we may write
$\bayesext = h_\infty\circ \theta$ for some function $h_\infty$
defined on the sphere $\sphere$.  Finally,
Assumption~\ref{hyp:cont_reg_func} ensures that $h_\infty$ may be
chosen as a continuous function. We summarize the discussion in the
following lemma.
\begin{lemma}\label{lem:angularMinimizer}
  Under Assumptions~\ref{hyp:bound} and~\ref{hyp:limit}, the extreme
  risk $\riskext$ has a minimizer (among all measurable functions)
  which may be written as $\bayesext(x) = h_\infty\circ\theta(x)$
  where $h_\infty: \sphere\to I$ is a bounded, continuous function.
\end{lemma}

The next result brings answers regarding the objective $(ii)$ outlined
above, by establishing a key connection between the (seemingly)
different problems of minimizing $\riskinf$ on the one hand, and
minimizing $\riskext$ on the other hand.  Recall from
Section~\ref{subsec:framework} that the extreme risk
$\riskext(f) = \E [ (f(X_\infty) - Y_\infty)^2 ]$ and the asymptotic
risk
$\riskinf(f) = \limsup_{t \rightarrow + \infty} \E [ (f(X) - Y)^2 \mid
\lVert X\rVert \s t ]$ are two different functionals, so that the
regression function $\bayesext$ is only defined as a minimizer of the
extreme risk $\riskext$ and not the asymptotic risk $\riskinf$.  In
the sequel we denote by $\riskext^*$ the minimum value of the extreme
risk, \ie
$ \riskext^*:=\inf_{f \text{ measurable}} R_{P_\infty}(f) =
\riskext(\bayesext)$. \new{The proof of
  Theorem~\ref{prop:conv_bayes_risk} is deferred to
  Section~\ref{sec:proof_thm_32} of the Appendix.}

\begin{theorem}\label{prop:conv_bayes_risk}
Under Assumptions \ref{hyp:bound} and  \ref{hyp:limit}, we have 
\begin{itemize}
\item[(i)]\label{prop:convergenceAngular} For any angular function of
  the kind $f(x) = h\circ\theta(x)$, where $h$ is a continuous
  function defined on $\sphere$, the conditional risk converges to the
  extreme risk, \ie $R_t(f) \tti R_{P_\infty}(f)$.  Thus for such
  prediction functions,
  $R_\infty(f) = \lim_{t\to+\infty} R_t(f) = R_{P_\infty}(f)$.
\end{itemize}
If in addition Assumption~\ref{hyp:cont_reg_func} is satisfied, then
the following assertions hold true.
\begin{itemize}
\item[(ii)]\label{prop:conv_bayes_risk_i} As $t \rightarrow +\infty$,
  the minimum value of $R_t$ converges to that of $\riskext$, \ie
  $R_t^* \tti \riskext^*$.
\item[(iii)]\label{equalityMinima} The minimum values of $\riskinf$
  and $\riskext$ coincide, \ie $\riskinf^* = \riskext^*$.
\item[(iv)]\label{prop:conv_bayes_risk_ii} The regression function
  $\bayesext$ minimizes the asymptotic conditional quadratic risk, \ie
  $\riskinf^* = \riskinf(\bayesext)$.
\end{itemize}
\end{theorem}

Observe that Theorem~\ref{prop:conv_bayes_risk} does not assert that
$R_t(f)$ converges to $\riskext(f)$ for all $f$, but the convergence
holds true for angular predictors $f=h\circ\theta$
(Property~\hyperref[prop:convergenceAngular]{\textit{(i)}} in the
statement).
Property~\hyperref[prop:conv_bayes_risk_ii]{\textit{(iv)}} discloses
that the solution $\bayesext$ of the extreme risk minimization problem
\new{$\min_{f \textrm{ measurable}}\riskext(f)$}, is also a minimizer
of the asymptotic conditional quadratic risk $\riskinf$ (and that the
minima coincide). Because $\bayesext = h_\infty\circ \theta$ is of
angular type, we thus obtain, under
Assumptions~\ref{hyp:bound},~\ref{hyp:limit}
and~\ref{hyp:cont_reg_func},
 \begin{equation}\label{eq:mainJustifProba}
    \inf_{f\text{measurable}} R_\infty(f) = \inf_{h\text{ measurable}} R_\infty(h\circ\theta). 
 \end{equation}
 In other words, the search for minimizers of $\riskinf$ may indeed be
 restricted\new{, without loss of generality,} to angular prediction
 functions.  This provides a first heuristic justification for the
 {\sc ROXANE} algorithm.  However in order to develop rigorous
 guarantees for the predictive performance of minimizers of the
 empirical criterion~\eqref{eq:emp_min_extr} computed by means of the
 {\sc ROXANE} algorithm, further assumptions regarding the class
 $\mathcal{H}$ of angular predictors are needed. In particular these
 additional assumptions ensure uniformity of the convergence result
 \hyperref[prop:convergenceAngular]{\emph{(i)}} from
 Theorem~\ref{prop:conv_bayes_risk}. This is the focus of the next
 section.

\subsection{Statistical Learning Guarantees}\label{sec:stats}

This section provides a nonasymptotic analysis of the approach
proposed for regression on extremes.  An upper confidence bound for
the excess of $R_{\infty}$-risk of a solution of
\eqref{eq:emp_min_extr} is established, when the class $\mathcal{H}$
over which empirical minimization is performed is of controlled
complexity, see Assumption~\ref{hyp:class} below.

The rationale behind the {\sc ROXANE} algorithm is to find an angular
predictive function that nearly minimizes the asymptotic conditional
quadratic risk $\riskinf$ \eqref{eq:asym_cond_risk}. Our ERM strategy
thus consists in solving an empirical version of the nonasymptotic
optimization problem
\begin{equation*}
\min_{h\in \mathcal{H}}R_{t}(h\circ\theta).
\end{equation*}
Recall that a heuristic justification for considering angular
classifiers is given by Eq.~\eqref{eq:mainJustifProba}, which is
itself a consequence of Theorem~\ref{prop:conv_bayes_risk}.  The
radial threshold $t$ is chosen as a relatively high quantile of the
empirical distribution of the radii $\lVert X_{i}\rVert$. In
particular, let $t_{n,k}$ denote the $1-k/n$ quantile of the norm
$\lVert X\rVert$, where $k\ll n $ is large enough so that a
statistical analysis remains realistic, but small enough so that the
distribution of $(X,Y)$ given that $\lVert X\rVert>t_{n,k}$ is close
to the limit $P_\infty$, see \eqref{eq:Pinf}. Then an empirical
version of $t_{n,k}$ is $\hat t_{n,k} = \lVert X_{(k)}\rVert$, the
$k^{th}$ largest order statistic of the norm already introduced in
Algorithm~\ref{algo}. In practice the number $k$ of retained extreme
statistics is a recurrent issue in EVA, for which no definite
theoretical answer exists, but which is a standard bias/variance
compromise. In our experiments, following standard practice we choose
$k$ by inspection of stability regions in Hill plots. In addition, in
a regression setting we consider feature importance summaries relative
to the radial variable, see Section~\ref{sec:exp} for details.

Summarizing, the objective minimized in Algorithm~\ref{algo} may be
viewed as an empirical version of the conditional risk $\risknk$ for a
predictive mapping of the form $h\circ\theta$. In the sequel we denote
by $\riskempnk$ this empirical objective
\begin{equation}\label{eq:emp_risk}
\riskempnk(f)=\frac{1}{k}\sum_{i=1}^k\Big( Y_{(i)}- f(X_{(i)})\Big)^2.
\end{equation}
The statistic above is not an average of independent random variables,
as it involves extreme order statistics of the norm. Thus
investigating its concentration properties \new{requires particular
  attention}.  The minimum is taken over a class $\mathcal{H}$ of
continuous bounded functions on $\mathbb{S}$ of controlled complexity
but hopefully rich enough to contain a reasonable approximant of
$h_\infty$ introduced in Lemma~\ref{lem:angularMinimizer}.  The
following assumption regarding $\mathcal{H}$ will turn out to be
sufficient to obtain a control of the deviations of the empirical
risk. In order to avoid measurability issues regarding supremum
deviations over the class $\mathcal{H}$, it is assumed throughout that
$\mathcal{H}$ is \emph{pointwise measurable}
(see~\cite{vandervaartWeakConvergenceEmpirical1996}, Example~2.3.4),
\ie\ that there exists a countable family
$\mathcal{H}_0\subset\mathcal{H}$, such that for all
$\omega\in\sphere$ and all $h\in\mathcal{H}$, there is a sequence
$(h_i)_{i\ge 1} \in\mathcal{H}_0$ such that
$h_i(\omega)\to h(\omega)$. This mild condition is satisfied in most
practical cases, in particular by parametric classes $\mathcal{H}$,
\ie classes indexed by a finite dimensional parameter
$\beta\in\rset^p$, which depend continuously on the parameter, \ie
such that $\lVert h_\beta- h_{\beta_n} \rVert_{\infty,\sphere} \to 0$
as $\beta_n\to\beta$.
 
\begin{assumption}\label{hyp:class}
  The pointwise measurable class $\mathcal{H}$ is a family of
  continuous, real-valued functions defined on $\mathbb{S}$; of {\sc
    VC} dimension $V_{\mathcal{H}}<+\infty$, and uniformly bounded by
  the same constant as the target $Y$ (see
  Assumption~\ref{hyp:bound}),
  $\forall h \in \mathcal{H}, \forall \omega \in \mathbb{S}$,
  $\vert h(\omega)\vert \leq M. $
\end{assumption}

Under the complexity hypothesis above, the following result provides
an upper confidence bound for the maximal deviations between the
conditional quadratic risk $\risknk$ and its empirical version
$\riskempnk$, uniformly over the class $\mathcal{H}$.

A similar result can be found in \cite{aghbalou2022cross} (Lemma C.3),
albeit within the more intricate setting of cross-validation. The
working assumptions in the cited reference are comparable, though not
identical; specifically, the VC assumption pertains to the loss class
rather than the class of prediction functions. The proof therein is
arguably more technical than necessary for the straightforward ERM
context considered here, primarily due to the need to address
dependencies between different folds of the cross-validation
scheme. We present a more concise, direct proof in Section~\ref{sec:D}
in the Appendix.

  Compared to the proof of Theorem~2 in \cite{jalalzai2018binary}, the
  main difference lies in the fact that here we focus on an empirical
  process indexed by a class of \emph{functions}, rather than by
  sets. Consequently, we cannot rely on Rademacher complexity bounds
  for a VC class of sets, which typically involve the shattering
  coefficient of the class and Sauer's lemma, see \emph{e.g.}
  \citep[]{BBL05}. Instead, we use complexity measures better suited
  to classes of functions.

In contrast to the approach in~\cite{jalalzai2018binary}, our argument
relies on polynomial control of the $L^2$-covering number of the class
$\big\{(x,y) \mapsto (h \circ \theta(x) - y)^2 \mid h \in
\mathcal{H}\big\}$, which leads to a control of expectations of
Rademacher processes indexed by functions, leveraging entropy bounds
\citep[][Proposition~2.1]{gine2001consistency}.

\begin{proposition}\label{prop:concentr1}
  Suppose that Assumptions \ref{hyp:bound} and \ref{hyp:class} are
  satisfied. Let $\delta\in (0,1)$. We have with probability larger
  than $1-\delta$
\begin{align*}
  \sup_{h\in \mathcal{H}}
  & \left\vert  \riskempnk(h\circ \theta)-\risknk(h\circ \theta) \right\vert
    \le \\
  &4 M^2\bigg(~
    \frac{2 \sqrt{ 2 \log( 3 / \delta)} + C \sqrt{V_{\mathcal{H}}}}{
    \sqrt{k}} ~+ ~
    \frac{  \frac{4}{3} \log( 3/\delta ) +  V_{\mathcal{H}}}{k} ~\bigg),
\end{align*}
where $C$ is a universal constant.
\end{proposition}
Proposition~\ref{prop:concentr1} controls only the statistical
deviations between the sub-asymptotic risk $\risknk$ and its empirical
version $\riskempnk$.  A control of the bias term $\risknk - \riskinf$
is given next, under appropriate complexity assumptions controlling
the complexity of class $\mathcal{H}$.  In particular
Assumption~\ref{hyp:class} can be traded against a total boundedness
assumption (Case~1. in Proposition~\ref{prop:prop2}) which is further
discussed below (Remark~\ref{rem:compactFamily}). Regarding the second
set of assumptions (Case 2. in Proposition~\ref{prop:prop2}), the
notation $\phi_{\theta,t}$ for $t\ge 1$ stands for the probability
density of the angular distribution
$\Phi_{\theta,t} = \mathcal{L}(\theta(X)\,|\, \lVert X \rVert \s t)$,
with respect to
$\Phi_{\theta,1} = \mathcal{L}(\theta(X)\,|\, \lVert X\rVert \s 1)$.
Indeed for any measurable set $A\subset \sphere$, if
$\PP[\Theta\in A \,|\, \lVert X\rVert \ge 1] = 0$ then also for any
$t\ge 1$, $\PP[\Theta\in A \,|\, \lVert X\rVert \ge t] = 0$, so that
$\Phi_{\theta,t}$ is indeed continuous with respect to
$\Phi_{\theta,1}$.
\begin{proposition}\label{prop:prop2}
  Suppose that Assumptions \ref{hyp:bound} and \ref{hyp:limit} are
  satisfied. Let $\mathcal{H}$ be a class of real-valued, continuous
  functions on $\mb{S}$. Assume that one of the two following
  conditions is satisfied.
\begin{enumerate}
	\item $\mathcal{H}$  is  totally bounded in the space $(\mathcal{C}(\sphere), \lVert\cdot \rVert_\infty)$ of continuous functions on $\sphere$ endowed with the supremum norm, or 
	\item $\mathcal{H}$ fulfills Assumption~\ref{hyp:class} and in
          addition, suppose that the conditional densities
          $\phi_{\theta,t}$ introduced above the statement satisfy
          $\sup_{t\geq 1,\; \omega \in
            \mb{S}}\phi_{\theta,t}(\omega)=D$, for some $0<D<\infty$.
\end{enumerate}
Then, as $t$ tends to infinity, we have
\begin{equation*}
\sup_{h\in \mathcal{H}}\left\vert  R_t(h\circ \theta)-R_{\infty}(h\circ \theta) \right\vert \tti 0.
\end{equation*}
\end{proposition}

The proof of Proposition~\ref{prop:prop2} is given in
Section~\ref{sec:proof_prop2} of Appendix~\ref{sec:proofs_main}.  The
two following remarks discuss the assumptions of
Proposition~\ref{prop:prop2}.

\begin{remark}[Totally bounded family of regression
  functions]\label{rem:compactFamily} Relying on a topological
  assumption on a set of regression functions such as total
  boundedness (\textit{i.e.} $\mathcal{H}$ may be covered by finitely
  many balls of radius $\varepsilon$, for any $\varepsilon>0$) is
  rather uncommon in statistical learning. However it turns out that
  this condition encompasses several standard algorithms.  Namely, if
  $\mathcal{H}$ is a parametric family indexed by a bounded parameter
  set, \ie $\mathcal{H} = \{h_\beta, \beta \in B\}$ for some
  $B\subset \rset^d$ of finite diameter, and if $h_\beta$ is
  Lipschitz-continuous with respect to $\beta$, \ie for some $C>0$,
  $\lVert h_\beta - h_\gamma\rVert_\infty \le C
  \lVert\beta-\gamma\rVert$ for all $\beta,\gamma\in B$, then
  $\mathcal{H}$ satisfies Condition 1. from
  Proposition~\ref{prop:prop2}.  As an example consider set of
  functions $h_\beta (\omega) = \langle \beta, \omega \rangle$ for
  $\omega\in\sphere$ with a bounded parameter set
  $B = \{\beta\in\rset^d: \lVert\beta\rVert_q\le \lambda\}$ for some
  fixed $\lambda>0$, where $\lVert\,\cdot\,\rVert_q$ is the $L^q$ norm
  on $\rset^d$, $q\ge 1$. The case $q=2$ (\emph{resp.} $q=1$)
  corresponds to a constrained Ridge (\emph{resp.} Lasso) regression.

\end{remark}

\begin{remark}[Bounded angular densities]\label{rem:bounded_dens}
  The second condition in Proposition~\ref{prop:prop2} implies that
  the angular measure $\Phi_{\theta,t}$ for large $t$ may not
  concentrate around sets that are negligible with respect to the
  `bulk' angular measure $\Phi_{\theta,1}$. This excludes situations
  where the limit angular measure $\Phi_{\theta}$ concentrates on
  lower dimensional subcones of $\rset^d$, whereas $\Phi_{\theta,1}$
  does not necessarily do so. This concentration phenomenon as
  $t\to+\infty$ % is sparser than $\Phi_{\theta,1}$,
  is precisely the framework considered in recent works on
  unsupervised dimension reduction for extremes where the goal is to
  uncover sparsity patterns in the limit angular measure $\Phi_\theta$
  which may not be representative of the bulk
  behavior~\citep{chiapino2019identifying,cooley2019decompositions,drees2021principal,goix2016sparse,goix2017sparse,meyer2021sparse}. How
  to relax Condition 2. in order to encompass such frameworks even
  though the family $\mathcal{H}$ does not satisfy Condition 1. is
  left to future research.
\end{remark}

Our main result below  summarizes the  results of
Section~\ref{sec:main} in the form of an upper confidence bound for
the excess of $\riskinf$-risk for any solution $\hatf$ of the problem
\begin{equation*}
\min_{h \in \mathcal{H}}\riskempnk(h\circ \theta).
\end{equation*}

\begin{theorem}[Bias-Variance decomposition for the excess of
  $R_\infty$ risk]\label{cor:concentr2}
  Let $\hatf = \hath\circ\theta$ be the prediction function issued by
  Algorithm~\ref{algo}.  Let
  Assumptions~\ref{hyp:bound},~\ref{hyp:limit},~\ref{hyp:cont_reg_func}
  and~\ref{hyp:class} be satisfied.  Recall $h_\infty$ from
  Lemma~\ref{lem:angularMinimizer} and that, from
  Theorem~\ref{prop:conv_bayes_risk},
  $\riskinf(h_\infty\circ\theta) = \inf_{h\textrm{
      measurable}}\riskinf(h\circ\theta) = \riskinf^*$.  For any
  $\delta>0$, with probability at least $1- \delta$, the excess
  $\riskinf$-risk of $\hatf$ satisfies
\begin{align}\label{eq:excess}
  \riskinf(\hatf)-\riskinf^* \le D_k + B_1(t_{n,k}) + B_2(\mathcal{H}), 
\end{align}
where $D_k, B_1,B_2$ are respectively a deviation term and two bias terms, 
\begin{equation*}
  \begin{cases}
    D_k = 8 M^2\Big(~
    \frac{2 \sqrt{ 2 \log( 3 / \delta)} + C \sqrt{V_{\mathcal{H}}}}{
    \sqrt{k}} ~+ ~
    \frac{  \frac{4}{3} \log( 3/\delta ) +  V_{\mathcal{H}}}{k} ~\Big)
    & \text{(deviations)} \\
    B_1(t) =  2\sup_{h \in \mathcal{H}} |R_\infty(h\circ\theta)- R_t(h\circ\theta)| & \text{(threshold bias)} \\
    B_2(\mathcal{H}) = \inf_{h \in\mathcal{H}} \riskinf(h \circ\theta) \;-\; \riskinf(h_\infty\circ\theta) &\text{(class bias)}. 
  \end{cases}
\end{equation*} The first bias term  $B_1(t_{n,k})$ in the above bound converges to zero as $n\to+\infty$, $k\to+\infty$, $k/n\to 0$ whenever the conditions of Proposition~\ref{prop:prop2} are met. 

  \end{theorem}
  \begin{proof}
    Assume for simplicity that the infimum of the $\riskinf$-risk over
    the class $\mathcal{H}$ is reached, \ie
    $\exists \hH\in\mathcal{H}: \riskinf(\hH\circ\theta) = \inf \{
    \riskinf(h\circ\theta), h \in\mathcal{H} \}$ (if this is not the
    case, consider an $\varepsilon$-minimizer $h_\varepsilon$ for
    arbitrarily small $\varepsilon$, and proceed).  Thus
    \begin{align}
      \riskinf(\hatf)
      & -\riskinf^* \le \riskinf(\hath\circ\theta) - \risknk(\hath\circ\theta )  + 
        \risknk(\hath\circ\theta ) - \riskempnk(\hath\circ\theta )  \nonumber \\
      &+ \riskempnk(\hath\circ\theta ) - \riskempnk(\hH\circ\theta ) + 
        \riskempnk(\hH\circ\theta ) - \risknk(\hH\circ\theta )  \nonumber \\
      &+ \risknk(\hH\circ\theta ) - \riskinf(\hH\circ\theta )+
        \riskinf(\hH\circ\theta ) - \inf_{h \textrm{ measurable}} \riskinf(h\circ\theta) \nonumber \\
      &+ \inf_{h \textrm{ measurable}} \riskinf(h\circ\theta) - \inf_{f \textrm{ measurable}} \riskinf(f). \nonumber
    \end{align}
    Because $\hath\circ\theta$ minimizes $\riskempnk$ and considering
    identity~\eqref{eq:mainJustifProba} (which holds because of
    Assumptions~\ref{hyp:bound},~\ref{hyp:limit},~\ref{hyp:cont_reg_func}),
    the above decomposition simplifies into
    \begin{align}
      \riskinf(\hatf) -\riskinf^*
      & \le 2\sup_{h\in\mathcal{H}}|\riskinf - \risknk|(h\circ\theta) + 
        2\sup_{h\in\mathcal{H}}|\risknk - \riskempnk|(h\circ\theta)  \nonumber \\
      &+    \riskinf(\hH\circ\theta ) - \inf_{h \textrm{ measurable}} \riskinf(h\circ\theta). 
        \nonumber 
\end{align}
The result follows by plugging in the deviation bound from
Proposition~\ref{prop:concentr1}.
\end{proof}

As it is generally the case in statistics of extremes, two types of
bias terms are involved in the upper bound~\eqref{eq:excess} of
Theorem~\ref{cor:concentr2}.  The first bias term $B_1(t)$ results
from the substitution of the conditional quadratic risk $\risknk$ for
its asymptotic limit $\riskinf$. While the weak additional assumptions
of Proposition~\ref{prop:prop2} ensure that this bias term vanishes as
$k/n\to 0$, a quantification of its decay rate would
require % to extend the conditional multivariate regular variation property and specify
second-order conditions, e.g. by extending the second order regular
variation setting of \cite{deHaan1996} to our context of joint regular
variation.
  
The second bias term is a model bias, induced by restricting the
family of all measurable functions on $\sphere$ to the class
$\mathcal{H}$ of controlled combinatorial complexity. It should be
noted that under the conditions of the statement,
Identity~\eqref{eq:mainJustifProba} ensures that restricting to
angular predictors does not induce any additional bias term compared
with considering a standard class for predictors taking the full
covariate (including the radius) as input.

\begin{remark}[Rate of convergence]
To establish the concentration bound stated in Proposition~\ref{prop:concentr1}, we employ general concentration results that are not ideally tailored for a regression context. A more detailed investigation might yield a bound on the stochastic error term of order $O(\log(k)/k)$, as suggested by standard concentration results (refer to \cite{GKKW02}, Section~11). This refined study is left to future work.
\end{remark}

\begin{remark}[Alternative to ERM]\label{rem:altern_erm}
  In the case where the output/response variable $Y$ is heavy-tailed
  (or possibly contaminated by a heavy-tailed noise), robust
  alternatives to the ERM approach exist and are preferable~(see
  \cite{lugosi2016risk}).  Extension of these robust alternatives to
  the present context of heavy-tailed input is beyond the scope of
  this paper but will be the subject of further
  research. 
\end{remark}

\section{Numerical Experiments and Case Study}\label{sec:exp}

We now investigate the performance of the approach previously
described and theoretically analyzed for regression on extremes from
an empirical perspective on several simulated and real datasets. The
code used to run our experiments is available at
\url{https://github.com/HuetNathan/extremeregression}.  The MSE in
extreme regions of angular regression functions output by specific
implementations of the {\sc ROXANE} algorithm are compared to those of
the classic regression functions, learned in a standard fashion. On
this occasion we propose a simple graphical diagnostic procedure
allowing to check visually whether the data meet our assumptions, in
particular Assumption~\ref{hyp:limit} which is central in our
work. More precisely we inspect the relative importance of the radial
variable $\lVert X\rVert $ for predicting $Y$ above increasing radial
thresholds.  We consider in Section~\ref{sec:simul_data} simulated
data in the additive and multiplicative models which are particular
instances of Example~\ref{ex:noise}. Section~\ref{sec:realdata}
develops a case study based on the financial dataset \emph{49 Industry
  Portfolios [Daily]} from Kenneth R.
French.

\subsection{Experimental Results on Simulated Data}\label{sec:simul_data}

As a first go, we focus on predictive performance of the {\sc ROXANE}
algorithm in terms of Mean Squared Error (MSE), with simulated data
following the general pattern detailed in Example~\ref{ex:noise}. More
precisely we consider an additive noise model and a multiplicative
noise model with heavy tailed design,
$Y = \Tilde{g}_0(X) + \varepsilon_0$, and
$Y = \varepsilon_1 \Tilde{g}_1(X)$, respectively.  Here, the noise
$\varepsilon_0$ is defined as a centered Gaussian variable, truncated
on the interval $[-1,1]$, with standard deviation $\sigma_0 = 0.1$,
with density $p_{\varepsilon_0}(z)$ proportional
to 
$ \1 \{|z| \le 1\} \exp(-z^2/(2\sigma_0^2))$.
The true regression function in the additive model is $f^*_0(x) =
  \Tilde{g}_0(x)$.  For the multiplicative model,
$\varepsilon_1$ is again a truncated Gaussian variable with the same
standard deviation
$\sigma_0$, however it is non-centered, with mean
$\mu=1$, and the truncation is performed outside the interval
$[0,2]$. % with mean $\mu =1$ and standard deviation $\sigma_1 = 0.1$,
The density $f_{\varepsilon_1}(z)$ for the noise
$\varepsilon_1$ is thus proportional to $\1 \{0 \le z \le 2\}
\exp(-(z-\mu)^2/(2\sigma_0^2))$ and the true regression function
  in the second model is simply $f^*_1(x) = \Tilde{g}_1(x)$. 

We then define the functions $\Tilde{g}_i$ as
$\Tilde{g}_0(x) = \beta^T\theta(x)(1+1/\lVert x\rVert )$, and
$\Tilde{g}_1(x) = \cos(1/\lVert x\rVert ) \\ \times \sum_{i=1}^{d/2}
(\theta(x)_{2i-1}-1/\lVert x\rVert ^{2})\sin(\pi
(\theta(x)_{2i}-1/\lVert x\rVert ^{2}) )$, for $x\in \rset^d$.  It is
shown in Section~\ref{sec:examples} from the Appendix
(Propositions~\ref{ex:add_noise} and \ref{ex:multiplicativeNoise})
that these two models satisfy our working assumptions, see also the
discussion following Example~\ref{ex:noise}. Concretely, the limit
regression functions are $f^*_{P_\infty, 0}(x) = \beta^\top\theta(x)$
and
$f^*_{P_\infty, 1}(x) = \sum_{i=1}^{d/2} \theta(x)_{2i-1} \sin(\pi
\theta(x)_{2i})$.

In the additive model (\emph{resp.} in the multiplicative model) the
design $X$ is generated according to a multivariate extreme value
distribution from the logistic family \citep{stephenson2003simulating}
with dependence parameter $\xi=1$, which means that extreme
observations occur very close to the axes (\emph{resp}. $\xi=0.7$,
meaning that the angular component of extreme observations is
relatively spread-out in the positive orthant of the unit sphere).
The input $1$-d marginals are standard Pareto with shape parameter
$\alpha = 1$ (\emph{resp.} $\alpha=3$).  We use the Euclidean norm to
define an extreme covariate,
$\lVert \cdot\rVert =\lVert \cdot\rVert _2$.

The simulated data is of dimension $d=7$ (\emph{resp.} $d=14$). For
both models, the size of the training dataset is
$n_{train} = 10\,000$, and the number of extreme observations retained
for training the {\sc ROXANE} algorithm is set to $k_{train}=1000$
($=n_{train}/10$).  The size of the test dataset is
$n_{test}=100\,000$ and the $k_{test}=10\,000$ ($=n_{test}/10$)
largest instances are used to evaluate predictive performance on
extreme covariates.  We consider three different regression algorithms
implemented in the \textit{scikit-learn} library \citep{scikit-learn}
with the default parameters, namely Ordinary Least Squares (OLS),
Support Vector Regression (SVR), %CART (tree)
and Random Forest (RF). Predictive functions are learned using
respectively $(i)$ the full training dataset, $(ii)$ a reduced dataset
composed of the $k_{train}$ largest observations
$X_{(1)}, \ldots X_{(k_{train})}$, and $(iii)$ an angular dataset
$\Theta_{(1)}, \ldots \Theta_{(k_{train})}$ consisting of the angles
of the $k_{train}$ largest observations.  These three options
correspond respectively to $(i)$ the default strategy (using the full
dataset), $(ii)$ a `reasonable' naive strategy (training on extreme
covariates for the purpose of predicting from extreme covariates),
$(iii)$ the ROXANE strategy that we promote in this paper,
corresponding to Algorithm~\ref{algo}. We evaluate the performance of
the outputs using the MSE computed on the test
set. Table~\ref{table:simu} shows the average MSE's when repeating
this experiment across $E=10$ independent replications of the
dataset. For the additive model the regression parameter $\beta$ is
randomly chosen for each replication, namely each entry of $\beta$ is
drawn uniformly at random over the interval $[0,1]$.

\begin{table}[t]
\vskip-0.4cm
\caption{Average MSE (and standard deviation) for regression functions trained using all observations, extreme observations and angles of extreme observations, over $10$ independent replications of the dataset generated in  the additive and the multiplicative noise models.}
\label{table:simu}
\vspace{0.2cm}
%\begin{center}
\centering
%\begin{small}
%\begin{sc}
\textsc{
\resizebox{\textwidth}{!}{
\begin{tabular}{lccccr}
\toprule
Methods/Models & Train on $X$ & Train on $X \mid \lVert X\rVert $ large & Train on $\Theta \mid \lVert X\rVert $ large  \\
\midrule
Add.: OLS & $ 23 {\scriptstyle\pm 29}$ & $ 3 {\scriptstyle\pm 6 }$ & $ \mathbf{0.003 {\scriptstyle\pm 0.001 }}$ \\
\phantom{Add.: }SVR & $ 0.13 {\scriptstyle\pm 0.01 }$ & $ 0.05 {\scriptstyle\pm 0.02 }$ & $ \mathbf{0.003 {\scriptstyle\pm 0.001 }}$ \\
%\phantom{Add.: }tree & $ 0.021 {\scriptstyle\pm 0.001 }$ & $ 0.012 {\scriptstyle\pm 0.003 }$ & $ \mathbf{0.008 {\scriptstyle\pm 0.001 }}$  \\
\phantom{Add.: }RF & $ 0.012 {\scriptstyle\pm 0.004 }$ & $ 0.007 {\scriptstyle\pm 0.002 }$ & $ \mathbf{0.004 {\scriptstyle\pm 0.001 }}$ \\
\midrule
Mult.: OLS & $ 0.006 {\scriptstyle\pm 0.001 }$ & $ 0.003 {\scriptstyle\pm 0.001 }$ & $ \mathbf{0.001 {\scriptstyle\pm 0.001 }}$  \\
\phantom{Mult.: }SVR & $ 0.0041 {\scriptstyle\pm 0.0002 }$ & $ 0.0038 {\scriptstyle\pm 0.0004 }$ & $ \mathbf{0.0034 {\scriptstyle\pm 0.0003 }}$ \\
%\phantom{Mult.: }tree & $ 0.0037 {\scriptstyle\pm 0.0001 }$ & $ 0.0028 {\scriptstyle\pm 0.0002 }$ & $ \mathbf{0.0008 {\scriptstyle\pm 0.0002 }}$  \\
\phantom{Mult.: }RF & $ 0.0020 {\scriptstyle\pm 0.0001 }$ & $ 0.0013 {\scriptstyle\pm 0.0001 }$ & $ \mathbf{0.0004 {\scriptstyle\pm 0.0001 }}$ \\
\bottomrule
\end{tabular}
}
%\end{sc}
%\end{small}
%\end{center}
%\vskip -0.3in
}

\vskip -0.4cm
\end{table}

With both models, the approach we promote for regression on extremes
clearly outperforms its competitors, no matter the algorithm (\ie the
model bias) considered. This paper being the first to consider
regression on extremes (see Remark \ref{rem:altern_erm} for a
description of regression problems of different nature with
heavy-tailed data), no other alternative approach is documented in the
literature.

Besides prediction performance, we propose to assess the validity of
our main modeling assumption (Assumption~\ref{hyp:limit}) by
inspecting the \emph{variable importance} (\emph{a.k.a.}
\emph{feature importance}, see e.g.~\cite{gromping2015variable} and
the references therein) of the radial variable $\lVert X\rVert $
compared with the angular variables $\Theta_j, j\le d$, for the
purpose of predicting the target $Y$.  Indeed, under
Assumption~\ref{hyp:limit}, the variables $Y$ and $\lVert X\rVert $
are asymptotically independent conditional on $\{\lVert X\rVert >t\}$
as $t\to+\infty$, so that the variable importance of
$\lVert X\rVert $, when restricting the training set to regions above
increasingly large radial thresholds, should in principle
vanish. 

We consider here two widely used measures of feature importance, Gini
importance--or Mean Decrease of Impurity,
\citep{breiman2017classification,wei2015variable}--and Permutation
feature importance \citep{Breiman2001,wei2015variable} in the context
of Random Forest prediction, as implemented in the
\textit{scikit-learn} library. Gini importance measures a mean
decrease of impurity in a forest of trees, between parent nodes
involving a split on the considered variables, and their child
nodes. Gini score is normalized so that the sum of all importance
scores across variables equals $1$. Permutation importance compares
the prediction performance of the original input dataset with the same
dataset where the values of the considered variable have been randomly
shuffled.  A large score indicates a high predictive value of the
variable for both measures.

\begin{figure}[ht] 
	\begin{minipage}{0.485\textwidth}
		%\begin{subfigure}
			\centering
			\stackunder[5pt]{\includegraphics[width=\textwidth]{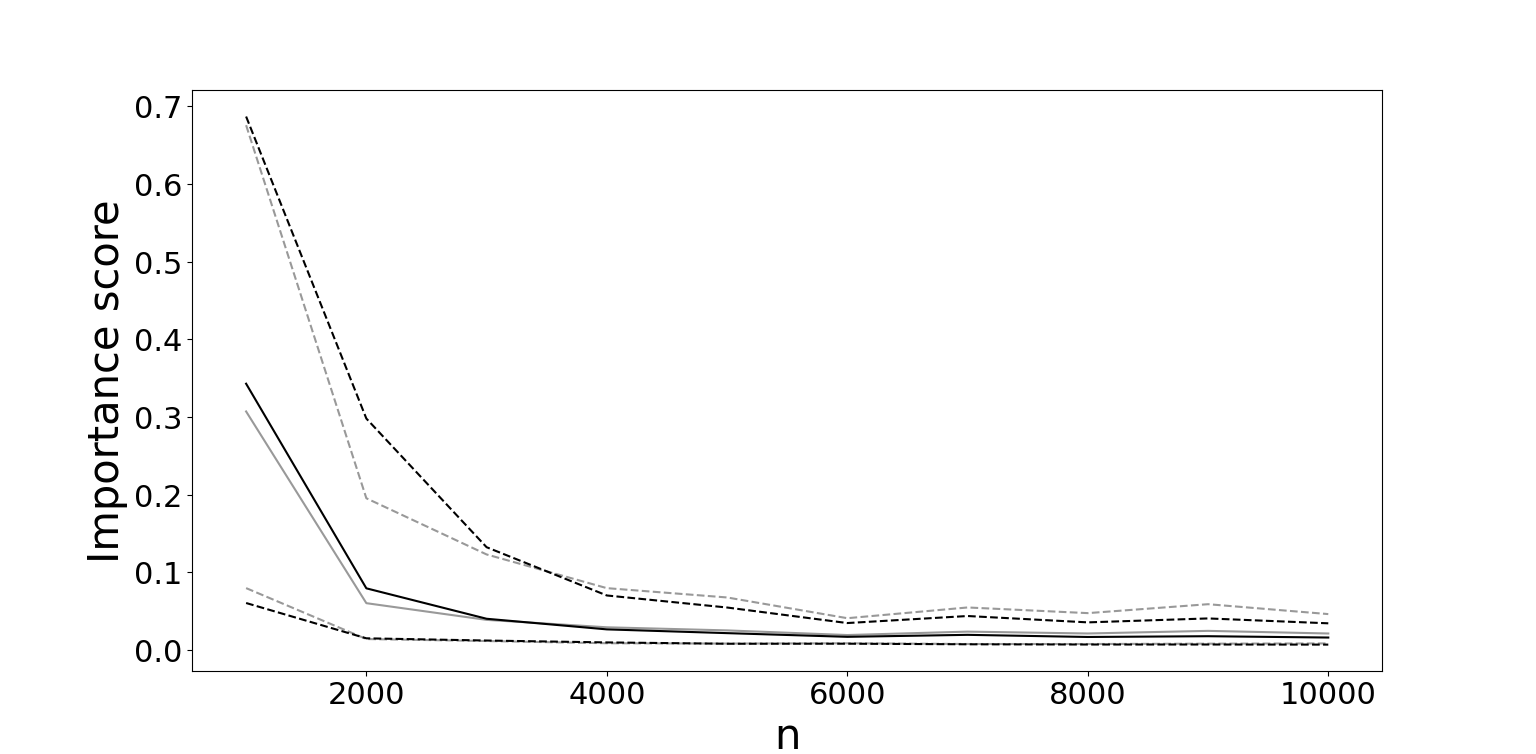}}{Additive model}
		%\end{subfigure}
	\end{minipage}
     	\hfill
     \begin{minipage}{0.485\textwidth}
     	%\begin{subfigure}
			%\centering
	\stackunder[5pt]{\includegraphics[width=\textwidth]{
            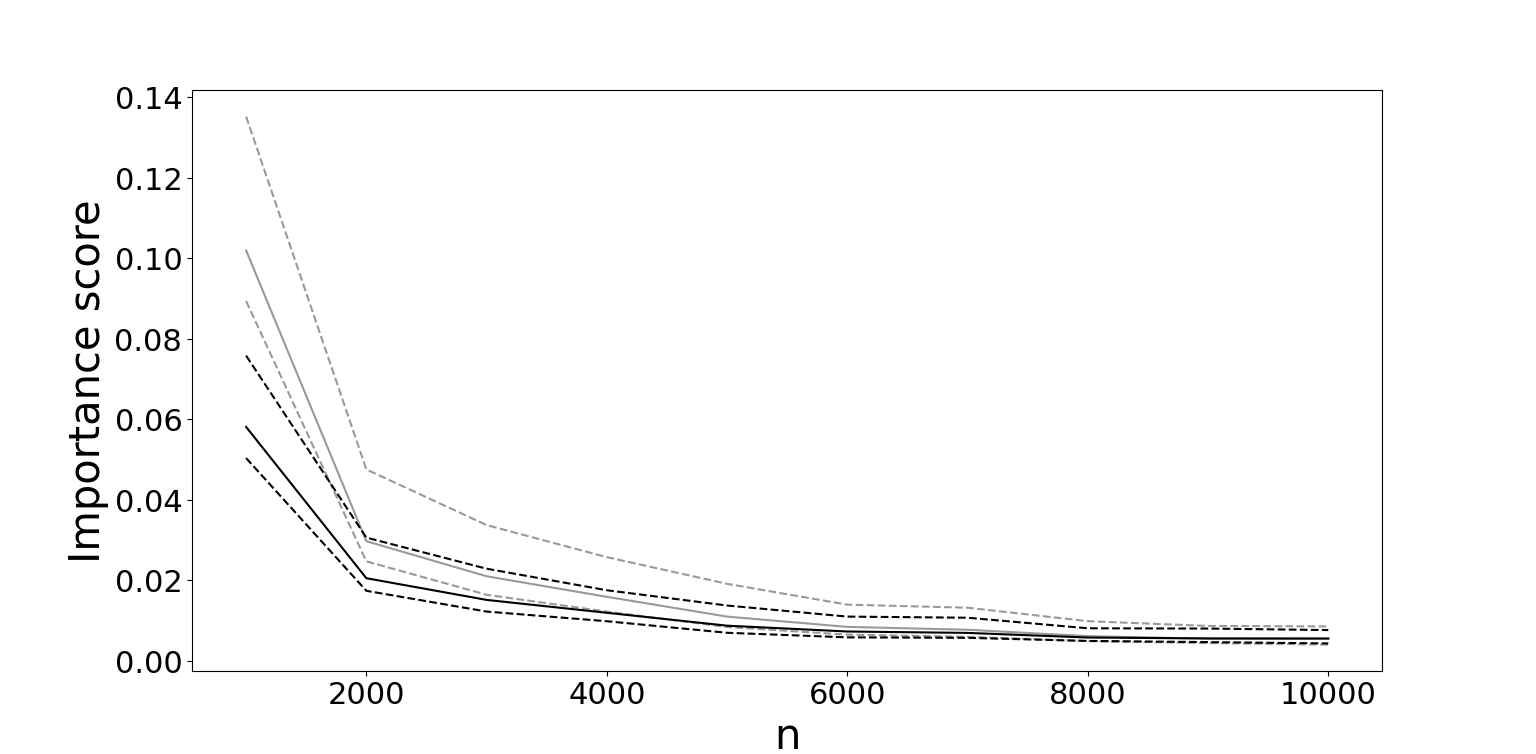}}{Multiplicative model}
		%\end{subfigure}
	\end{minipage}
        \caption{\label{fig:imp_simu} Average permutation and Gini
          importance measures of the radial variable using the RF
          algorithm in the additive noise model (left) and the
          multiplicative noise model (right) over $10$ replications,
          as a function of the total sample size $n_{imp}$ for fixed
          extreme training size $k_{imp}$.  Solid black line: average
          Gini importance. Solid grey line: average Permutation
          importance. Dashed lines: empirical 0.8-interquantile
          ranges.}
\end{figure}

The aim of this second experiment is to illustrate the decrease of the
radial feature importance for reduced datasets involving increasingly
(relatively) large inputs. To cancel out the perturbation effect of
reduced sample sizes, we fix a training size $k_{imp} = 1 000$ and we
simulate increasingly large datasets of size
$n_{imp} \in \{k_{imp}, 2k_{imp}, \ldots, 10 k_{imp} \}$ in the
additive and multiplicative models described above. Then for
$j\in\{1,\ldots,10\}$ the $k_{imp}$ largest observations in terms of
$\lVert X\rVert $ among $n_{imp} = j k_{imp}$ are retained, a random
forest is fitted with input variables
$(\lVert X\rVert , \Theta_1, \ldots, \Theta_d)$, and the Gini and
Permutation scores are computed.  Figure~\ref{fig:imp_simu} shows the
average scores obtained over $10$ independent experiments, together
with interquantile ranges, as a function of the full sample size
$n_{imp}$. In both models, the decrease of both scores is obvious. In
particular in terms of Gini measure, the relative importance of the
radius decreases from $38\%$ to $1\%$ for the additive model and from
$6\%$ to $<1\%$ for the multiplicative model.

\subsection{Case Study on Real Data}\label{sec:realdata}

\begin{figure}[t]
  \centering
\includegraphics[width=\textwidth/2]{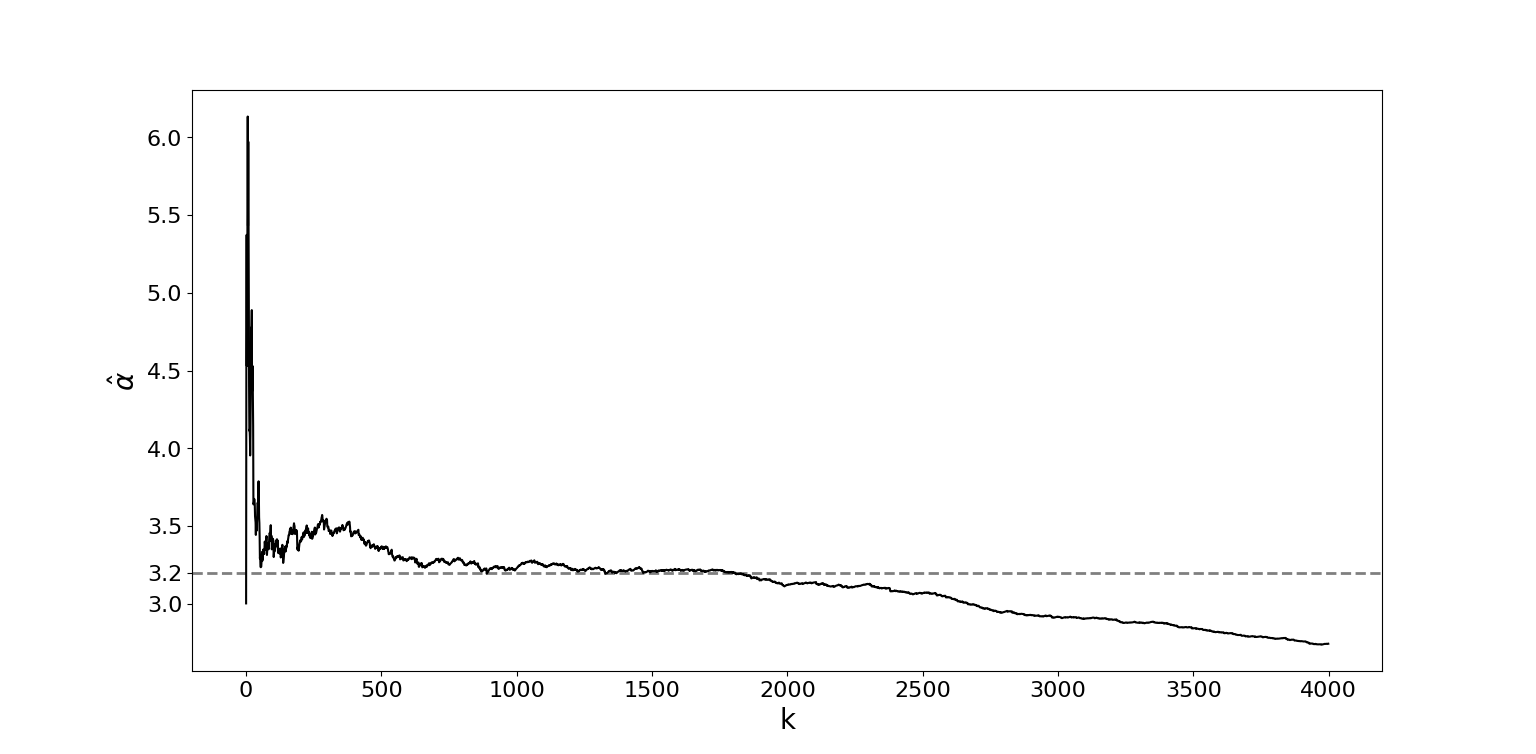}
\caption{\label{fig:Hillplot}Hill plot for the radial variable of the 49 Industry Portfolio Daily dataset: estimation of the extreme value index $\gamma=1/\alpha$ with the Hill estimator using the $k$ largest order statistics of $\lVert X\rVert $, as a function of $k$. 
}
\end{figure}

Encouraged by this first agreement between theoretical and numerical results, experiments on real data are conducted.  We place ourselves in the setting of Example~\ref{ex:predictionRVvect} where the target is one particular variable in a multivariate regularly varying random vector.  We consider a financial dataset, namely \emph{49 Industry Portfolios [Daily]} from Kenneth R. French - Data Library (\url{https://mba.tuck.dartmouth.edu/pages/faculty/ken.french/data_library.html}). A study of extremal clustering properties within this dataset has already been carried out by \cite{meyer2023multivariate}. This dataset comprises %value-averaged weighted
daily returns of 49 industry portfolios, within the time span from January 5th, 1970 to October 31st, 2023. Rows containing any NA values are removed, resulting in a dataset of dimension $d=49$ and size $n = 13577$. Figure~\ref{fig:Hillplot} displays a Hill plot of the radial variable (w.r.t. $\lVert \cdot\rVert _2$), with a rather  wide stability region,  roughly between $k=500$ and $k=2000$,  which suggests that regular variation is indeed present,  with regular variation index $\alpha\approx 3.2$. %
We consider separately the first three variables as output (target)
variables, namely \emph{Agric} (\ie "Agriculture"), \emph{Food} (\ie
"Food Products"), and \emph{Soda} (\ie "Candy and Soda"). Each choice
of a target variable defines a regression problem, which involves
predicting the target based on a covariate vector of \new{dimension
  $d=48$, composed of all the other variables.}  The dataset is
randomly split into a test set of size $n_{test}=4073$ ($30\%$ of the
data), and a train set of size $n_{train} = 9504 =n -
n_{test}$. % with the rest.
As suggested by the Hill plot (Figure~\ref{fig:Hillplot}), the number
$k_{train}$ of extreme observations used at the training step is set
to $k_{train} = \lfloor n_{train} / 5 \rfloor = 1900$. On the other
hand, at the testing step, to evaluate the extrapolation performance
of our method, we fix $k_{test}$ to a smaller fraction of the test
set, $k_{test} = \lfloor n_{test} / 10 \rfloor=407$.  In this setting,
paralleling our experiments with simulated data, we compare in
Table~\ref{table:real} the performance of regression functions learned
using the full training dataset (first column), the truncated version
composed of the $k_{train}$ largest observations (second column) and
the angles of the truncated version (ROXANE, promoted approach, third
column).  Again, we consider the OLS, SVR, and RF algorithms. To
  make the OLS algorithm competitive with the other two, which are
  better suited for high-dimensional settings, a preliminary, naive
  dimension reduction step is performed before training the OLS
  algorithm. Specifically, only the 10 covariates most correlated with
  the output variable are retained in the covariate vector for OLS,
  where the correlation is estimated over the entire training set (not
  only extremes).

  Second, regarding the nature of the target, we have endeavored to
  make the comparison as fair as possible. Specifically, we train
  ROXANE (Algorithm~\ref{algo}) on the rescaled target
  \( Y = \tilde{X}_{d+1} / \lVert \tilde{X} \rVert \), where
  \(\tilde{X} = (\tilde{X}_1, \ldots, \tilde{X}_{d+1})\), in
  accordance with our theory, but we evaluate the output in terms of
  MSE on the back-transformed (raw) target $\tilde{X}_{d+1}$. In
  practice, the output \(\hat{Y}\) from ROXANE is plugged into the
  formula \(\tilde{X}_{d+1} = Y \lVert X \rVert / \sqrt{1 - Y^2}\),
  where \(X = (X_1, \ldots, X_d)\), which yields an estimate
  \(\hat{X}_{d+1}\). This approach is chosen for the sake of realism
  in potential applications where the focus would be on the raw target
  rather than the normalized version.  In contrast, for the two other
  competitors (first two columns of Table 2), as there is no guiding
  theory, we proceed in a naive yet potentially efficient manner. That
  is, the training step is also performed using the raw target
  $\tilde{X}_{d+1}$.

  This setup could potentially disadvantage ROXANE, as, unlike the
  other two competitors, the minimization problem at the training step
  differs from that at the testing step.  

  The results gathered in Table~\ref{table:real} are the average MSE's
  obtained when repeating $10$ times the procedure described above
  with random splits of the dataset into a train and a test set. These
  results provide evidence that conditionally on the other (covariate)
  variables being large, our method ensures, in most cases, better
  reconstruction of the target variable than the default strategy
  (first column) and the intermediate strategy (second column). For
  predicting the \emph{Soda} variable however, the default strategy
  with OLS obtains the best scores. This suggests that convergence of
  the conditional distribution of excesses towards its limit as
  in~\eqref{eq:multivariate} is somewhat slower for the subvector
  $(\tilde X_1,\ldots, \tilde X_{d+1})$ where $\tilde X_{d+1}$ is
  \emph{Soda} and $\tilde X_1,\ldots, \tilde X_d$ are the $10$
  selected variables based on their correlation with \emph{Soda}.

\begin{figure}[h!] 
	\begin{minipage}{0.49\textwidth}
		%\begin{subfigure}
			\centering
			\stackunder[5pt]{\includegraphics[width=\textwidth]{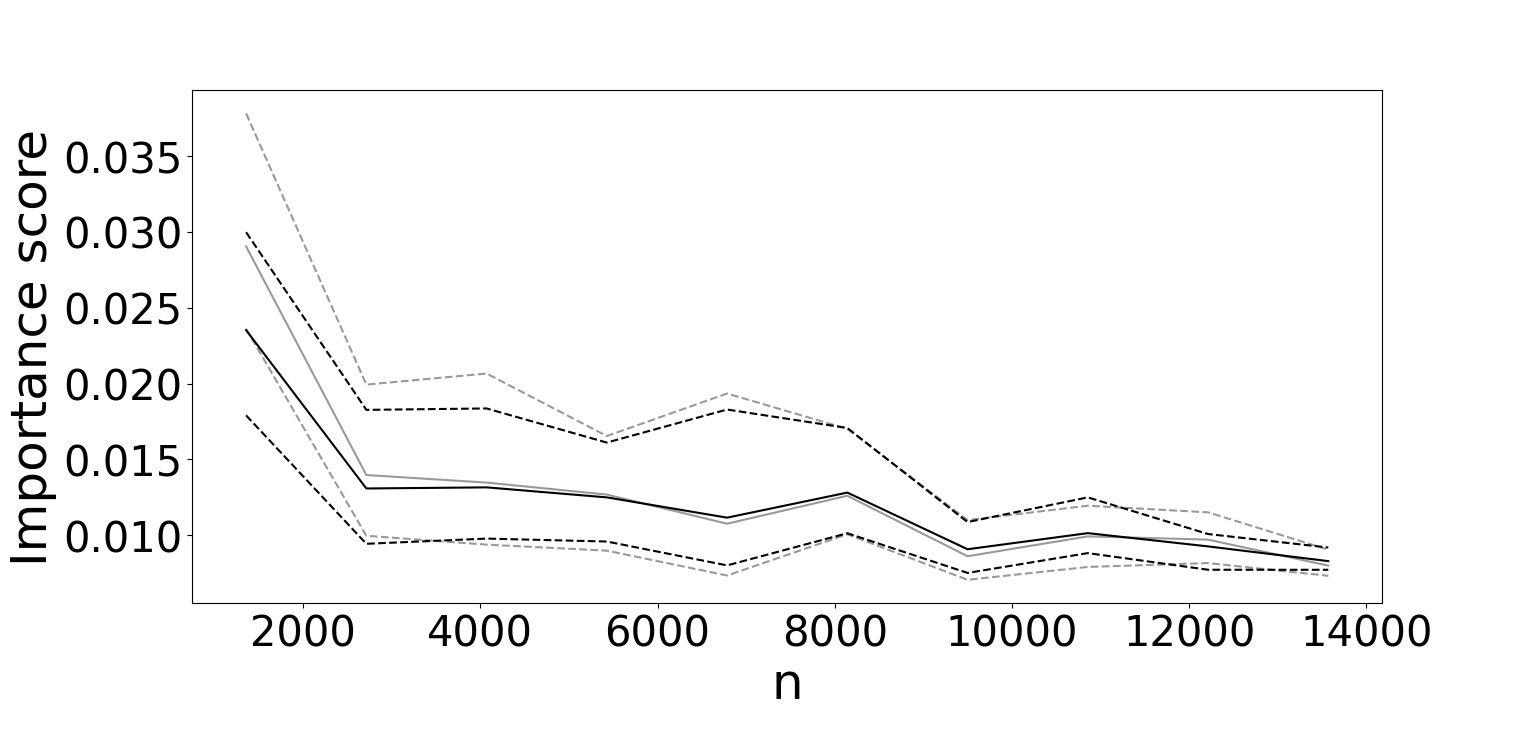}}{\emph{Agric}}
		%\end{subfigure}
	\end{minipage}
     	\hfill
     \begin{minipage}{0.49\textwidth}
     	%\begin{subfigure}
			%\centering
	\stackunder[5pt]{\includegraphics[width=\textwidth]{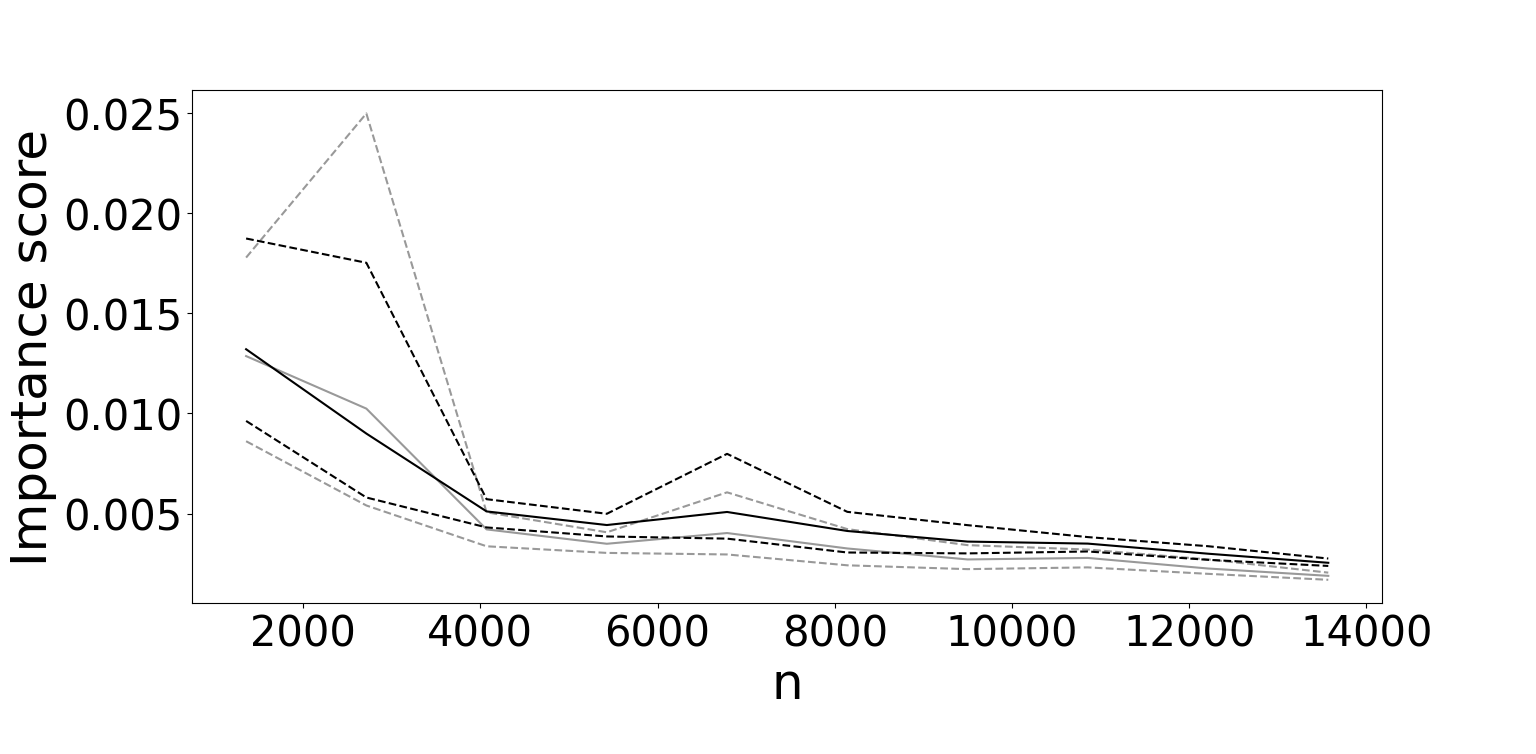}}{\emph{Food}}
		%\end{subfigure}
	\end{minipage}
	\center
	\begin{minipage}{0.49\textwidth}
     	%\begin{subfigure}
			%\centering
	\stackunder[5pt]{\includegraphics[width=\textwidth]{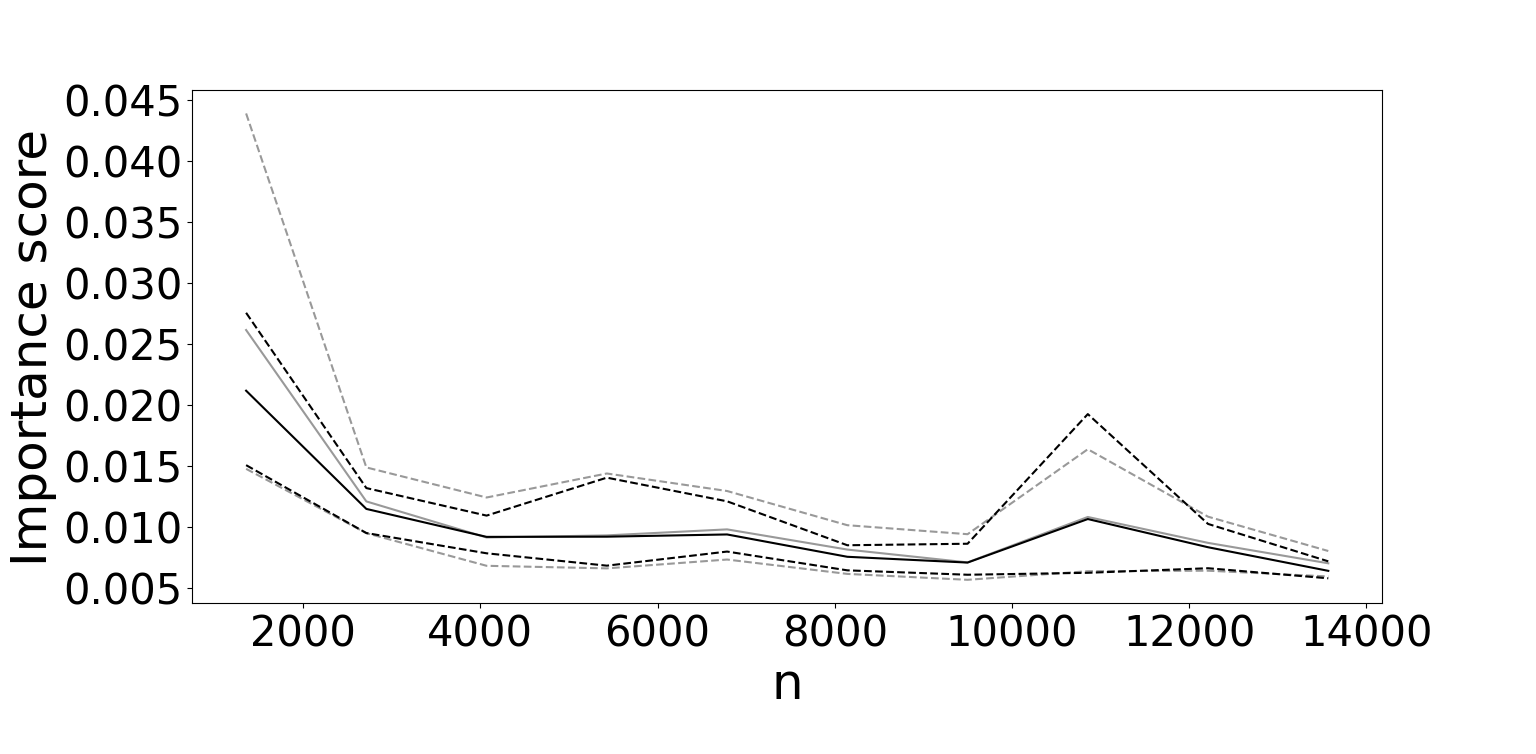}}{\emph{Soda}}
		%\end{subfigure}
	\end{minipage}
        \caption{\label{fig:imp_real}Average permutation and Gini
          importance measures of the radial variable for predicting
          \emph{Agric} (top left), \emph{Food} (top right) and
          \emph{Soda} (bottom) variables using the RF over $10$
          randomly shuffled datasets. At each measurement, $1357$
          extreme observations are selected from a dataset whose total
          size increases from $1357$ to $13570$ with increments of
          $1357$. Solid black line: average Gini importance. Solid
          grey line: average Permutation importance. Dashed lines:
          empirical 0.8-interquantile ranges.}
\end{figure}

  This intuition is confirmed by the graphs of variable importance
  displayed in Figure~\ref{fig:imp_real}, again paralleling the ones
  of Figure~\ref{fig:imp_simu} and fully described in
  Section~\ref{sec:simul_data}. In Figure~\ref{fig:imp_real}, for
  simplicity, the importance scores are computed in a prediction task
  where the covariate vector includes all the available variables,
  except from the target ($48$ of them). Also the target variable for
  the RF algorithm is the rescaled variable
  $Y = \tilde X_{d+1} / \lVert \tilde X \rVert $.  Whereas the radial
  importances decreases monotonically when the target variable in
  \emph{Agric} and \emph{Food}, the third panel dedicated to the
  target variable \emph{Soda} displays a local maximum in radial
  importance around $n=11\,000$. This value corresponds to a ratio
  $k/n \approx 0.12$ which is near the ratio $1/10$ considered for the
  testing step in our experimental results reported in
  Table~\ref{table:real}. This may explain our comparatively poor
  results for this particular variable.  However for all three target
  variables, overall, both Gini and Permutation importance score
  decrease significantly, as the ratio $k/n$ decreases.  In particular
  for Gini importance, the relative radial importances are
  approximately $2\% \approx 1/48$ when $n=k$, which is to be expected
  when all variables have equal importance. On the other hand when
  $n=10 k$, all three Gini importances are less than $1\%$.

\begin{table}[t!]
\vskip-0.4cm
\caption{\label{table:real}Average MSE (and standard deviation) for predictive functions learned using all observations, extremes ($20\%$) and angles of the extreme observations with output variables \emph{Agric} over $10$ random splits of each dataset.}
\vspace{-0.1cm}
%\begin{center}
\centering
%\begin{small}
%\begin{sc}

\textsc{
\resizebox{\textwidth}{!}{
\begin{tabular}{lccccr}
\toprule
Methods/Models & Train on $X$ & Train on $X \mid \lVert X\rVert $ large & Train on $\Theta \mid \lVert X\rVert $ large \\
\midrule
\emph{Agric}: OLS & $ 3.30 {\scriptstyle\pm 0.47}$ & $ 3.26 {\scriptstyle\pm 0.47 }$ & $\mathbf{3.25 {\scriptstyle\pm 0.44 }}$\\
\phantom{\emph{Agric}: }SVR & $ 4.76 {\scriptstyle\pm 0.56 }$ & $ 3.98 {\scriptstyle\pm 0.51 }$ & $ \mathbf{3.74 {\scriptstyle\pm 0.50 }}$ \\
%\phantom{\emph{Agric}: }\textcolor{red}{tree} & $ 7.16 {\scriptstyle\pm 0.63 }$ & $ 6.40 {\scriptstyle\pm 0.66 }$ & $ \mathbf{6.29 {\scriptstyle\pm 0.54 }}$ \\
\phantom{\emph{Agric}: }RF & $ 3.47 {\scriptstyle\pm 0.47 }$ & $ 3.48 {\scriptstyle\pm 0.47 }$ & $ \mathbf{3.28 {\scriptstyle\pm 0.52 }}$
\\
\midrule
\emph{Food}: OLS & $ 0.69 {\scriptstyle\pm 0.087 }$ & $ \mathbf{0.678 {\scriptstyle\pm 0.082 }}$ & $ 0.680 {\scriptstyle\pm 0.085 }$\\
\phantom{\emph{Food}: }SVR & $ 1.8 {\scriptstyle\pm 0.4 }$ & $ 1.3 {\scriptstyle\pm 0.4 }$ & $ \mathbf{0.87 {\scriptstyle\pm 0.08 }}$ \\
%\phantom{\emph{Food}: }\textcolor{red}{tree} & $ \mathbf{1.38 {\scriptstyle\pm 0.17 }}$ & $ 1.42 {\scriptstyle\pm 0.01 }$ & $ 1.44 {\scriptstyle\pm 0.14}$\\
\phantom{\emph{Food}: }RF & $ 0.70 {\scriptstyle\pm 0.13 }$ & $ 0.72 {\scriptstyle\pm 0.12 }$ & $ \mathbf{0.63 {\scriptstyle\pm 0.08 }}$
\\
\midrule
\emph{Soda}: OLS & $ \mathbf{2.35 {\scriptstyle\pm 0.21 }}$ & $ 2.37 {\scriptstyle\pm 0.21 }$ & $ 2.42 {\scriptstyle\pm 0.21 }$\\
\phantom{\emph{Soda}: }SVR & $ 4.0 {\scriptstyle\pm 0.5 }$ & $ 3.1 {\scriptstyle\pm 0.5 }$ & $ \mathbf{2.8 {\scriptstyle\pm 0.2 }}$ \\
%\phantom{\emph{Soda}: }\textcolor{red}{tree} & $ 5.2 {\scriptstyle\pm 0.4 }$ & $ \mathbf{4.9 {\scriptstyle\pm 0.5 }}$ & $ 5.0 {\scriptstyle\pm 0.7 }$ \\
\phantom{\emph{Soda}: }RF & $ 2.46 {\scriptstyle\pm 0.28 }$ & $ 2.46 {\scriptstyle\pm 0.25 }$ & $ \mathbf{2.34 {\scriptstyle\pm 0.18 }}$
\\
\bottomrule
\end{tabular}
%}
}
}
\end{table}

\section{Conclusion}\label{sec:conclusion}
We have provided a sound ERM approach to the generic problem of statistical regression on extreme values. The asymptotic framework we have developed crucially relies on the (novel) notion of \textit{joint regular variation} w.r.t. some multivariate component. When the distribution of the pair $(X,Y)$ is regularly varying w.r.t. the first component, the problem can be stated and analyzed in a rigorous manner. We have described sufficient conditions under which the optimal solution  can be nearly recovered with nonasymptotic guarantees by implementing a variant of the ERM principle, based on the angular information carried by a fraction of the largest observations only. We have also carried out numerical experiments to support the approach promoted, highlighting the necessity of using a dedicated methodology to perform regression on extreme samples with guarantees.

\new{
Our work paves the way for several natural extensions. First, our choice of the quadratic loss is motivated by simplicity and for illustrative purposes. Other losses, such as the pinball loss, will be considered in future work, which could serve as a first step toward generalizing the results of \cite{buritica2024progression} to multivariate settings. Additionally, to address high-dimensional problems, the ROXANE algorithm can naturally be extended to incorporate penalized loss functions, such as LASSO regression, as in \cite{clemenccon2025weak}, or be combined with tailored variable selection procedures, as developed in \cite{de2022extreme}.

For the sake of simplicity and clarity, we have chosen to work with a bounded response variable \( Y \), while proposing a rescaling mechanism to enforce this assumption for unbounded targets. An alternative approach would be to consider unbounded response variables from the outset, although the technical price to pay would be non-negligible since concentration tools requiring boundedness would no longer be applicable.

In such a context, a natural alternative to the one-component regular variation Assumption~\ref{hyp:limit} would be to allow for a rescaling of the target \( Y \) in the left-hand side, say \( Y/c(t) \). In other words, to assume `partial regular variation' \citep[see Chapter 3 of][and the references therein]{kulik2020heavy} of the pair \((X, Y)\), thus connecting the statistical learning framework developed in the present paper with the vast literature related to hidden regular variation \citep{resnick2002hidden} and conditional extremes \citep{heffernan2007limit}.
}

\clearpage
%%%%%%%%%%%%%%%%%%%%%%%%%%%%%%%%%%%%%%%%%%%%%%
%% Example with multiple Appendixes:        %%
%%%%%%%%%%%%%%%%%%%%%%%%%%%%%%%%%%%%%%%%%%%%%%
\begin{appendix}
  The Appendix is structured as follows.  In
  Section~\ref{sec:cond_mult_reg_var}, regular variation with respect
  to the first component, as introduced in Assumption~\ref{hyp:limit},
  is rephrased into equivalent conditions that facilitate connection
  with existing literature on regular variation.
  Section~\ref{sec:proofs_section_proba} gathers auxiliary results and
  some technical proofs for the results stated in
  Section~\ref{sec:background}. The proofs of our main results from
  Section~\ref{sec:main} are gathered in Section~\ref{sec:proofs_main}.

\section{Multivariate Regular Variation w.r.t. the Covariable}\label{sec:cond_mult_reg_var}
This section makes explicit the connection between
Assumption~\ref{hyp:limit} and the regular variation framework on a
metric space developed in~\cite{lindskog2014regularly}. We also
provide alternative formulations of Assumption~\ref{hyp:limit}.
Following whenever possible the notations
of~\cite{lindskog2014regularly}, let $\mathcal{Z} = \rset^d\times I$
where we recall $I=[-M,M]$ (in~\cite{lindskog2014regularly} the
ambient space $\mathcal{Z}$ is denoted by $\sphere$ which interferes
with our notation for the unit sphere). The ambient space
$\mathcal{Z}$ is endowed with the Euclidean product
metric,
$$d((x_1,y_1), (x_2,y_2)) = \sqrt{\lVert x_1-x_2\rVert^2 +(y_1-y_2)^2 },$$ so
that $(\mathcal{Z},d)$ is a complete separable metric space.  Define a
scalar `multiplication' on $\mathcal{Z}$ as
$\lambda.(x,y) = (\lambda x,y)$, $\lambda> 0$, which is continuous and
satisfies the associativity property
$\lambda_1 . (\lambda_2 .z) =(\lambda_1\lambda_2) . z$, and $1.z =
z$. This scalar multiplication induces a scaling operation on sets,
$\lambda A = \{\lambda.z, z\in A\}$ for $A\subset \mathcal{Z}$.
Consider the set
$\mathbb{C} = \{0_{\rset^d}\}\times I\subset \mathcal{Z}$. Then
$\mathbb{C}$ is a closed set which is preserved by the above scaling
operation, \ie it is a closed cone. For $z=(x,y)$ we have
$d(z,\mathbb{C}) = \lVert x\rVert $, whence
$d(x,\mathbb{C}) < d(\lambda x, \mathbb{C})$ for $\lambda> 1$.  Thus
Assumptions A1, A2, A3 in~\cite{lindskog2014regularly}, Section~3, are
satisfied. Let $\mathbb{O} = \mathcal{Z}\setminus\mathbb{C}$ and
introduce
$\mathbb{C}^r = \{z \in \mathbb{O}: d(z, \mathbb{C})>r\}, r\ge 0$.
In~\cite{lindskog2014regularly}, the class of Borel measures on
$\mathbb{O}$ whose restriction to $\mathcal{Z}\setminus \mathbb{C}^r$
is finite for any $r>0$ is denoted by $\Mo $. Then convergence of a
sequence of measures $\mu_n \in \Mo $ towards $\mu\in\Mo $ is defined
as convergence of functional evaluations $\mu_n(f)\to\mu(f)$ for
$f\in\mathcal{C}_{\mathbb{O}}$, the class of continuous functions on
$\mathcal{Z}$ which vanish on a neighborhood of $\mathbb{C}$, \ie
whose support is a subset of $\mathbb{C}^r$ for some $r>0$. A measure
$\nu\in\Mo $ is called \emph{regularly varying} with limit measure
$\mu\in\Mo$ and scaling sequence $b_n\in\rset$, if $b_n$ is
increasing, regularly varying in $\rset$ and if the sequence of
measures $b_n\nu(n\,\cdot\,)$ converges in $\Mo $ towards $\mu$ (see
Definitions~3.1, 3.2 in~\cite{lindskog2014regularly}). From the
Portmanteau Theorem 2.1 in~\cite{lindskog2014regularly} and the series
of equivalences in Theorem 3.1 of the same reference, our
Assumption~\ref{hyp:limit} is equivalent to assuming that the
distribution $P$ of the random pair $(X,Y)$ is regularly varying in
$\Mo $ with scaling sequence $b_n$ and limit measure $\mu$, with the
notations of Section~\ref{subsec:framework}.

\begin{theorem}\label{thm:limit}
  Let $\mathbb{O},\mathbb{C}$ be defined as above the statement, let  $\mu\in\Mo$ be a nonnull measure and  let $b(t)$ be a regularly varying function on $\rset^+$ with index $\alpha>0$. Let  $(X,Y)\sim P$ be a random pair valued in $\rset^d\times I$. %Recall that $\B=\{x \in \rset^d, \lVert x\| \le 1\}$.
The following assertions are equivalent. 
\begin{itemize}
\item[(i)]\label{eq:cond_reg_var_appendix} The random pair $(X,Y)$ satisfies Assumption~\ref{hyp:limit} from the main paper with limit measure $\mu$ and normalizing function $b$. 
\item[(ii)]\label{eq:cond_reg_var_appendix2} 
  For any bounded and continuous function 
  $h:\mathbb{O}\to\rset$ 
  that vanishes in a neighborhood of $\mathbb{C}$, \ie whose support is included in $\mathbb{C}^r$ for some $r>0$,   
\begin{equation*}
  \lim_{t\rightarrow +\infty} b(t)\mathbb{E}\left[h(t^{-1}X,Y) \right]=
  \int_{\mathbb{O}} h \ud\mu. %(x,y)\in \mathbb{O}}h(x,y)d\mu(x,y).
\end{equation*}

\item[(iii)]\label{eq:cond_reg_var_appendix4} There exists a finite  measure $\Phi$ on $\mb{S} \times I$ such that
\begin{equation*}\label{eq:cond_reg_var_appendix41}
\frac{\mb{P}\{\theta(X) \in B, Y \in A,  \lVert X\rVert  \s tr \}}{\mb{P}\{\lVert X\rVert  \s t \}} \tti c   r^{-\alpha}\Phi(B \times A)
\end{equation*}
 for all $r>0$ and $A \in \mathcal{B}(I)$, $B\in \mathcal{B}(\mb{S})$  such that $\Phi (\partial (B \times A) ) =0$, with $c = \Phi(\sphere\times I)^{-1}$. 
\end{itemize}

\end{theorem}

\begin{proof}
  \textbf{$\hyperref[eq:cond_reg_var_appendix]{\textit{(i)}}
    \Leftrightarrow
    \hyperref[eq:cond_reg_var_appendix2]{\textit{(ii)}}$.}
  Condition~$\hyperref[eq:cond_reg_var_appendix2]{\textit{(ii)}}$ in
  the statement is precisely Definition 3.2 of regular variation in
  $\Mo$ of~\cite{lindskog2014regularly}, regarding the measure $P$
  restricted to $\mathbb{O}$. The equivalence with our
  Assumption~\ref{hyp:limit} is a direct application of the
  Portmanteau theorem 2.1 in~\cite{lindskog2014regularly}.

  \textbf{$\hyperref[eq:cond_reg_var_appendix4]{\textit{(iii)}}
    \Leftrightarrow
    \hyperref[eq:cond_reg_var_appendix]{\textit{(ii)}}$. } We
  generalize the argument of \cite{lindskog2014regularly}, Example 3.4
  and we verify that we fit into the context of Example 3.5 of the
  same reference. The argument in Example~3.5 (see also Example~3.4)
  in~\cite{lindskog2014regularly} relies on a continuous mapping
  argument (Theorem 2.3 in the same reference). Introduce the `polar
  coordinate transform' $T(x,y) = (\lVert x\rVert , \theta(x), y )$, for
  $(x,y)\in\mathbb{O}$, where we recall $\theta(x) = x/\lVert x\rVert $. Then
  $T$ is a homeomorphism from $\mathbb{O}$ onto
  $\mathbb{O}' = (\rset_+\setminus\{0\})\times\sphere\times I =
  \mathcal{Z}'\setminus\mathbb{C}'$ with
  $\mathcal{Z}' = \rset_+\times\sphere\times I$,
  $\mathbb{C}' = \{0\}\times\sphere\times I$. The space $\mathcal{Z}'$
  is endowed with a continuous scalar multiplication
  $\lambda.(r, \omega, y) = (\lambda r, \omega, y)$ for
  $\lambda\ge 0$, which is compatible with the mapping $T$ in the
  sense that $\lambda.T(z) = T(\lambda.z)$. The scalar multiplication
  on $\mathcal{Z}'$ satisfies the same associativity and monotonicity
  properties as the one on $\mathcal{Z}$.  The mapping $T$ has the
  property that if $A'\subset\mathbb{O}'$ is bounded away from
  $\mathbb{C}'$ then also $T^{-1}(A')\subset\mathbb{O}$ is bounded
  away from $\mathbb{C}$.  The conditions of Example 3.5 in
  \cite{lindskog2014regularly} are thus satisfied, so that regular
  variation of the joint distribution $P$ (restricted to $\mathbb{O}$)
  in $\Mo$ is equivalent to regular variation of the image measure
  $T_\star P$ (restricted to $\mathbb{O}'$), with limit measure
  $\mu' = T_\star \mu$, and with the same scaling function $b(t)$.  In
  other words
  Condition~$\hyperref[eq:cond_reg_var_appendix]{\textit{(ii)}}$ is
  equivalent to the fact that for any measurable sets
  $B\subset\sphere, C\in I$ such that
  $\mu( \partial (\mathcal{C}_B\times C))=0$, where
  $\mathcal{C}_B= \{t \omega, t\ge 1, \omega\in B\}$, we have 
  \begin{align*} 
    &    b(t)\PP[\lVert X\rVert  > tr, \theta(X) \in B , Y\in C  ] \\
   \tti &  \mu\{(x,y): \lVert x\rVert \ge r, \theta(x)\in B, y\in C \} \\ = &  \mu( r.\{ (x,y): \lVert x\rVert \ge 1 , \theta(x)\in B, y\in C  \}) \\
    = &  r^{-\alpha} \mu\{ (x,y): \lVert x\rVert \ge 1 , \theta(x)\in B, y\in C  \}, 
  \end{align*}
  where the last identity follows from the homogeneity of $\mu$
  (Theorem 3.1 in~\cite{lindskog2014regularly}).  Define the angular
  measure $\Phi$ on $\sphere\times I$ as in~\eqref{eq:definePhiJoint}
  from the main paper,
  $\Phi(B\times C) = \mu\{(x,y)\in \mathbb{O} : \lVert x\rVert \ge 1,
  \theta(x)\in B, y\in C\}$. Then $\Phi$ is a finite measure and the
  latter display writes equivalently 
    \begin{align} 
    b(t)\PP[\lVert X\rVert  > tr, \theta(X) \in B , Y\in C  ] \tti r^{-\alpha}\Phi(B\times C), \label{eq:rv-angular} 
  \end{align}
  for all measurable sets $B\subset\sphere, C\in I$ such that $\Phi(\partial(B\times C))= 0$. If~\eqref{eq:rv-angular} holds then also, taking $B=\sphere, C=I, r=1$ we have
  \begin{align*}
     b(t)\PP[\lVert X\rVert  > t  ] \tti  \Phi(\sphere \times I), 
  \end{align*}
  and taking the ratio of~\eqref{eq:rv-angular} with the latter
  displays yields
  Condition~$\hyperref[eq:cond_reg_var_appendix4]{\textit{(iii)}}$ of
  the statement. Conversely if
  $\hyperref[eq:cond_reg_var_appendix4]{\textit{(iii)}}$ holds, then
  letting $b(t)= \Phi(\sphere \times I)/\PP[\lVert X\rVert > t ]$, we
  obtain~\eqref{eq:rv-angular}, which is equivalent to
  Condition~$\hyperref[eq:cond_reg_var_appendix]{\textit{(ii)}}$.
\end{proof}

\section{Proofs of the Results in Section~\ref{sec:background}}\label{sec:proofs_section_proba}
This section gathers the proofs of the claims in
Example~\ref{ex:noise} and auxiliary results for the proof of
Proposition~\ref{prop:exampleMissingComponent}. 
\subsection{Proofs and Additional Results concerning  Example~\ref{ex:noise}}\label{sec:examples}
In this section, we show that a generic heavy-tailed regression model (Example~\ref{ex:noise}) satisfies the requirements of our assumptions. Subsequently, we establish that two widely used models, the additive and multiplicative noise models,  % (see Remark~\ref{rem:examples}),
constitute particular  instances of that  generic model.

\begin{proposition}\label{prop:B}  In the setting of Example~\ref{ex:noise}, 
   the random pair $(X,Y)$ satisfies Assumption~\ref{hyp:bound},~\ref{hyp:limit} and~\ref{hyp:cont_reg_func}. In particular, the  limit distribution $P_\infty$ in  Equation~\eqref{eq:Pinf} 
   is given by
\begin{equation*}
 P_\infty = \mathcal{L}(X_\infty, g_\theta(X_\infty \slash \lVert X_\infty\rVert ,\varepsilon)),
\end{equation*}
 where $X_\infty$ follows the limit distribution
  \begin{equation*}
    Q_\infty = \lim_{t\to+\infty} \mathcal{L}(t^{-1}X \mid \lVert X\rVert  \s t ).  
      \end{equation*}
\end{proposition}
\begin{proof}
  Assumption~\ref{hyp:bound} is obviously fulfilled with
  $M=\sup_{x,z\in \mb{R}^d \times \mb{R}}| g(x,z) |$. Regarding
  Assumption~\ref{hyp:limit} and the limit distribution, we consider a
  bounded and Lipschitz function
  $l:\mb{R}^d\times \mb{R}\rightarrow \mathbb{R}$.  For all $t>0$,
  writing $\Theta = \lVert X\rVert ^{-1} X$, we have
\begin{align*}
  \mathbb{E}\left[ l(t^{-1}X,Y) \mid \lVert X\rVert  \s t \right]
  &=\mathbb{E}\left[ l(t^{-1}X,g(X,\varepsilon)) \mid \lVert X\rVert  \s t \right] \nonumber \\
  &= \mathbb{E}\left[ l(t^{-1}X,g_\theta(\Theta,\varepsilon)) \mid \lVert X\rVert  \s t \right]    \\
  &+  \mathbb{E}\left[ l(t^{-1}X,g(X,\varepsilon))  - l(t^{-1}X,g_\theta(\Theta,\varepsilon)) \mid \lVert X\rVert  \s t \right]. % - \mathbb{E}\left[ l(t^{-1}X,g_\theta(\Theta,\varepsilon)) \mid \|X\| \s t \right].
\end{align*}
Since $\varepsilon$ is independent from $X$, writing $\Theta_\infty = \lVert X_\infty\rVert ^{-1} X_\infty$, the regular variation of $X$ and continuity of $l$ and $g_\theta$ imply that 
\begin{equation}\label{eq:26}
\mathbb{E}\left[ l(t^{-1}X,g_\theta(\Theta,\varepsilon)) \mid \lVert X\rVert  \s t \right] \rightarrow \mathbb{E}\left[ l(X_\infty,g_\theta(\Theta_\infty,\varepsilon))\right].
\end{equation}
Because $l$ is Lipschitz continuous (for some Lipschitz  constant $C$) and $X$ and $\varepsilon$ are independent, we have
\begin{align*}
  \Big| \mathbb{E}&\left[ l(t^{-1}X,g(X,\varepsilon))  -   l(t^{-1}X,g_\theta(\Theta,\varepsilon))\mid \lVert X\rVert  \s t \right] \Big|    % \mathbb{E}\left[ l(t^{-1}X,g_\theta(\Theta,\varepsilon)) \mid \lVert X\rVert  \s t \right] \Big|  
                                                       \\
  & \le C \E \Big[| g(X,\varepsilon) - g_\theta(\Theta,\varepsilon) | \, \big| \lVert X\rVert  \s t\Big]\\
 & \le C  \E \Big[ \sup_{\lVert x\rVert  \s t} |g(x,\varepsilon) - g_\theta(\theta(x), \varepsilon)| %\,   \big| \,    \lVert X\rVert  \s t
  \Big]. \label{eq:27} %
%\rightarrow 0,
\end{align*}
The right-hand side tends to zero as  $t \rightarrow +\infty$, % where the convergence is
from  the dominated convergence theorem which applies because $\sup_{\lVert x\rVert  \s t} |g(x,\varepsilon) - g_\theta(x/ \lVert x\rVert , \varepsilon) | \m M$ and because of our model assumption~\eqref{eq:condition_example}.   % Equation~\eqref{eq:condition_appendix3} hold. Finally
Thus Assumption~\ref{hyp:limit} is satisfied  and % we have, with notation of Section~\ref{subsec:framework},  % Equation~\eqref{eq:Pinf},
% $(X_\infty, Y_\infty) \overset{d}{=}
$P_\infty =\mathcal{L}(X_\infty, g_\theta(\Theta_\infty,\varepsilon))$.

We now show that  Assumption~\ref{hyp:cont_reg_func} also holds true by proving the stronger condition (i) from Proposition~\ref{prop:sufficientConditionsForAssum3}. 
For  $x \in \rset^d$ with $\lVert x\rVert  \s t$, we have by independence of $X$ and $\varepsilon$, 
\begin{align*}
|f^*(x) - \bayesext(\theta(x)) | & = \Big| \E [ g(x,\varepsilon)] - \E [ g_\theta(\theta(x),\varepsilon)] \Big| \\ 
&\m \EE \Big[\sup_{\lVert x\rVert  \s t} \big|g(x,\varepsilon) - g_{\theta}(\theta(x),\varepsilon) \big|\Big],
\end{align*}
which entails as in~\eqref{eq:26} that $\sup_{\lVert x\rVert  \s t} |f^*(x) - \bayesext(x / \lVert x\rVert ) | \rightarrow 0$, as $t \rightarrow +\infty$. Since $g$ is assumed continuous and bounded, $f^*$ is continuous. Thus, the sufficient condition (i) from Proposition~\ref{prop:sufficientConditionsForAssum3} is satisfied, which shows that 
 Assumption~\ref{hyp:cont_reg_func} holds true. 
\end{proof}

We now turn to the two sub-examples given by the additive and
multiplicative noise models mentioned after Example~\ref{ex:noise}
from the main paper. We show that under mild assumptions, both
sub-examples indeed satisfy the conditions specified in
Proposition~\ref{prop:B}.

\begin{proposition}\label{ex:add_noise} Consider the additive noise model
\begin{equation*}
Y=\Tilde{g}(X)+\varepsilon,
\end{equation*}
where $X$ is a regularly varying random vector in $\rset^d$ such that 
\begin{equation*}
\mathcal{L}(t^{-1}X \mid \lVert X\rVert  \s t ) \rightarrow \mathcal{L}(X_\infty),
\end{equation*}
as $t \rightarrow + \infty$, $\varepsilon$ is a bounded real-valued
random variable defined on the same probability space independent from
$X$ and $\Tilde{g}_\theta$ % :\mb{R}^d \rightarrow \mathbb{R}
is a bounded, continuous function on $\rset ^d$ which converges
uniformly to some angular mapping
$\Tilde{g}_\theta:\sphere \rightarrow \mb{R}$, in the sense that
\begin{equation*}\label{eq:condition_appendix}
\sup_{\lVert  x\rVert  \s t}| \Tilde{g}(x)- \Tilde{g}_\theta(\theta(x))|\rightarrow 0 \text{ as } t\rightarrow +\infty.
\end{equation*}
Then, the random pair $(X,Y)$ satisfies the requirements of
Proposition~\ref{prop:B} with
$M=\sup_{x\in \mb{R}^d}\vert \Tilde{g}(x) \vert+\vert\vert \varepsilon
\vert\vert_{\infty}$. The limit distribution $P_\infty$
in % random variables involved in the limit in
Equation~\eqref{eq:Pinf} is
\begin{equation*}
P_\infty = \mathcal{L}\big(X_\infty, \Tilde{g}_\theta( \theta(X_\infty)) + \varepsilon \big).
\end{equation*}
\end{proposition}

\begin{proof}
  Because $\varepsilon$ is almost surely bounded, there exists
  $m_\varepsilon \in \rset_+$ a nonnegative real-number such that
  $\varepsilon \overset{a.s.}{\in}
  [-m_\varepsilon,+m_\varepsilon]$. Consider the mapping
  $g : (x,z) \in \rset^d \times [-m_\varepsilon,+m_\varepsilon]
  \mapsto g(x)+z$ and
  $g_\theta : (\omega,z) \in \sphere \times
  [-m_\varepsilon,+m_\varepsilon] \mapsto
  \Tilde{g}_\theta(\omega)+z$. The function $g$ is continuous and
  bounded by
  $M=\sup_{x\in \rset^d}| \Tilde{g}(x) | + m_\varepsilon$  %\| \varepsilon \|_{\infty}$
  and the pair $(g,g_\theta)$ satisfies
  Equation~\eqref{eq:condition_example}.  % \eqref{eq:condition_appendix3}
  % with $g_\theta$ 
  Indeed  for all $z \in [-m_\varepsilon,+m_\varepsilon]$, 
\begin{equation*}
\sup_{\lVert x\rVert  \s t} |g(x,z)- g_\theta(\theta(x),z)| = \sup_{\lVert x\rVert  \s t}|\Tilde{g}(x) - \Tilde{g}_\theta(\theta(x))| \rightarrow 0,
\end{equation*}
as $t \rightarrow + \infty$, which concludes the proof. 
\end{proof}

\begin{proposition}\label{ex:multiplicativeNoise} Consider the multiplicative noise model
\begin{equation*} \label{eq:noise2}
Y=\varepsilon \Tilde{g}(X),
\end{equation*}
where $(X, \varepsilon)$ and
$\Tilde{g}$ are as in
Proposition~\ref{ex:add_noise}. % is a regularly varying random vector in $\rset^d_+$ such that
Then, the random pair
$(X,Y)$ satisfies the requirements of Proposition~\ref{prop:B} with
$M=\sup_{x\in \rset^d}| \tilde g(x) | \times \| \varepsilon
\|_{\infty}$ and
% limit random variables involved in Equation
the limit distribution
$P_\infty$ in~\eqref{eq:Pinf} is given by $P_\infty =
\mathcal{L}(X_\infty, \varepsilon
\Tilde{g}_\theta(\theta(X_\infty)))$, where
$\Tilde{g}_\theta$ and
$X_\infty$ are as in Proposition~\ref{ex:add_noise}.
\end{proposition}
\begin{proof}
  Consider the mapping $g(x,z) = z
  \Tilde{g}(x)$ and $g_\theta(\omega,z) =
  z\Tilde{g}_\theta(\omega)$. Let
  $m_\varepsilon$ be as in the proof of
  Proposition~\ref{ex:add_noise}.  On the domain $\rset^d\times
  [-m_\varepsilon, m_\varepsilon]$, the function
  $g$ is continuous and bounded by $M= m_\varepsilon \sup_{x\in
    \rset^d}| \Tilde{g}(x) | $.  %\varepsilon
  % \|_{\infty}$
  The pair $(g,g_\theta)$ satisfies~\eqref{eq:condition_example} %  Equation
  % \eqref{eq:condition_appendix3} % with $g_\theta$
  since for all
  $z \in [-m_\varepsilon,+m_\varepsilon]$
\begin{equation*}
  \sup_{\lVert x\rVert  \s t} |g(x,z)- g_\theta(x/\lVert x\rVert ,z)| \le  m_\varepsilon  \sup_{\lVert x\rVert  \s t}|\Tilde{g}(x) - \Tilde{g}_\theta(\theta(x))| \xrightarrow[t\to\infty]{} 0,
\end{equation*}
%as $t \rightarrow + \infty$,%
which concludes the proof. 
\end{proof}

\subsection{Auxiliary results for the proof of
  Proposition~\ref{prop:exampleMissingComponent}} % concerning Example~\ref{ex:predictionRVvect}}
 \label{sec:proof_example2}

 \begin{lemma}[Uniform Convergence of marginals of $p$. ]\label{lem:Unifcvmarginal}~
   Under the assumptions of
   Example~\ref{ex:predictionRVvect}, % for any $\tilde x\neq 0_{\rset^d}$
   we have
  \begin{equation*}
    \begin{gathered}
      \sup_{\lVert  x \rVert  = 1} \left| \int_\rset t^{-1} h(t)  p(t x, z) \ud z - q(x) \right| \tti 0, \quad \text{where } \\q( x) = \int_z q(x,z) \ud z,  \quad  \text{ and } h(t) = t^{d+1}/\PP[\lVert \tilde X\rVert \s t].     
    \end{gathered}
 \end{equation*}

%  with  $h(t) = t^{d+1}/\PP[\|X\|>t]$. 
\end{lemma}
\begin{proof}
  We adapt the arguments of the proof of Theorem 2.1 of
  \cite{de1987regular} to our context. With the notation $h$ from our
  statement, our uniform convergence assumption~(\ref{eq:cvDensity})
  becomes
  \begin{equation*}
    \sup_{\omega\in\sphere_{d+1}} | h(t)p(t\omega) - q(\omega)|\tti 0. 
  \end{equation*}
  Now
  \begin{align*}
    \int_\rset t^{-1} h(t)  p(tx, z) \ud z
    % &= \frac{h(t)}{h(t\lVert \tilde x\rVert )} \int_{\rset } h(t\lVert \tilde x\rVert )
    % p(t\tilde x, tr  ) t\ud r \\
    & = \int_\rset h(t)  p(tx, t r )  \ud r,  \\
    \end{align*}
    so that
    \begin{align*}
      \sup_{\lVert  x \rVert  = 1} \left| \int_\rset t^{-1} h(t)  p(tx, z) \ud z - q(x) \right|
      & \le \int_\rset \sup_{\lVert  x \rVert  = 1} \left|
         h(t)  p(tx, tr ) -  q(x,r ) \right|\ud r. 
    \end{align*}
    For fixed $r\in\rset$, because $\lVert (x,r)\rVert  \ge \lVert x \rVert \ge 1$, the integrand in the right-hand side is less than
    \begin{align*}
      \sup_{\lVert \tilde u\rVert \ge 1} \left|
         h(t)  p(t \tilde u  ) -  q(\tilde u) \right|. 
    \end{align*}
    The latter display tends to zero as $t\to +\infty$ because of~(\ref{eq:cvDensitybis}). To conclude, we need to upper bound the integrand by an integrable function of $r$, in order to apply dominated convergence. We thus write 
    \begin{align*}
      \sup_{\lVert  x \rVert  = 1} &\left|
      h(t)  p(t x, tr ) -  q( x,r ) \right|
                                     \\
      & \le \sup_{\lVert  x \rVert  = 1}   h(t)  p(t x, tr ) + 
        \sup_{\lVert   x \rVert  = 1}  q( x,r )   \\
        & =   \underbrace{  \sup_{\lVert   x \rVert  = 1} \frac{h(t)}{h(t\, \lVert  (x,r ) \rVert )
          }}_{A(t,r)}
        \underbrace{  \sup_{\lVert  x \rVert  = 1} h(t \, \lVert ( x,r ) \rVert ) 
           p\Big(t \, \lVert  ( x,r)  \rVert  \;  \theta( x,r)\Big)}_{B(t,r)}   +
           \underbrace{\sup_{\lVert  x \rVert  = 1}  q(x,r )}_{C(t,r)},
  \end{align*}
  where $\theta(x,r)\in\sphere_{d+1}$. % We now show that a domination condition is satisfied regarding the product $A(t,r) B(t,r)$ in order to apply dominated convergence. 
  \begin{itemize}
  \item 
  The function $h$ is regularly varying with positive index $d+1+\alpha$.
  By Karamata representation (Proposition~0.5 of \cite{resnick2013extreme}), for $t$ large enough (say $t\ge t_0$),
  for any $s\ge 1  $, we have
\begin{equation*}
\frac{h(t)}{h(ts)} \le 2 s^{-d-\frac{\alpha}{2}+1}.
\end{equation*}

Thus for $t\ge t_0$, for all $r\in\rset$,
\begin{equation}
  \label{eq:boundA_Karamata}
A(t,r) \le 2 \lVert ( x,r)\rVert ^{-d-\alpha/2-1} \le 2 (1+r^p)^{\frac{-d-\alpha/2-1}{p}},   
\end{equation}

which is an integrable function of $r$ for any $d\ge 1,\alpha>0$. 
% for some $\epsilon>0$ which depends on the (fixed) value of $\|\tilde x\|$. 
\item because $\lVert ( x,r) \rVert \ge \lVert  x\rVert  \ge 1$ we have for all $t \ge t_0$ large enough, uniformly over $ x$ such that  $\lVert  x\rVert =1$ and $ r\in\rset$, 
  \begin{align*}
    % B(t,r)
    % &\le  \sup_{u\ge 1 }
   \Big|   h(t \, \lVert  ( x,r ) \rVert )
      p\Big(t \, \lVert  ( x,r ) \rVert  \;  \theta( x,r)\Big) -  q\Big(\theta(x,r)\Big) \Big| \le 1,
  \end{align*}
  thus for $t\ge t_0$, for all $r$, 
  \begin{align*}
    B(t,r) \le \sup_{\omega \in \sphere_{d+1}}q(\omega) + 1, 
  \end{align*}
which is a finite constant. 
% The $B(t,r)$ converges uniformly as $t\to\infty$ to $q(\theta(r))$ which is bounded on $\sphere$,  thus for $t$ large enough, $B(t,r)\le C$ for some constant $C$.
\item We may also upper bound $C(t,r)$ by an integrable function of $r$, since by homogeneity of $q$, 
  \begin{equation*}
    C(t,r)
    =  \sup_{\lVert  x\rVert  = 1}  \lVert (x,r)\rVert ^{-d-\alpha-1}q\big(\theta(x,r)\big) 
     \le \max_{\omega \in \sphere_{d+1}} (q(\omega)) (1+r^p)^{\frac{{-d-\alpha-1}}{p}}, 
  \end{equation*}
  which is integrable for $d\ge 1$ and $\alpha>0$. 
\end{itemize}
As a consequence of the above three points, the quantity
$A(t,r) B(t,r) + C(t,r)$ is upper bounded by an integrable function of
$r$. The result follows by dominated convergence.
 \end{proof}
 \begin{lemma}[Upper and lower bounds for the marginals of $q$]\label{lem:boundsMarginalQ}
   Under the conditions of Example~\ref{ex:predictionRVvect}, there
   exists positive constants $c,C >0$ such that for all $x\in\rset^d$
   such that $\lVert x\rVert =1$,
   \begin{align*}
    c \le  \int q(x,z)\ud z \le C.
   \end{align*}
 \end{lemma}
 \begin{proof}
   For $x\in\rset^d$ such that $\lVert x\rVert  =1$, and $z\in\rset$ we have
\begin{equation*}
q(x,z) = (1+z^p)^{\frac{-\alpha-d-1}{p}}q(\theta(x,z)).
\end{equation*}
  
The results follows with
\begin{equation*}
  c = (\min_{\omega \in \sphere_{d+1}} q(\omega)) \; \int (1+z^p)^{\frac{-\alpha-d-1}{p}} \ud z \text{ and } C = (\max_{\omega \in \sphere_{d+1}} q(\omega) ) \; \int (1+z^p)^{\frac{-\alpha-d-1}{p}} \ud z.
\end{equation*}
 \end{proof}

\section{Proofs of the Results in Section~\ref{sec:main}}\label{sec:proofs_main}
\subsection{Proof of Theorem~\ref{prop:conv_bayes_risk}}\label{sec:proof_thm_32}

\hspace{\parindent}(i) In view of Characterization (iii) from
Theorem~\ref{thm:limit} (see also~\eqref{eq:cond_reg_var3}),
Assumption~\ref{hyp:limit} implies that the conditional distribution
  $$\mathcal{L}(\Theta, Y,\lVert X\rVert /t \;|\; \lVert X\rVert >t) $$ converges weakly
  to the distribution of
  $(\Theta_\infty,Y_\infty, \lVert X_\infty\rVert )$.  Now if
  $f = h\circ \theta$ is a prediction function on $\rset^d$, where $h$
  is a continuous function defined on $\sphere$, then by compactness
  of $\sphere$ the function $(\theta,y)\mapsto (h(\theta)-y)^2 $ is
  automatically bounded and continuous on the domain
  $\sphere\times[-M,M]$.  Thus by weak convergence we obtain as
  $t\to+\infty$,
  \begin{equation*}
  R_t(f) = \EE[ (h(\Theta) - Y)^2 \,|\, \lVert X\rVert >t ]     \to  \EE[ (h(\Theta_\infty) - Y_\infty)^2] = R_{P_\infty}(f). 
  \end{equation*}

  \hspace{\parindent}(ii) Recall that $R_t^* = R_t(f^*)$ where $f^*$
  is the regression function for the pair $(X,Y)$ and
  $\riskext^* = \riskext(\bayesext)$ where $\bayesext$ is the
  regression function for the pair $(X_\infty,Y_\infty)$ defined in
  Lemma~\ref{lem:angularMinimizer}. Now we decompose $R_t^*$ as
 %   Recall that $\bayesext(tx) = \bayesext(x)$. 
    \begin{align*}
      R_t^{*}
      &= \EE[ (Y - f^*(X))^2 \,|\, \lVert X \rVert \s t] \\
      &= \underbrace{\EE[ (Y - \bayesext(X))^2 \,|\, \lVert X\rVert \s t]}_{A_t}  +
        \underbrace{\EE[ ( \bayesext(X) -f^*(X) )^2 \,|\, \lVert X\rVert \s t]}_{B_t} \\
      &+  
        \underbrace{ 2\EE[ (Y - \bayesext(X))(\bayesext(X) - f^*(X) ) \,|\, \lVert X\rVert \s t]}_{C_t} .\\
    \end{align*}
    The first term $A_t$ is simply $R_t(\bayesext)$. From
    Lemma~\ref{lem:angularMinimizer}, $\bayesext$ is an angular
    function, thus Property (i) of the statement implies that
    $A_t\to \riskext(\bayesext)$, which is $\riskext^*$.
    
    We now show that the second and third terms $B_t,C_t$ vanish. We
    use that, as a consequence of Assumption~\ref{hyp:bound},
    $\forall x \in \rset^d$, $|\bayesext(x)| \le M$ and,
    $|f^*(x)| \le M$. Thus
\begin{equation*}
  B_t \le 4M^2 \EE[ | \bayesext(X) -f^*(X)|  \,|\, \lVert X\rVert \s t ]. % \tti 0, 
\end{equation*}
Assumption~\ref{hyp:cont_reg_func} ensures that the latter display
converges to $0$ as $t\to \infty$. %to obtain the convergence to $0$.
Similarly, using Assumptions~\ref{hyp:bound}
and~\ref{hyp:cont_reg_func} again, we obtain
\begin{equation*}
  | C_t | \le 4M^2 \EE[ | \bayesext(X) -f^*(X)|  \,|\, \lVert X\rVert \s t ] \tti 0.
\end{equation*}
We have proved that $R_t^* \tti \riskext^*$.

\hspace{\parindent}(iii) Recall from the introduction that
$\riskinf^* = \riskinf(f^*) = \limsup_t R_t(f^*)$. Because of (ii), in
fact $R_t(f^*)$ converges to $\riskext^*$. Thus
  $$ \limsup_t R_t(f^*) = \lim_t R_t(f^*) = \riskext^*, $$ and the result follows.
 
  \hspace{\parindent}(iv) From Property (iii) of the statement, we
  have $\riskinf^* =\riskext(\bayesext)$. Now, Property (i) of the
  statement and the angular nature of $\bayesext$
  (Lemma~\ref{lem:angularMinimizer}) imply that
  $\riskext(\bayesext) = \riskinf(\bayesext)$.
%    \end{itemize}

\subsection{Proof of Proposition~\ref{prop:concentr1}}\label{sec:D}
We recall for convenience a Bernstein-type inequality due to
C. McDiarmid (see Theorem~3.8 of~\cite{mcdiarmid1998concentration})
which is a key ingredient of the proof of
Proposition~\ref{prop:concentr1}.
\begin{lemma}[Bernstein-type inequality, \cite{mcdiarmid1998concentration}] \label{lem:Bernstein}
Let $X=(X_{1:n})$ with $X_i$  taking  values in a set $\mathcal{X}$ and let $f$ be a real-valued function defined on $\mathcal{X}^n$. Let $Z = f(X_{1:n})$. Consider the positive deviation functions, defined for $1 \le i \le n$ and for $x_{1:i} \in \mathcal{X}^i$, 
\begin{equation*}
g_i(x_{1:i}) = \E [Z | X_{1:i}=x_{1:i}] - \E [Z | X_{1:i-1}=x_{1:i-1}].
\end{equation*}
Denote by $b$ the maximum deviation 
\begin{equation*}
b = \max_{1 \le i \le n} \sup_{x_{1:i} \in \mathcal{X}^i} g_i(x_{1:i}).
\end{equation*}
Let $\hat{v}$ be the supremum of the sum of conditional variances, % defined by
\begin{equation*}
\hat{v} = \sup_{(x_1,...,x_n) \in \mathcal{X}^n} \sum_{i=1}^n \sigma_i^2(f, x_{1:i-1}),
\end{equation*}
where $\sigma_i^2(f, x_{1:i-1})) = \Var[g_i(X_{1:i}) \,|\, X_{1:i-1} = x_{1:i-1}]$.
If $b$ and $\hat{v}$ are both finite, then 
\begin{equation*}
\mb{P}\{Z - \EE[Z] \ge \varepsilon \} \le \exp\bigg(\frac{-\varepsilon^2}{2(\hat{v}+b\varepsilon/3)}\bigg),
\end{equation*}
for $\varepsilon > 0$.
\end{lemma}

We now proceed with the proof of Proposition~\ref{prop:concentr1}
Introduce an intermediate risk functional
\begin{equation*}
\risknktilde(h \circ \theta) = \frac{1}{k} \sum_{i=1}^n\Big(h(\theta(X_i))-Y_i\Big)^2 \1 \{ \lVert X_i\rVert  \ge t_{n,k}\},
\end{equation*}
and notice that $\EE[\risknktilde(h\circ\theta)] = \risknk(h\circ\theta)$. 
Our proof is based on  the following risk decomposition, %in order to separate the study in two parts
\begin{equation}\label{eq:decomp_prop2}
  \begin{aligned}
    \sup_{h\in \mathcal{H}}&\Big|   \riskempnk(h\circ \theta)-\risknk(h\circ \theta)\Big| \\
   &  \le \sup_{h\in \mathcal{H}}\Big|   \riskempnk(h\circ \theta)-\risknktilde(h\circ \theta)\Big| +
    \sup_{h\in \mathcal{H}}\Big|  \risknktilde(h\circ \theta)-\risknk(h\circ \theta)\Big|.
    \end{aligned}
\end{equation}

Regarding the first term on the right-hand side of Inequality \eqref{eq:decomp_prop2}, % the triangle inequality yields
\begin{align*}
  \sup_{h\in \mathcal{H}} &\Big| \riskempnk(h\circ \theta)  -\risknktilde(h\circ \theta)\Big| \\
  &= \sup_{h\in \mathcal{H}}\frac{1}{k} \Big| \sum_{i=1}^n \Big(h\circ\theta(X_i) -Y_i\Big)^2 \big(\1 \{ \lVert X_i\rVert  \ge t_{n,k}\} - \1 \{ \lVert X_i\rVert  \ge \lVert X_{(k)}\rVert \}\big)\Big|  \\
  &\le \frac{4M^2}{k}  \sum_{i=1}^n \big|\1 \{ \lVert X_i\rVert  \ge t_{n,k}\} - \1 \{ \lVert X_i\rVert  \ge \lVert X_{(k)}\rVert \}\big|.
\end{align*}
The number of nonzero terms inside the sum in the above display is the number of indices $i$ such that ` $\lVert X_i\rVert  < \lVert X_{(k)}\rVert $ and $\lVert X_i\rVert  \ge t_{n,k}$', or the other way around.  In other words 
\begin{align*}
  \Big\{\big|
  &\1 \{ \lVert X_i\rVert  \ge t_{n,k}\}
    - \1 \{ \lVert X_i\rVert  \ge \lVert X_{(k)}\}\big| \neq 0 \Big\} \\
  &\subset 
    \Big(  \big\{t_{n,k}\le X_i < X_{(k)}  \big\} \cup
    \big\{  X_{(k)} \le X_i <  t_{n,k} \big\} \Big). 
\end{align*}
Considering separately the cases where  $X_{(k)}\le t_{n,k}$ and
$X_{(k)}>t_{n,k}$ we obtain 
\begin{align*}
  \sup_{h\in \mathcal{H}}\Big| \riskempnk(h\circ \theta)-
  \risknktilde(h\circ \theta)\Big|
  & \le \frac{4M^2}{k}\Big| \sum_{i=1}^n  \1\{ \lVert X_i\rVert  \ge t_{n,k} \} -  k \Big|. 
\end{align*}
Notice that $\sum_{i=1}^n \1\{ \lVert X_i\rVert \ge t_{n,k} \}$
follows a Binomial distribution with parameters $(n, k/n)$.  The
(classical) Bernstein inequality as stated e.g., in
\cite{mcdiarmid1998concentration}, Theorem~2.7, yields
\begin{align}
  \mb{P} \Big\{ &\sup_{h\in \mathcal{H}}\Big| \riskempnk(h\circ \theta)-\risknktilde(h\circ \theta)\Big| \ge \varepsilon \Big\} \nonumber \\
                &  \le \mb{P}\Big\{ \Big| \sum_{i=1}^n  \1\{ \lVert X_i\rVert  \ge t_{n,k} \} -  k \Big| \ge k\varepsilon / (4M^2) \Big\} \nonumber \\
                & \le 2 \exp \Big( \frac{-k\varepsilon^2}{32M^4+8M^2\varepsilon/3} \Big).  \label{eq:1}
%&\le 2 \exp \Big( \frac{-k\varepsilon^2}{64M^4+8M^2\varepsilon/3} \Big)
\end{align}

We now  turn to the second term of Inequality~\eqref{eq:decomp_prop2}, and we apply
Lemma~\ref{lem:Bernstein} to the function
$$f((x_1,y_1),...,(x_n,y_n)) =
\sup_{h \in \mathcal{H}} \Big| \frac{1}{k}\sum_{i=1}^n\Big(h\circ
\theta(x_i) - y_i \Big)^2 \1 \{ \lVert x_i\rVert \ge t_{n,k} \} -
\risknk(h \circ \theta) \Big|, $$ so that
$f( (X_1,Y_1),...,(X_n,Y_n)) = \sup_{h\in\mathcal{H}} \big|
\risknktilde(h\circ\theta) - \risknk(h\circ\theta)\big|$.  With the
notations of Lemma~\ref{lem:Bernstein}, the maximum of the positive
deviations and the maximum sum of variances satisfy respectively
$b \le 4M^2/k$ and $\hat{v} \le 16M^4/k$. Thus
\begin{equation}\label{eq:2}
  \begin{gathered}
    \mb{P}\Big\{\sup_{h\in \mathcal{H}}\Big|  \risknktilde(h\circ \theta)-\risknk(h\circ \theta)\Big|  - \EE [\sup_{h\in \mathcal{H}}\Big|  \risknktilde(h\circ \theta)-\risknk(h\circ \theta)\Big| ] \ge \varepsilon \Big\} \\
    \le  \exp \Big( \frac{-k \varepsilon^2}{32M^4+8M^2\varepsilon /3}\Big).     
  \end{gathered}
\end{equation}

The last step consists in bounding from above the expected deviations in the above display, that is 
\begin{equation*}
\EE \sup_{h\in \mathcal{H}}\Big|  \risknktilde(h\circ \theta)-\risknk(h\circ \theta)\Big|.
\end{equation*}

%To do so, we follow the lines of proofs of Lemma~13 and Lemma~14 in \cite{goix2015learning}. 
Let $\varepsilon_1,\ldots,\varepsilon_n$ be $n$ independent,
$\{0,1\}$-valued Rademacher random variables and introduce the
Rademacher average
\begin{equation*}
  % \risknktilde^\varepsilon
  \mathcal{R}_k^\varepsilon
  = \sup_{h \in \mathcal{H}}\frac{1}{k}
  \Big| \sum_{i=1}^n \varepsilon_i (h\circ\theta(X_i)-Y_i)^2
  \1 \{ \lVert X_i\rVert  \ge t_{n,k} \} \Big|.
\end{equation*}
Following a standard symmetrization argument as \emph{e.g.} in the
proof of Lemma~13 in
\cite{goix2015learning}, 
we obtain
\begin{equation}
  \EE \sup_{h\in \mathcal{H}}\Big|  \risknktilde(h\circ \theta)-\risknk(h\circ \theta)\Big|  \le 2\EE [  \mathcal{R}_k^\varepsilon ].\label{eq:boundExpDev-Rademacher}
\end{equation}
Let $(X_1^k,Y_1^k),...,(X_n^k,Y_n^k)$ be independent replicates, also
independent from the $X_i,Y_i$'s, such that
$\mathcal{L}(X_i^k,Y_i^k) = \mathcal{L}\big((X,Y) \,|\, \lVert X\rVert
\ge t_{n,k} \big)$. By Lemma~2.1 of \cite{lhaut2022uniform}, we have
\begin{equation*}
  \sum_{i=1}^n \varepsilon_i (h\circ\theta(X_i)-Y_i)^2 \1 \{ \lVert X_i\rVert  \ge t_{n,k} \}
  \overset{d}{=}\sum_{i=1}^\mathcal{K} \varepsilon_i (h\circ\theta(X_i^k)-Y_i^k)^2,
\end{equation*}
where $\mathcal{K} \sim Bin(n,k/n)$ is independent from the $\varepsilon_i,X_i,Y_i$'s. Then,  write
\begin{equation}
  \EE[\mathcal{R}_k^\epsilon]= \frac{1}{k}
  \mb{E} \bigg[  \mb{E}\Big[  \sup_{h \in \mathcal{H}}\big| \sum_{i=1}^\mathcal{K}\varepsilon_i (h\circ\theta(X_i^k)-Y_i^k)^2 \big|  \,\Big|\, \mathcal{K}\Big]\bigg].
   \label{eq:boundRademacher-conditioning}
\end{equation}

We first control the conditional expectation in the above display for
any fixed value $\mathcal{K} = m \le n$. For this purpose, we apply
Proposition~2.1 of \cite{gine2001consistency} to the class of
functions
${\mathcal{G}} = \{g(x,y) = (h\circ\theta(x) - y)^2,
h\in\mathcal{H}\}$.

Notice first that for $g_i(x,y) = (h_i\circ\theta(x) - y)^2, i=1,2$ and $Q$ any probability measure on $\rset^d\times[-M,M]$ we have
\begin{align*}
  & \lVert g_1-g_2\rVert _{L^2(Q)}  \\
  =&\sqrt{\EE_Q\bigg([ ( h_1\circ\theta(X) - h_2\circ\theta(X)) (h_1\circ\theta(X)+h_2\circ\theta(X) - 2Y)]^2\bigg)}\\
  \le &  4M \lVert h_1-h_2\rVert _{L^2(Q_X\circ\theta^{-1})}, 
\end{align*}
where $Q_X$ is the marginal distribution of $Q$ regarding the first
component $X\in\rset^d$. Thus the covering number
$\mathcal{N}(\mathcal{G}, L_2(Q),\tau)$ for the class $\mathcal{G}$,
relative to any $L_2(Q)$ radius $\tau$ is always less than than
$\mathcal{N}(\mathcal{H}, L_2(\tilde Q),\tau/(4M))$ for the class
$\mathcal{H}$, where $\tilde Q = Q_X\circ\theta^{-1}$.  Now the class
$\mathcal{H}$ has envelope function $H = M\1_{\sphere}(\,\cdot\,)$ and
has VC-dimension $V_{\mathcal{H}}<\infty$, thus Theorem 2.6.7
in~\cite{vandervaartWeakConvergenceEmpirical1996} yields a control of
its covering number,
$$\mathcal{N}(\mathcal{H}, L_2(\tilde Q),\tau M ) \le (
A/\tau)^{2V_{\mathcal{H}}}$$ for some universal constant $A>0$
not depending on $\tilde Q$ nor $\mathcal{H}$. We obtain
\begin{equation*}
  \mathcal{N}(\mathcal{G}, L_2(Q),\tau ) \le ( 4AM^2 /\tau)^{2V_{\mathcal{H}}}. 
\end{equation*}
Now $\mathcal{G}$ has envelope function $G = 4M^2\1_{\rset^d\times \sphere}$. The previous display writes equivalently
\begin{equation}
  \mathcal{N}(\mathcal{G}, L_2(Q),\tau \lVert G\rVert _{L^2(Q)} ) \le ( A /\tau)^{2V_{\mathcal{H}}}. \label{eq:controlCovering}
\end{equation}
 Inequality~\eqref{eq:controlCovering} is precisely the  first step of the proof of 
% An inspection of the proof of
 Proposition 2.1 in~\cite{gine2001consistency} (see Inequality 2.2 in
 the cited references), so that their upper bound on the Rademacher
 process applies with VC constant $v = 2 V_{\mathcal{H}}$. The upper
 bound of their statement involves
 $\sigma^2= \sup_g \EE g^2 \le 16M^4$ and
 $U=\sup_g \lVert g\rVert _\infty\le 4M^2$, thus we may take
 $\sigma=U = 4M^2$.  We obtain
\begin{equation*}
  \EE \sup_{h \in \mathcal{H}} \Big| \sum_{i=1}^m\varepsilon_i (h\circ\theta(X_i^k)-Y_i^k)^2 \Big|  \le C' 4M^2(V_{\mathcal{H}} + \sqrt{ m V_{\mathcal{H}}}),  
\end{equation*}
for some other   universal constant  $C'$. Injecting the latter control into~\eqref{eq:boundRademacher-conditioning} yields, using the concavity of the squared root function and $\EE[\mathcal{K}] = k$, 
\begin{equation}
  \EE[\mathcal{R}_k^\varepsilon]\le \frac{1}{k}
    C' 4M^2(V_{\mathcal{H}} + \EE\,[ \sqrt{ \mathcal{K} }]\sqrt{V_{\mathcal{H}}})    
  \le \frac{1}{k}
  C' 4M^2(V_{\mathcal{H}} +  \sqrt{ k }\sqrt{V_{\mathcal{H}}})    .  \label{eq:controlRademacherExtreme}
\end{equation}
Combining~\eqref{eq:boundExpDev-Rademacher}
and~\eqref{eq:controlRademacherExtreme} we obtain
\begin{align}\label{eq:esp}
\EE \sup_{h\in \mathcal{H}}\Big|  \risknktilde(h\circ \theta)-\risknk(h\circ \theta)\Big|  \le 2\EE [\mathcal{R}_k^\epsilon] &  
  \le C4M^2\Big(\frac{V_{\mathcal{H}}}{k}+ \sqrt{\frac{V_{\mathcal{H}}}{k}}\Big),
\end{align}
with $C = 2 C'$. Finally, combining Equations~\eqref{eq:1}, \eqref{eq:2} and \eqref{eq:esp} yields
\begin{equation*}
  \begin{gathered}
\mb{P} \Big\{ \sup_{h\in \mathcal{H}}\Big| \riskempnk(h\circ \Theta)-\risknk(h\circ \Theta) \Big| \ge \varepsilon + C4M^2\Big(\frac{V_{\mathcal{H}}}{k}+ \sqrt{\frac{V_{\mathcal{H}}}{k}}\Big) \Big\} \\\le 3 \exp \Big( \frac{-k\varepsilon^2}{16(8M^4+M^2\varepsilon/3)} \Big),    
  \end{gathered}
\end{equation*}
which concludes the proof after solving for $3 \exp \big( -k\varepsilon^2/(16(8M^4+M^2\varepsilon/3) ) \big)=\delta$.

\subsection{Proof of Proposition~\ref{prop:prop2}}\label{sec:proof_prop2}
  %\begin{enumerate}
\hspace{\parindent}1. For $t\ge 1$ and $h\in\mathcal{H}$, write $r_t(h) = R_t(h\circ\theta)$. 
For all $h_1,h_2 \in \mathcal{H}$, and $t\ge 1$,  we have
\begin{align}
 & |r_t(h_1) - r_t(h_2)|%  \\
  % &
    = | R_t(h_1 \circ \theta ) - R_t(h_2 \circ \theta ) |  \nonumber \\
   &=\Big|\; \EE [h_1(X)^2 - h_2(X)^2 + 2Y (h_1(X)-h_2(X)) \mid \lVert X\rVert  \ge t] \Big|
  \nonumber \\
  &\le \EE \big[ |(h_1(X) +h_2(X))(h_1(X) - h_2(X))| \mid \lVert X\rVert  \ge t \big]  \nonumber \\
 &+    2\EE \big[ |Y (h_1(X) - h_2(X)) | \mid \lVert X\rVert  \ge  t \big] \nonumber \\
&\le 4M \lVert  h_1 - h_2 \rVert _\infty \label{eq:boundrt}, 
\end{align}
where we have used Assumption~\ref{hyp:bound} to obtain the last inequality.
Similarly,
\begin{align}
  &\riskext(h_1\circ\theta) - \riskext(h_2\circ\theta) \nonumber \\
  &\le \EE  |(h_1(\Theta_\infty) +h_2(\Theta_\infty))(h_1(\Theta_\infty) - h_2(\Theta_\infty))|   \nonumber \\
  % & \quad \dots
 &+    2\EE |Y_\infty (h_1(\Theta_\infty) - h_2(\Theta_\infty )) | \nonumber\\
&\le 4M \lVert  h_1 - h_2 \rVert  _\infty, \label{eq:boundrinf}
\end{align}
Let $\varepsilon>0$. By total boundedness there exists a family
$ h_1,\ldots, h_L\in\mathcal{H}$ such that
$\mathcal{H} \subset \cup_{i=1,\ldots, L} B(h_i,\varepsilon)$. Here
$B(h,\varepsilon)$ denotes the ball of radius $\varepsilon$ in
$(\mathcal{C}(\sphere), \lVert \,\cdots\,\rVert )$. Now because of
Assumption~\ref{hyp:limit} (see Theorem~\ref{prop:conv_bayes_risk},
(i)) we have $r_t(h_i)\to \riskext(h_i\circ\theta)$ as $t\to\infty$,
for all fixed $i$. Thus there exists some $T>0$ such that for all
$i\in\{1,\ldots, L\}$
$|r_t(h_i) - \riskext(h_i\circ\theta) | \le \varepsilon$. Now for any
$h\in\mathcal{H}$ and $t\ge T$, using~\eqref{eq:boundrt}
and~\eqref{eq:boundrinf} there exists $i\le L$ such that
$$\max(|r_t(h) - r_t(h_i)| , |\riskext(h\circ\theta) - \riskext(h_i\circ\theta)|\le 4M\varepsilon, $$ so that
\begin{align*}
  |r_t(h) - \riskext(h\circ\theta)| 
  &\le |r_t(h) - r_t(h_i)| + |r_t(h_i) - \riskext(h_i\circ\theta)|  \\
  &+ [\riskext(h_i\circ\theta) - \riskext(h\circ\theta)| \\
  &\le 8 M \varepsilon + \varepsilon. 
\end{align*}
Because $\riskext(h\circ\theta) = \riskinf(h\circ\theta)$ (Theorem~\ref{prop:conv_bayes_risk}-(i)), the proof is complete. 

2. The VC-class property of $\mathcal{H}$
  (Assumption~\ref{hyp:class}) ensures that for any probability
  measure $Q$ on $\sphere$, and any $\varepsilon>0$, the covering
  number $\mathcal{N}(\varepsilon,\; \mathcal{H},\; L_1(Q) )$ is
  finite (see e.g.,
  \cite{vandervaartWeakConvergenceEmpirical1996}, Section 2.6.2). Our
  first step is to build such a probability measure $Q$ which dominates
  both the $\Phi_{\theta,t}$'s and $\Phi_{\theta}$, in such a way that
  $\EE[ |h_1 - h_2|(\Theta)\, |\, \lVert X\rVert >t]$ and
  $\EE[|h_1 - h_2|(\Theta_\infty)]$ are both controlled by
  $\int_\sphere |h_1- h_2|\ud Q  = \lVert h_1 - h_2\rVert _{L_1(Q)}$. % .

  Let $Q = \frac{1}{2}(\Phi_{\theta,1} + \Phi_{\theta})$. Then
  $\Phi_{\theta}$ is absolutely continuous with respect to $Q$, and so
  is each $\phi_t, t\ge 1$, in view of the discussion above the
  statement in the main paper. In addition we have
  $\sup_{\omega \in \sphere} |\ud \Phi_{\theta} /\ud Q (\omega)| \le 2
  $ and from Condition 2.  also
  $\sup_{\omega \in \sphere, t\ge 1} |\ud \Phi_{\theta,t} /\ud Q
  (\omega)| \le 2 D$.

  For any $h_1, h_2$
  in $\mathcal{H}$, following the argument leading to~\eqref{eq:boundrt} we obtain 
\begin{align*}
  \left\vert  r_t(h_1)- r_t(h_2)  
    \right\vert  &
  \leq 4M\int_{\sphere}\left\vert h_1 -h_2 \right\vert  \ud \Phi_t \\
&  \leq 8MD\int_{\sphere}\left\vert h_1 -h_2 \right\vert \ud Q   = 8MD\,\lVert h_1 - h_2\rVert _{L_1(Q)}. 
\end{align*}

Also, we have%Similarly,
\begin{align*}
  & | \riskext(h_1\circ\theta) - \riskext(h_2\circ\theta) | \\
  &\le \EE  |(h_1g(\Theta_\infty) +h_2(\Theta_\infty))(h_1(\Theta_\infty) - h_2(\Theta_\infty))|  \\%  + \dots \nonumber \\
  % & \quad \dots
   &+  2\EE |Y_\infty (h_1(\Theta_\infty) - h_2(\Theta_\infty )) | \nonumber\\
  &\le 4M \EE[ | h_1 - h_2|(\Theta_\infty)] \le 8 M \lVert  h_1 - h_2\rVert _{L_1(Q)} . % \label{eq:boundrinf}
\end{align*}
Let $\varepsilon>0$.  Since the covering number of the class
$\mathcal{H}$ for the $L_1(Q)$-norm is finite, for some
$L\le \mathcal{N}(\varepsilon,\; \mathcal{H},\; L_1(Q) )$, there
exists $h_1,\; \ldots,\; h_L \in \mathcal{H}$ such that each
$h\in\mathcal{H}$ is at $L_1(Q)$-distance at most $\varepsilon$ from
one of the $h_i$'s. The rest of the proof follows the same lines as
the argument following~\eqref{eq:boundrinf}, up to replacing the
infinity norm with the $L_1(Q)$-norm on $\mathcal{H}$.

\end{appendix}

%%%%%%%%%%%%%%%%%%%%%%%%%%%%%%%%%%%%%%%%%%%%%%
%% Acknowledgements                         %%
%% should be provided in the                %%
%% Acknowledgements section.                %%
%%%%%%%%%%%%%%%%%%%%%%%%%%%%%%%%%%%%%%%%%%%%%%
\begin{acks}[Acknowledgments]
The authors would like to thank the anonymous referees, an Associate
Editor and the Editor for their constructive comments that improved the
quality of this paper.
\end{acks}

%%%%%%%%%%%%%%%%%%%%%%%%%%%%%%%%%%%%%%%%%%%%%%
%% Funding information, if any,             %%
%% should be provided in the                %%
%% funding section.                         %%
%%%%%%%%%%%%%%%%%%%%%%%%%%%%%%%%%%%%%%%%%%%%%%
\begin{funding}
  Anne Sabourin acknowledges the support of the French ANR (ANR
  project EXSTA, ANR-23-CE40-0009). Nathan Huet's research was funded
  by Hi! Paris and the `Programme d'Intelligence Artificielle de l'IP
  Paris' from the French ANR.
\end{funding}

\bibliographystyle{imsart-nameyear} % Style BST file (imsart-number.bst or imsart-nameyear.bst)
\bibliography{bibliExtremeClassif}

\end{document}